\documentclass[fontsize=12pt]{article}

\usepackage[english]{babel} 
\usepackage{amsmath,amsfonts,amsthm} 
\usepackage[utf8]{inputenc}
\usepackage{float}
\usepackage{lipsum} 
\usepackage{blindtext}
\usepackage{graphicx} 
\usepackage{caption}
\usepackage[sc]{mathpazo} 
\usepackage[T1]{fontenc} 
\usepackage{microtype} 
\usepackage[hmarginratio=1:1,top=32mm,columnsep=20pt]{geometry} 
\usepackage{multicol} 
\usepackage{booktabs} 
\usepackage{float} 
\usepackage{hyperref} 
\usepackage{lettrine} 
\usepackage{paralist} 
\usepackage{abstract} 
\usepackage{titlesec} 
\usepackage{natbib, url}

\usepackage{times}
\usepackage[T1]{fontenc}
\usepackage{microtype}
\usepackage{multicol}
\usepackage{xcolor}
\usepackage{comment}
\usepackage{tikz}
\usetikzlibrary{intersections}
\usepackage{newtxmath}
\usepackage[justification=centering]{subfig}
\usepackage[utf8]{inputenc}
\usepackage{multirow}
\usepackage[scale=0.90]{sourcecodepro}
\usepackage{algorithm}
\usepackage[noend]{algpseudocode}
\usepackage[normalem]{ulem} 

\makeatletter
\def\algbackskip{\hskip-\ALG@thistlm}
\makeatother

\renewcommand\thesection{\Roman{section}} 

\titleformat{\section}[block]{\Large\bfseries\filright}{\thesection.}{1em}{} 
\titleformat{\subsection}[hang]{\bfseries\filright}{\thesubsection}{1em}{} 

\usepackage{fancyhdr} 
\usepackage[bottom]{footmisc}

\numberwithin{equation}{section}

\newcommand{\idenI}{\textrm{I}}

\newcommand{\normal}{\textrm{normal}} 
\newcommand{\samp}[2]{#1^{(#2)}}
\newcommand{\rmsup}[2]{#1^{\textrm{#2}}}
\newcommand{\bigo}[1]{\mathcal{O}(#1)}

\newenvironment{algo-fig}[2]{
	\begin{figure}[t!]
		$\textbf{#1}(#2)$
		\vspace*{2pt}
		\hrule
	}{
	\end{figure}
}

\newenvironment{algo-args}{
	\vspace*{-2pt}
	\begin{itemize}\setlength{\itemsep}{-2pt}
	}{
	\end{itemize}\setlength{\itemsep}{0pt}
}

\newenvironment{algo-algo}{
	\vspace*{-4pt}
	\textit{Algorithm}:\vspace*{-4pt}
	\begin{enumerate}\setlength{\itemsep}{-2pt}
	}{
	\end{enumerate}\setlength{\itemsep}{0pt}
}

\newenvironment{algo-return}{
	\vspace*{-8pt}
	\textit{Return}:
	\vspace*{-4pt}
	\begin{itemize}\setlength{\itemsep}{-2pt}
	}{
	\end{itemize}\setlength{\itemsep}{0pt}
	\hrule
}

\graphicspath{{./pics/}}

\title{\vspace{-15mm}\fontsize{24pt}{10pt}\selectfont\textbf{Pathfinder: Parallel quasi-Newton variational inference}} 

\author{
\large
{\textsc{Lu Zhang}}\\
{\normalsize Department of Statistics, Columbia University}\\
{\normalsize \href{mailto:lz2786@columbia.edu}{lz2786@columbia.edu}}\\
\\
\large
{\textsc{Bob Carpenter}}\\
{\normalsize Center for Computational Mathematics, Flatiron Institute}\\
{\normalsize \href{mailto:bcarpenter@flatironinstitute.org}{bcarpenter@flatironinstitute.org}}\\
\\
\large
{\textsc{Andrew Gelman}}\\
{\normalsize Departments of Statistics and Political Science, Columbia University}\\
{\normalsize \href{mailto:gelman@stat.columbia.edu}{gelman@stat.columbia.edu}}\\
\\
\large
{\textsc{Aki Vehtari}}\\
{\normalsize Department of Computer Science, Aalto University}\\
{\normalsize \href{mailto:aki.vehtari@aalto.fi}{aki.vehtari@aalto.fi}}\\
}
\providecommand{\keywords}[1]{\textbf{\textit{Keywords:}} #1}
\begin{document}
\maketitle 
\thispagestyle{fancy} 

\label{firstpage}

\begin{abstract}
We propose Pathfinder, a variational method for approximately sampling from differentiable log densities.  
Starting from a random initialization, Pathfinder locates normal approximations to the target density along a quasi-Newton optimization path, with local covariance estimated using the inverse Hessian estimates produced by the optimizer.  Pathfinder returns draws from the approximation with the lowest estimated Kullback-Leibler (KL) divergence to the true posterior.  

We evaluate Pathfinder on a wide range of posterior distributions, demonstrating that its approximate draws are better than those from automatic differentiation variational inference (ADVI) and comparable to those produced by short chains of 
dynamic Hamiltonian Monte Carlo (HMC), as measured by 1-Wasserstein distance.  Compared to ADVI and short dynamic HMC runs, Pathfinder requires one to two orders of magnitude fewer log density and gradient evaluations, with greater reductions for more challenging posteriors.  Importance resampling over multiple runs of Pathfinder improves the diversity of approximate draws, reducing 1-Wasserstein distance further and providing a measure of robustness to optimization failures on plateaus, saddle points, or in minor modes.  The Monte Carlo KL divergence estimates are embarrassingly parallelizable in the core Pathfinder algorithm, as are multiple runs in the resampling version, further increasing Pathfinder's speed advantage with multiple cores.
\end{abstract}

\keywords{Variational inference; Hamiltonian Monte Carlo; quasi-Newton optimization; Laplace approximation; importance resampling}
\newpage
\section{Introduction}
Obtaining efficient, scalable, and robust posterior inference remains the primary challenge in advanced Bayesian computation.  The difficulty of this challenge has led researchers to develop a wide range of posterior approximation algorithms.  One of the most popular classes of approximate methods is variational inference (VI), which searches for a tractable approximate distribution that minimizes Kullback-Leibler (KL) divergence to the posterior. Although VI is typically faster than Monte Carlo sampling \citep{blei2017variational}, popular approaches such as black-box variational inference \citep{ranganath2014black} and automatic differentiation variational inference \citep{kucukelbir2017automatic} can fail to converge due to the high variance of nested gradient estimates~\citep{dhaka2021challenges}.

In this paper, we develop Pathfinder, an algorithm that locates approximations to the target density along a quasi-Newton optimization path. Starting from a random initialization in the tail of the posterior distribution, the quasi-Newton optimization trajectory can quickly move from the tail, through the body of the distribution, to a mode or pole. By evaluating the ELBO in parallel for the normal approximations along the optimization path generated by L-BFGS, a popular quasi-Newton method, Pathfinder can quickly find a region of high probability mass from which to draw approximate samples.
Novel contributions of this paper include (1) new VI algorithms that use curvature information of the target distribution collected by optimization trajectories to propose approximate distributions; (2) an efficient sampling algorithm for the normal approximations estimated from quasi-Newton inverse Hessian approximations; (3) the design of Pathfinder, which allows evaluating the evidence lower bound (ELBO) in parallel for each normal approximation. Hence, Pathfinder can be greatly accelerated by parallel computing, which is not possible with existing VI algorithms that directly minimize KL divergence, all of which are sequential.

\begin{figure}[t!]
	\centering{
		\includegraphics[width=0.95\textwidth, keepaspectratio]{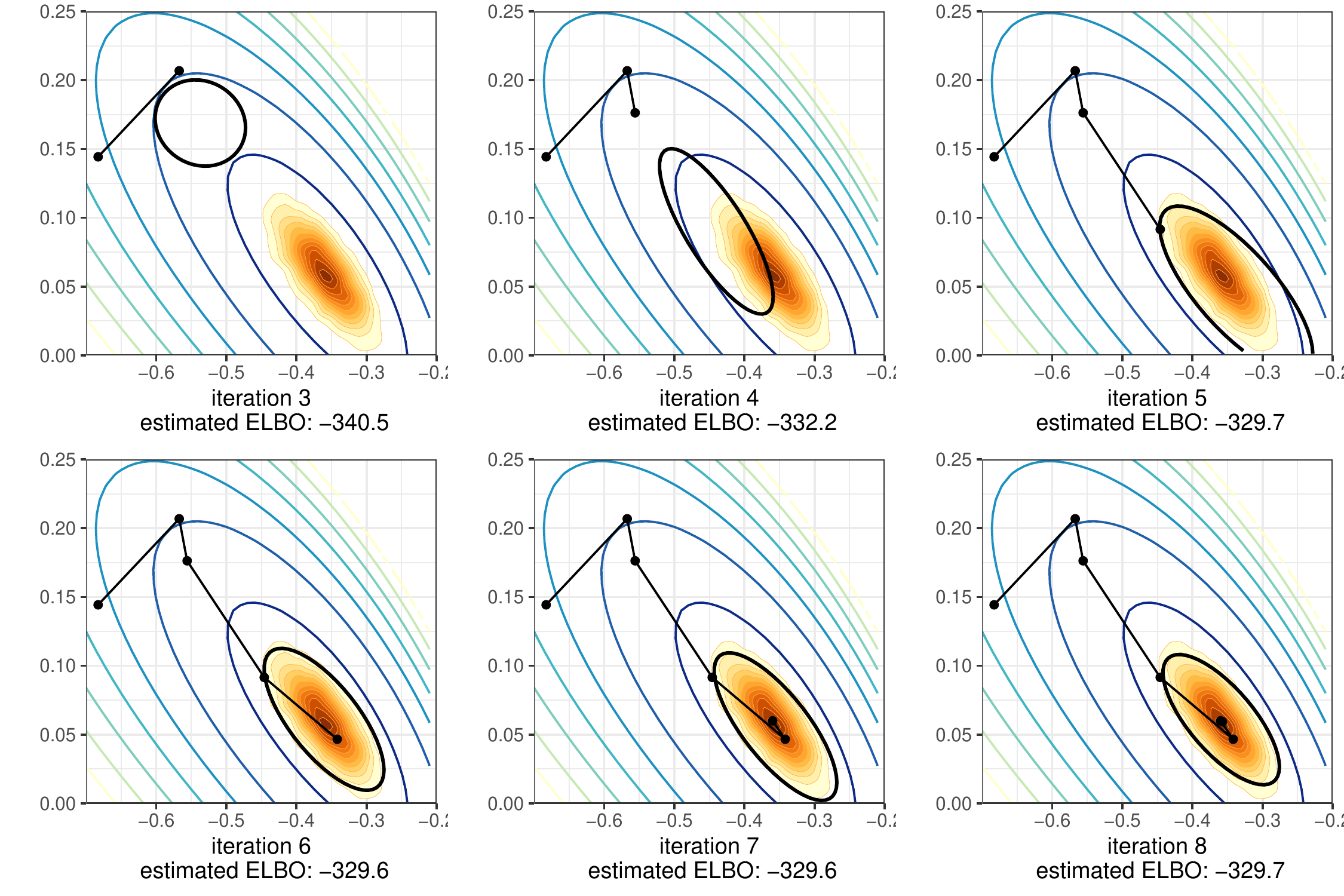}}
	\caption{\it A series of normal approximations (black ellipse for 95\% central region) along a quasi-Newton optimization path (black polyline) for the posterior of a logistic regression model (green to blue contours), whose high probability mass region is indicated by the yellow to orange contours. As the optimization path reaches the mode, the central 95\% region of the normal approximation closely matches the high probability mass region of the target density.
	}\label{fig: illustrate_pics1}
\end{figure}
Figure~\ref{fig: illustrate_pics1} illustrates the evolution of approximate posteriors along the optimization path for a density with a single mode.  In cases with posterior modes, Pathfinder provides a Laplace-like approximation of the posterior density using the quasi-Newton optimizer's efficient inverse Hessian estimate for covariance.
\begin{figure}[t!]
	\centering{
		\includegraphics[width=0.95\textwidth, keepaspectratio]{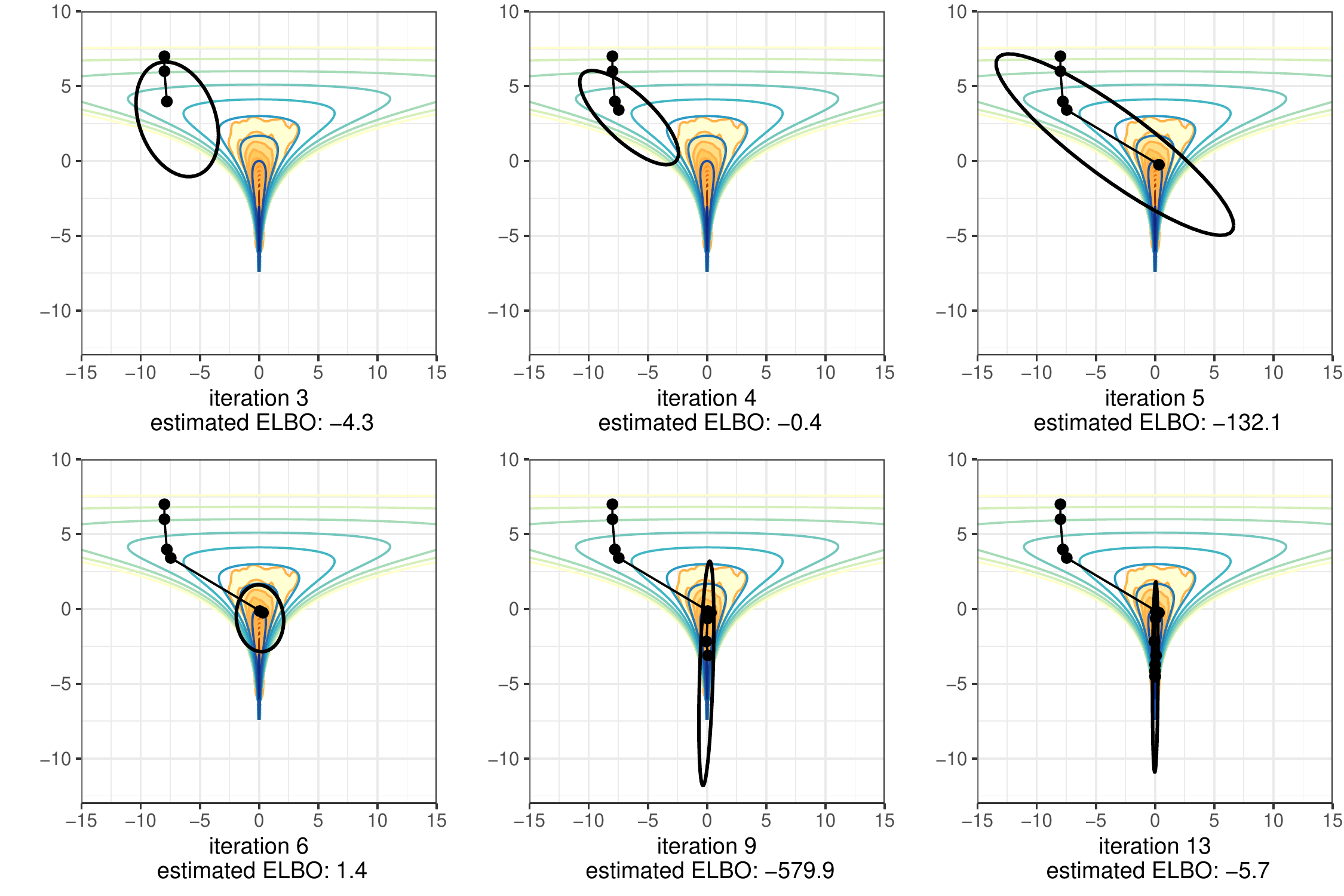}}
	\caption{\it A series of normal approximations (black ellipse for 95\% central region) along a quasi-Newton optimization path (black polyline) for a funnel-like posterior density with no mode (green to blue contours), whose high probability mass region is indicated by the yellow to orange contours. The normal approximations with the highest ELBO value (lower left) occurs at a point on the optimization trajectory before it heads off toward the pole at negative infinity on the vertical axis.  The ELBO values rise and fall along the optimization path.}\label{fig: illustrate_pics2}
\end{figure}
Figure~\ref{fig: illustrate_pics2} shows how Pathfinder behaves for unbounded target densities like the funnel, where it balances the competing goals of high entropy and containment within the target density to stop before heading off to a pole.  In both cases, the use of approximate inverse Hessian information allows the quasi-Newton optimizer to move quickly and stably into and through the high probability region of the posterior.

Multimodal distributions can be approximated by running several instances of Pathfinder in parallel from different initialization points, followed by filtering with importance resampling.  In Section~\ref{resampling-pathfinder-algorithm.sec}, we describe this multi-path version of Pathfinder. We adapt importance sampling using Pareto smoothed importance weights \citep{yao2018yes, vehtari2019pareto} to importance resampling to obtain more stable approximate draws.

We evaluate the performance of Pathfinder experimentally in Section~\ref{sec: simulation}. We compare Pathfinder to automatic differentiation variational inference (ADVI), a state-of-the-art variational inference algorithm \citep{kucukelbir2017automatic}.\footnote{We also evaluated a robust version of ADVI from \cite{dhaka2020robust}, which provided similar results in its mean-field form, but the implementation failed to converge in its dense form.}  We also compare to the approximate draws generated by running many short MCMC chains with dynamic Hamiltonian Monte Carlo in the form of the no-U-turn sampler \citep{hoffman2014no} as refined by \cite{betancourt2017conceptual}.\footnote{We use Stan's implementation of ADVI and dynamic HMC \citep{stan2021ref}.}  The short MCMC chains can be viewed as the first stage of warmup for MCMC sampling or as a variational inference algorithm in its own right, following \cite{hoffman2020black}.  

We evaluate approximations to the target density based on the discrete form of 1-Wasserstein distance, with the target density defined by long runs of dynamic HMC thinned to roughly independent draws.  Wasserstein distance measures how much one distribution would need to be distorted to match the other.  Unlike the asymmetric KL divergence measure, Wasserstein distance is a proper distance metric obeying symmetry and the triangle inequality.

Over a diverse set of 20 models from the \texttt{posteriordb} evaluation set \citep{mans2021posteriordb}, we found Pathfinder's approximations ranged from slightly worse to much better than those of ADVI using diagonal covariance (mean field), ADVI with dense covariance (full rank), and dynamic HMC using short chains (75 iterations).  Pathfinder required one to two orders of magnitude fewer log density and gradient evaluations than these systems \textit{without parallelization}.  We further explore Pathfinder's features and limitations based on case studies of high-dimensional models in Section~\ref{sec: simulation}. 

Although we frame Pathfinder as a form of variational inference, its development was motivated by the question of how to efficiently generate a handful of approximate draws with which to initialize asymptotically exact Markov chain Monte Carlo (MCMC) methods such as dynamic HMC.   The best initialization for MCMC is a draw from the posterior, as that leads to a stationary Markov chain.  Initializing MCMC with an approximate draw from the posterior allows us to skip this first stage of MCMC adaptation (sometimes called ``burn-in'').  In Section~\ref{sec: Birthday}, we demonstrate the benefits of using Pathfinder to initialize MCMC through an analysis of a Gaussian process model. 

\section{Pathfinder}\label{sec: pathfinder}

This section describes the Pathfinder algorithm for generating approximate draws from a differentiable target density known only up to a normalizing constant. 
We follow the presentation of the basic Pathfinder algorithm in Section~\ref{pathfinder-algorithm.sec} with a multi-path version in Section~\ref{resampling-pathfinder-algorithm.sec}, which runs multiple optimization paths and uses importance resampling to select draws. Resampling from multiple normal approximations better matches  non-normal target densities and also mitigates the problem of L-BFGS getting stuck at local optima or in saddle points on plateaus. 
To discriminate the Pathfinder algorithms in Section~2.1 and Section~2.2, we refer to the former one as single-path Pathfinder and the latter one as multi-path Pathfinder. The remaining sections provide details of the algorithms used in the inner loop of Pathfinder. Section~\ref{subsec: L-BFGS_optim} provides relevant details of L-BFGS optimization, Sections~\ref{subsec: V_local_Gauss_approx} and \ref{sec:sampling} present the algorithm for evaluating and sampling from the normal approximations in Pathfinder, Section~\ref{subsec: est_DIV} explains the Monte Carlo evaluations of the evidence lower bound, and Section~\ref{subsec: PS-IR} describes the importance resampling algorithm implemented in the multi-path Pathfinder algorithm.   We review connections to related methods in Section~\ref{sec:related}.
\subsection{Pathfinder algorithm}\label{pathfinder-algorithm.sec}
The Pathfinder algorithm begins by drawing from an initialization distribution $\pi_0$, from which it follows a quasi-Newton optimization trajectory. We use L-BFGS to generate an optimization trajectory $\samp{\theta}{0:L} = (\samp{\theta}{0}, \ldots, \samp{\theta}{L})$ towards a local maximum (or pole) of the log density $\log p(\theta)$, where $\samp{\theta}{0}$ denotes the initial point and the superscript indicates the iteration. In applications to Bayesian posterior sampling, the target density $p(\theta)$ is the posterior $p(\theta \mid y)$, where $\theta \in \mathbb{R}^N$ represents the $N$-dimensional parameter vector and $y$ denotes the observations. The exact posterior is often intractable, and practitioners have to resort to iterative algorithms like MCMC and VI to obtain posterior samples or inferences based on a log probability function, which is the log density of the posterior up to an additive constant. For the ease of explanation, we use $\log p(\theta)$ to refer to the tractable log probability function. Based on the exploration of the optimization trajectory, Pathfinder generates local normal approximations of the target density using the gradient and curvature information collected along the optimization trajectory. To obtain the approximations, we develop a function {\sc $\alpha$-recover} that uses the optimization trajectory and the gradient along the optimization path $\nabla \log p(\samp{\theta}{0:L})$ to compute a diagonal estimate of the covariance matrix of the approximation for each iteration. The {\sc $\alpha$-recover} routine returns the diagonal elements of the covariance estimation for all iterations $\samp{\alpha}{1:L}$, and provides indicators $\samp{\xi}{1:L}$ of whether a pair of the updates of the position and gradient along the optimization path should be used in further covariance estimation or not. It also returns the updates of locations $\samp{s}{1:L}$ and gradients of the negative log joint density $\samp{z}{1:L}$ along the optimization path, where $\samp{s}{l} = \samp{\theta}{l} - \samp{\theta}{l-1}$ and $\samp{z}{l} = \nabla \log p(\samp{\theta}{l-1}) - \nabla \log p(\samp{\theta}{l})$. We describe the algorithm {\sc $\alpha$-recover} and provide pseudocode in Section~\ref{subsec: V_local_Gauss_approx}. Then for each iteration, we generate a local approximation based on the second-order Taylor series expansion and the quasi-Newton inverse Hessian estimate of covariance.  We further develop an evaluation algorithm that generates samples and estimates the evidence lower bound (ELBO) in parallel for all local approximations. In particular, we propose an efficient sampling algorithm {\sc BFGS-sample} in Section~\ref{sec:sampling}, which returns $K$ samples $\samp{\phi}{l, 1:K}$ and their log-densities under the approximation $\log q(\samp{\phi}{l, 1:K})$ for the local approximation at iteration $l$. Here we use double superscripts to distinguish samples from different approximations, the first superscript indexes the approximation and the second superscript indexes the draws. We calculate a Monte Carlo estimate of the ELBO $\samp{\lambda}{l}$ for approximation $l$ with Algorithm~\ref{ELBO-est.fig} in Appendix~\ref{appendix: pseudocode}. The last step in Pathfinder selects the approximation that maximizes the evidence lower bound (equivalently, minimizes Kullback-Leibler divergence to the target density), and then generates $M$ draws from the best normal approximation. 
Algorithm~\ref{pathfinder-algorithm.fig} presents pseudocode for Pathfinder. We later show in Section~\ref{sec: simulation} that Pathfinder is not particularly sensitive to its tuning parameters, which include an initial distribution, maximum number of L-BFGS iterations ($\rmsup{L}{max}$), L-BFGS convergence tolerance ($\rmsup{\tau}{rel}$), size of the history used to approximate the inverse Hessian ($J$), and the number of Monte Carlo draws used to evaluate the ELBO ($K$).

\begin{algorithm}
	\caption{Single-path Pathfinder 
	}\label{pathfinder-algorithm.fig}
	\hspace*{\algorithmicindent} \textbf{Input:} \\
	\hspace*{\algorithmicindent} \hspace*{\algorithmicindent} $\log p$: differentiable log density function of dimension $N$\\
	\hspace*{\algorithmicindent} \hspace*{\algorithmicindent} $\pi_0$: initial distribution \\
	\hspace*{\algorithmicindent} \hspace*{\algorithmicindent} $\rmsup{L}{max}$: maximum number of L-BFGS iterations \\
	\hspace*{\algorithmicindent} \hspace*{\algorithmicindent} $\rmsup{\tau}{rel}$: relative tolerance for convergence of L-BFGS \\
	\hspace*{\algorithmicindent} \hspace*{\algorithmicindent} $J$: 
	size of the history used to approximate the inverse Hessian \\
	\hspace*{\algorithmicindent} \hspace*{\algorithmicindent} $K$: number of Monte Carlo draws to evaluate ELBO \\
	\hspace*{\algorithmicindent} \hspace*{\algorithmicindent} $M$: number of approximate posterior draws to return \\
	\hspace*{\algorithmicindent} \textbf{Output:}\\
	\hspace*{\algorithmicindent} \hspace*{\algorithmicindent} $\samp{\psi}{1}, \ldots, \samp{\psi}{M}$: draws from ELBO-maximizing normal approximation \\
	\hspace*{\algorithmicindent} \hspace*{\algorithmicindent} $\log q(\samp{\psi}{1}), \ldots, \log q(\samp{\psi}{M})$: log density of draws in ELBO-maximizing normal approximation
	\begin{algorithmic}[1]
		\Procedure{Pathfinder}{$\log p, \pi_0, L, \rmsup{\tau}{rel}, J, K, M$}
		\State sample $\samp{\theta}{0} \sim \pi_0$
		\State let $\left(\samp{\theta}{0:L},  \nabla \log p(\samp{\theta}{0:L})\right) = \textsc{L-BFGS}\!\left(\log p, \samp{\theta}{0}, J, \rmsup{\tau}{rel}, \rmsup{L}{max} \right)$
		\State let $(\samp{\alpha}{1:L}, \samp{\xi}{1:L}, \samp{s}{1:L}, \samp{z}{1:L}) = \textsc{$\alpha$-recover}\!\left(\samp{\theta}{0:L},  \nabla \log p(\samp{\theta}{0:L}), J \right)$ 
		\For{$l \in 1:L$ in parallel} 
		\State let $\samp{\phi}{l, 1:K}, \log q(\samp{\phi}{l, 1:K}) 
		= \textsc{BFGS-sample}(\samp{s}{1:l}, \samp{z}{1:l}, \samp{\theta}{l}, \nabla \log p(\samp{\theta}{l}), \samp{\alpha}{l}, \samp{\xi}{1:l}, K)$ 
		\For{$k \in 1:K$}
		\State evaluate and store $\log p(\samp{\phi}{l, k})$
		\EndFor
		\State let $\samp{\lambda}{l} 
		= \textsc{ELBO}\!\left(\log p(\samp{\phi}{l, 1:K}), \log q(\samp{\phi}{l, 1:K})\right)$
		\EndFor
		\State let $l^* = \textrm{arg max}_l \, \samp{\lambda}{l}$
		\State let $\samp{\psi}{1:M}, \log q(\samp{\psi}{1:M})
		= \textsc{BFGS-sample}(\samp{s}{1:l^\ast}, \samp{z}{1:l^\ast}, \samp{\theta}{l^\ast}, \nabla \log p(\samp{\theta}{l^\ast}), \samp{\alpha}{l^\ast}, \samp{\xi}{1:l^\ast}, M)$ 
		\EndProcedure
	\end{algorithmic}
\end{algorithm}

The computational cost of single-path
Pathfinder is dominated by the  (1) L-BFGS optimization, (2) sampling from the approximate normal distributions, and (3) evaluating the evidence lower bounds. The cost of L-BFGS optimization is dominated by log density and gradient evaluations, the number of which will be determined by the tuning parameters of L-BFGS, including  $\rmsup{\tau}{rel}$, $\rmsup{L}{max}$, and $J$. The cost of sampling from the normal approximations is modest because we efficiently rescale and rotate standard normal draws using the factored L-BFGS covariance approximation. Estimating the evidence lower bound requires a number of log density evaluations equal to the number of Monte Carlo draws~($K$). The accuracy and reliability of the Monte Carlo estimates can be diagnosed without regard to a reference distribution using a diagnostic based on the Pareto $k$ statistic \citep{vehtari2019pareto,dhaka2021challenges}.  Sampling from the normal approximation and estimating the evidence lower bound may be done in parallel across the points on the optimization trajectory. This makes L-BFGS the serialization bottleneck for this algorithm. L-BFGS is economical in its calls to gradient and log density functions because of its ability to leverage quasi-Newton estimates of local curvature.

In some cases, the optimization path terminates at the initialization point and in others it can fail to generate a positive definite inverse Hessian estimate.  In both of these settings, Pathfinder essentially fails.  Rather than worry about coding exceptions or failure return codes, Pathfinder returns the last iteration of the optimization path as a single approximating draw with $\infty$ for the approximate normal log density of the draw.  This ensures that failed fits get zero importance weights in the multi-path Pathfinder algorithm, which we describe in the next section.

\subsection{Multi-path Pathfinder algorithm}\label{resampling-pathfinder-algorithm.sec}

\begin{algorithm}
	\caption{Multi-path Pathfinder 
	}\label{resampling-pathfinder-algorithm.fig}
	\hspace*{\algorithmicindent} \textbf{Input:} \\
	\hspace*{\algorithmicindent} \hspace*{\algorithmicindent} $\log p$: differentiable log density function of dimension $N$\\
	\hspace*{\algorithmicindent} \hspace*{\algorithmicindent} $\pi_0$: initial distribution\\
	\hspace*{\algorithmicindent} \hspace*{\algorithmicindent} $\rmsup{L}{max}$: maximum number of L-BFGS iterations\\
	\hspace*{\algorithmicindent} \hspace*{\algorithmicindent} $\rmsup{\tau}{rel}$: relative tolerance for convergence of L-BFGS\\
	\hspace*{\algorithmicindent} \hspace*{\algorithmicindent} $J$: size of the history used to approximate the inverse Hessian\\
	\hspace*{\algorithmicindent} \hspace*{\algorithmicindent} $K$: number of Monte Carlo draws to evaluate ELBO\\
	\hspace*{\algorithmicindent} \hspace*{\algorithmicindent} $I$: number of independent Pathfinder runs\\
	\hspace*{\algorithmicindent} \hspace*{\algorithmicindent} $M$: number of draws returned by each Pathfinder run\\
	\hspace*{\algorithmicindent} \hspace*{\algorithmicindent} $R$: number of draws returned by importance resampling ($R \ll IM$)\\
	\hspace*{\algorithmicindent} \textbf{Output:}\\
	\hspace*{\algorithmicindent} $\samp{\psi}{1}, \ldots, \samp{\psi}{R}$: approximate draws from target density $p$
	\begin{algorithmic}[1]
		\Procedure{MultipathPathfinder}{$\log p, \pi_0, \rmsup{L}{max}, \rmsup{\tau}{rel}, J, K, I, M, R$}
		\For{$i \in 1:I$ in parallel}
		\State let $\samp{\phi}{i, 1:M},
		\log q(\samp{\phi}{i, 1:M})
		= \textrm{Pathfinder}(\log p, \pi_0, \rmsup{L}{max}, \rmsup{\tau}{rel}, J, K, M)$
		\State compute and store target log densities $\log p(\samp{\phi}{i, 1}), \ldots, \log p(\samp{\phi}{i, M})$
		\EndFor
		\State let $(\samp{\psi}{1}, \samp{i}{1}), \ldots, (\samp{\psi}{R}, \samp{i}{R}) = \textrm{PS-IR}(\samp{\phi}{1:I, 1:M}, \log \tilde{q}(\samp{\phi}{1:I, 1:M}), \log \tilde{p}(\samp{\phi}{1:I, 1:M}), R)$, \\
		\hspace*{\algorithmicindent} where  $\log \tilde{q}(\samp{\phi}{i, m}) = \log \frac{1}{I} \cdot  \textrm{multi-normal}(\samp{\phi}{i, m} \mid \samp{\mu}{i}, \samp{\Sigma}{i})) = \log q(\samp{\phi}{i, m}) - \log(I)$
		\\
		\hspace*{\algorithmicindent} and $\log \tilde{p}(\samp{\phi}{i, m}) = \log \frac{1}{I} \cdot p(\samp{\phi}{i, m}) = \log p(\samp{\phi}{i, m}) - \log(I)$.
		\EndProcedure
	\end{algorithmic}
\end{algorithm}
The multi-path Pathfinder algorithm is given in Algorithm~\ref{resampling-pathfinder-algorithm.fig}. It runs Pathfinder $I$ times in parallel. For run $i$ of Pathfinder, it saves the approximate samples $\samp{\phi}{i, 1:M}$. Then it generates $R$ approximate draws based on all of the approximate draws $\samp{\phi}{1:I, 1:M}$ by importance resampling. The resulting approximation generalizes the normal distribution of Pathfinder to a mixture of $I$ normal distributions, which improves approximations for distributions that are far from normal. Importance resampling from a mixture of normals also reduces the variability arising from single runs using random initial values and stochastic ELBO estimates to choose the best normal approximation along the trajectory. In particular, for more than one run of Pathfinder, we augment the parameter space to include the discrete index $i \in 1{:}I$ indicating the mixture component that generated the parameters.  With the assumption that all runs of Pathfinder return the same number of draws and that the mixture components are equally weighted, the resulting proposal distribution of draws and mixture indicator is
\[
\tilde{q}(\phi, i)
=
\frac{1}{I} \cdot \textrm{multi-normal}(\phi \mid \samp{\mu}{i}, \samp{\Sigma}{i}),
\]
where $\textrm{multi-normal}(\samp{\mu}{i}, \samp{\Sigma}{i})$ is the normal approximation selected by run $i$ of Pathfinder. The corresponding joint target distribution can be extended in the same way, to
\[
\tilde{p}(\phi, i) = \frac{1}{I} \cdot p(\phi).
\]
With the augmented parameter space, we can importance resample across different runs of Pathfinder without recomputing the marginal proposal distribution of $\phi$ for all draws. The importance resampling procedure based on several proposal densities from multiple runs of Pathfinder is a form of multiple importance sampling.  More specifically, we use the scheme labeled N1 by \citet{elvira2019generalized}, which has the least computational burden among the proper alternatives and can be computed in parallel in each individual run of Pathfinder.
Furthermore, optimization paths stuck in minor modes, at saddle points, or on plateaus can be eliminated through the natural weighting of importance resampling. 
The second step of the multi-path Pathfinder algorithm in Algorithm~\ref{resampling-pathfinder-algorithm.fig} uses importance resampling, based on the joint (parameter and mixture indicator) log density of the proposal density function $\tilde{q}(\phi, i)$ and target density function $\tilde{p}(\phi, i)$.   The pseudocode for Pareto-smoothed importance resampling (PS-IR) is provided in Algorithm~\ref{ps-sir.fig}.

Multi-path Pathfinder performs $I$ completely independent runs of Pathfinder, so that its expected number of operations is $I$ times as many operations as are expected from Pathfinder.   Because these independent runs can be executed asynchronously and each takes roughly the same amount of work, wall time for multi-path Pathfinder should be only slightly higher than that of Pathfinder.  The importance resampling step is fast, but it requires all runs of Pathfinder to complete before it is executed, making the expected time to run multi-path Pathfinder a bit longer than that of the slowest of the independent Pathfinder chains.  There is a bit of additional parallelizable work to evaluate log densities in the approximation and in the target density.  After this evaluation, resampling only requires normalization, random number generation, and selection, all of which are fast.

\subsection{L-BFGS optimization}\label{subsec: L-BFGS_optim}

For minimizing the objective function $-\log p(\theta)$, Newton steps move in the direction of the inverse Hessian of $-\log p(\theta)$ times the gradient, with all derivatives being respect to the parameter vector $\theta$,
\[
\delta = (-\nabla^2 \log p(\theta ))^{-1} \cdot \nabla \log p(\theta ).
\]
Quasi-Newton methods are so called because they use an approximation of the inverse Hessian. 

The BFGS optimization algorithm is a quasi-Newton method that approximates the inverse Hessian through updates of positions and gradients of the objective function at positions along the optimization path \citep{broyden1970convergence,fletcher1987practical,goldfarb1970family,shanno1970conditioning}. The limited-memory BFGS (L-BFGS) algorithm limits the size of the history of finite differences for greater scalability and to allow the local inverse Hessian estimates to adapt to varying curvature along the optimization path \citep{nocedal1980updating}.  There are several prominent implementations of L-BFGS that vary in their details;  we use the version introduced by \citet{byrd1995limited} and detailed by \citet{zhu1997algorithm}.\footnote{We use the L-BFGS-B implementation in the R function \texttt{stats::optim()} \citep{r2019language}.} The pseudocode for Zhu et al.'s version of L-BFGS (without bounds) is listed in Algorithm~\ref{l-bfgs-algorithm.fig} in Appendix~\ref{appendix: pseudocode}.  

The standard L-BFGS algorithm approximates inverse Hessians using the previous $J$ updates of positions and gradients of the objective function along the optimization path. For iteration $l$, let $S = \begin{bmatrix} S_1 & \cdots & S_J\end{bmatrix}$ and $Z = \begin{bmatrix} Z_1 & \cdots & Z_J \end{bmatrix}$ be $N \times J$ matrices that store the previous $J$ updates of positions and gradients, (i.e., $S_j = \samp{\theta}{l- J + j} - \samp{\theta}{l-J + j - 1}$, $Z_j = \nabla \log p(\samp{\theta}{l- J + j})- \nabla \log p(\samp{\theta}{l-J + j - 1})$ for $j = 1, \ldots, J$). 
Let $\alpha$ be an $N$-vector that stores the diagonal elements of an initial diagonal inverse Hessian estimate. Following \citet[eq.\ 2.6]{byrd1994representations}, the estimated inverse Hessian at iteration $l$ based on the previous $J$ positions can be formulated as 
\begin{equation}\label{eq: inv-Hessian_construct}
	\samp{\Sigma}{l} = \textrm{diag}(\alpha) + \beta \cdot \gamma \cdot \beta^\top\;,
\end{equation}
where  
\begin{equation}\label{eq: beta_gamma}
	\beta = \begin{bmatrix} 
		\textrm{diag}(\alpha) \cdot Z & S
	\end{bmatrix}\;,\;
	\gamma = \begin{bmatrix}
		0 & -\,E^{-1}
		\\[2pt] 
		-\, E^{-\top} & E^{-\top} \cdot \left(\textrm{diag}(\eta) + Z^{\top} \cdot \textrm{diag}(\alpha) \cdot Z\right) \cdot E^{-1}
	\end{bmatrix}\;,
\end{equation}
with
\begin{align*}
	E_{i, j} = \begin{cases}
		S_i^{\top} \cdot Z_{j} & \textrm{if} \ i \leq j
		\\
		0 & \textrm{otherwise}
	\end{cases}\;, 
	\eta = \begin{bmatrix}
		S_{1}^{\top} \cdot Z_{1}
		\ \cdots \
		S_{J}^{\top} \cdot Z_{J}
	\end{bmatrix}\;.		
\end{align*}
Although L-BFGS never explicitly constructs $\samp{\Sigma}{l}$, in Section~\ref{sec:sampling} we show how its factored form can be used to derive an efficient algorithm to sample from the normal distributions with covariance $\samp{\Sigma}{l}$, which is the key step in the Monte Carlo estimator of the ELBO.


\subsection{Local density approximations along the optimization path}\label{subsec: V_local_Gauss_approx}

The proposed normal approximations in Pathfinder are located along the optimization path.  The second-order Taylor series expansion of the target log density $\log p(\theta \mid y)$ at a point $\samp{\theta}{l}$ on the optimization path is
\[
\begin{array}{l}
	\log p(\theta \mid y) 
	\\[4pt]
	\quad \approx \log p(\samp{\theta}{l} \mid y) + \nabla \log p(\samp{\theta}{l} \mid y) \cdot (\theta - \samp{\theta}{l}) - \frac{1}{2} (\theta - \samp{\theta}{l})^\top \cdot \textrm{H}(\samp{\theta}{l}) \cdot (\theta - \samp{\theta}{l})
	\\[4pt]
	\quad = \log \hat{p} (\theta\mid y),
\end{array}
\]
where $\textrm{H}(\theta) = -\nabla^2 \log p(\theta \mid y) = -\nabla^2 \log p(\theta)$ is the Hessian function mapping points to the matrix of second derivatives of $-\log p(\theta)$ with respect to $\theta$ at that point. 
The approximate distribution $\hat{p}(\theta \mid y)$ is a second-order Taylor series expansion around $\samp{\theta}{l}$, which produces a multivariate normal approximation with mean $$\samp{\mu}{l} = \samp{\theta}{l} + \textrm{H}^{-1}(\samp{\theta}{l}) \cdot \nabla \log p(\samp{\theta}{l} \mid y) =  \samp{\theta}{l} + \textrm{H}^{-1}(\samp{\theta}{l}) \cdot \nabla \log p(\samp{\theta}{l})$$ and covariance $\textrm{H}^{-1}(\samp{\theta}{l})$.  At a mode, the first-order term drops out and we are left with a standard Laplace approximation. 

Pathfinder reconstructs the factors of the inverse Hessian approximations as needed using the optimization trajectory, as shown in Algorithm~\ref{Cov-Estimate-algorithm.fig}.  We construct covariance estimates rather than caching the estimates from the optimization algorithm for three reasons.  First, it gives us the flexibility to use a different inverse Hessian approximation than that used in the optimization algorithm.  In our implementation, we use standard L-BFGS to generate an optimization path whose diagonal inverse Hessian estimation $\textrm{diag}(\alpha)$ is a scaled identity matrix, while we use \citet[eq.\ 4.9]{gilbert1989some} in the recovery of the diagonal inverse Hessian estimation to allow the elements in $\alpha$ to vary. Second, the separation allows us to set up a more restricted condition to filter out sharp updates for the inverse Hessian estimation.    In Algorithm~\ref{Cov-Estimate-algorithm.fig} line 5, the condition of selecting the correct pairs $(\samp{s}{l}, \samp{z}{l})$ filters out candidates whose gradient changes $10^{12}$ times more than the position on the direction of the update of gradient. Third, the outputs of the {\sc $\alpha$-recover} procedure enable us to recover the factors in the covariance estimation \eqref{eq: inv-Hessian_construct} for each iteration in parallel without saving matrices $\beta$ and $\gamma$ in \eqref{eq: beta_gamma}, which reduces total memory overhead and reduces communication costs.  We store a sequence of indicators to pick out pairs of positions and gradients to use in covariance estimates. 

%

\begin{algorithm}
	\caption{Diagonal inverse Hessian estimation. 
	}\label{Cov-Estimate-algorithm.fig}
	\hspace*{\algorithmicindent} \textbf{Input:} \\
	\hspace*{\algorithmicindent} \hspace*{\algorithmicindent} $\samp{\theta}{0}, \ldots, \samp{\theta}{L} \in \mathbb{R}^N$: optimization path\\
	\hspace*{\algorithmicindent} \hspace*{\algorithmicindent} $\nabla \log p(\samp{\theta}{0}),
	\ldots, \nabla \log p(\samp{\theta}{L}) \in \mathbb{R}^N$: gradients of log density of $\samp{\theta}{0:L}$ \\
	\hspace*{\algorithmicindent} \hspace*{\algorithmicindent} $J$: size of the history used to approximate the inverse Hessian \\
	\hspace*{\algorithmicindent} \textbf{Output:}\\
	\hspace*{\algorithmicindent} \hspace*{\algorithmicindent} $(\samp{\alpha}{1}, \ldots, \samp{\alpha}{L})$: The diagonal elements of the initial inverse Hessian approximation\\
	\hspace*{\algorithmicindent} \hspace*{\algorithmicindent} $(\samp{\xi}{1}, \ldots, \samp{\xi}{L})$: The indicator of whether the updates of position and gradient are \\
	\hspace*{\algorithmicindent} \hspace*{\algorithmicindent} included in the inverse-Hessian approximation or not.\\
	\hspace*{\algorithmicindent} \hspace*{\algorithmicindent} $(\samp{s}{1}, \ldots, \samp{s}{L})$: The updates of position of the optimization path\\
	\hspace*{\algorithmicindent} \hspace*{\algorithmicindent} $(\samp{z}{1}, \ldots, \samp{z}{L})$: The updates of the gradient ($-\nabla \log p$) of the optimization path\\
	\begin{algorithmic}[1]
		\Procedure{$\alpha$-recover}{$\samp{\theta}{0:L}, \nabla \log p(\samp{\theta}{0:L}), J$}
		\State let $\samp{\alpha}{0} = 1_N$ where $1_N$ denotes the vector of 1's in $\mathbb{R}^N$ 
		\hfill $\bigo{N}$
		\For{$l \in 1:L$} \hfill $\bigo{L J  N}$
		\State let $\samp{s}{l} = \samp{\theta}{l} - \samp{\theta}{l-1}$ and let $\samp{z}{l} = \nabla \log p(\samp{\theta}{l-1}) - \nabla \log p(\samp{\theta}{l})$
		\If {${s}^{(l)\top} \samp{z}{l} < \epsilon \cdot \|\samp{z}{l}\|^2\;$ with $\epsilon = 10^{-12}$}
		\State let $\samp{\xi}{l} = 1$
		\State let $a = z^{(l)\top} \cdot \textrm{diag}(\samp{\alpha}{l-1}) \cdot \samp{z}{l}$; $b = {z}^{(l)\top} \cdot \samp{s}{l}$; $c = {s}^{(l)\top} \cdot \textrm{diag}(\samp{\alpha}{l-1})^{-1} \cdot \samp{s}{l}$
		\hfill $\bigo{N}$ 
		\For{$n \in 1:N$} \hfill $\bigo{N}$
		\State let $\samp{\alpha}{l}_n = \left(\frac{\displaystyle a}{\displaystyle b \cdot \samp{\alpha}{l-1}_n} + \frac{\displaystyle z_n^{(l)2}}{\displaystyle b}- \frac{\displaystyle a \cdot s_{n}^{(l)2}}{\displaystyle b \cdot c \cdot {\samp{\alpha}{l-1}_n}^2}\right)^{-1}$ \hfill $\bigo{1}$
		\EndFor
		\Else 
		\State {let $\samp{\alpha}{l} = \samp{\alpha}{l-1}$ and $\samp{\xi}{l} = 0$}
		\EndIf
		\EndFor
		\EndProcedure
	\end{algorithmic}
\end{algorithm}


Pathfinder evaluates the evidence lower bound for all local approximations to select the best normal approximation. 
By considering all normal approximations from the tail to the mode or pole of the posterior distribution, Pathfinder can quickly find approximations that generate draws in the high probability region of the target density.  

\subsection{Sampling from the approximation and evaluating the log density of a draw}\label{sec:sampling}

To implement Pathfinder, we need to be able to sample from the approximating distributions in order to evaluate the evidence lower bound.  Furthermore, we need to be able to evaluate the log density of these sampled points for importance resampling.  While these operations could be implemented by factoring our inverse Hessian approximations $\samp{\Sigma}{l}$, this would require $\mathcal{O}(N^3)$ operations in $N$ dimensions.  

Fortunately, the outer product representation may be used along with a thin QR factorization and Cholesky factorization in order to produce draws and their log density in $\mathcal{O}(N  J^2 + J^3)$ operations using only $\mathcal{O}(N  J + J^2)$ memory, where $J$ is the history size of L-BFGS and $N$ is the number of dimensions.  That is, the algorithm is linear in dimensionality, with a constant factor determined by the history size for L-BFGS Hessian approximations.

To achieve this efficiency, the Hessian can be factored as
\begin{equation}\label{eq: Hk_reformat}
	\samp{\Sigma}{l} = 
	\textrm{diag}(\alpha^{\frac{1}{2}}) \left( \idenI +\textrm{diag}(\alpha^{-\frac{1}{2}}) \cdot \beta \cdot \gamma \cdot \beta^{\top} \cdot \textrm{diag}(\alpha^{-\frac{1}{2}}) \right) \textrm{diag}(\alpha^{\frac{1}{2}}),
\end{equation}
where we have simplified notation by dropping the superscripts on $\samp{\alpha}{l}$
.  Next, let
\[
Q \cdot \tilde{R} = \textrm{diag}(\alpha^{-\frac{1}{2}}) \cdot \beta
\]
be the thin QR-factorization of $\textrm{diag}(\alpha^{-\frac{1}{2}}) \cdot \beta$, so that $Q$ is an $N \times 2J$ matrix with orthonormal columns and $\tilde{R}$ is a $2J \times 2J$ upper triangular matrix.   Let $P$ be the $N \times (N - 2J)$ matrix with orthonormal columns that makes $\begin{bmatrix} Q & P \end{bmatrix}$ an $N \times N$ orthogonal matrix.  We can then factor the inverse Hessian estimate as
\begin{equation}\label{eq: Hk_factor}
	\samp{\Sigma}{l} = T \cdot T^{\top},
\end{equation}
where
\[
T = \textrm{diag}(\alpha^{\frac{1}{2}})
\cdot
\begin{bmatrix}
	{Q} \cdot {\tilde{L}} & {P}
\end{bmatrix},
\]
and $\tilde{L}$ is defined through Cholesky decomposition to satisfy
$
{\tilde{L}} \cdot {\tilde{L}}^{\top}
= \idenI + {\tilde{R}} \cdot \gamma \cdot {\tilde{R}}^{\top}
$.
We provide a more detailed derivation of \eqref{eq: Hk_factor} in Appendix~\ref{appendix: Hk_factor_deriv}.

If we draw a standard normal $N$-vector
$v \sim \textrm{multi-normal}(0, \textrm{I})$,
then we can translate, scale, and rotate it so that it is a draw from the approximate distribution,
\[
\samp{\mu}{l} + T \cdot v 
\, \sim \,
\textrm{multi-normal}(\samp{\mu}{l}, \samp{\Sigma}{l}),
\]
where the location vector is computed via
\begin{align*}
	\samp{\mu}{l} = \samp{\theta}{l} + \samp{\Sigma}{l} \cdot \nabla \log p(\samp{\theta}{l} \mid y) =\samp{\theta}{l} + \textrm{diag}(\alpha) \cdot \nabla \log p(\theta) + \beta \cdot \gamma \cdot \beta^\top \cdot \nabla \log p(\theta),
\end{align*}
which only involves matrix vector multiplication and requires order $\mathcal{O}(JN + J^2)$ operations and memory.
Next, consider generating $u \sim \textrm{multi-normal}(0, \idenI)$
and setting
\[
v = \begin{bmatrix} {Q} & {P} \end{bmatrix}^{\top} \cdot u.
\]
It follows that $
v \sim \textrm{multi-normal}(0, \textrm{I})
$,
with the orthogonality between columns of ${P}$ and ${Q}$ allowing us to produce  draws $\samp{\mu}{l} + T \cdot v 
\sim \textrm{multi-normal}(\samp{\theta}{l}, \samp{\Sigma}{l})$
defined by
\begin{eqnarray}\label{eq: sample_trans}
	\samp{\mu}{l} + T \cdot v 
	& = & \samp{\mu}{l} +  \textrm{diag}(\alpha^{\frac{1}{2}}) ({Q} \cdot {\tilde{L}} \cdot {Q}^{\top} \cdot u + {P} \cdot {P}^\top \cdot u) \nonumber
	\\[4pt] 
	& = & \samp{\mu}{l} + \textrm{diag}(\alpha^{\frac{1}{2}}) ({Q} \cdot {\tilde{L}} \cdot {Q}^{\top} \cdot u + u - {Q} \cdot {Q}^\top \cdot u)\;.
\end{eqnarray}
Using the factorization in \eqref{eq: Hk_factor}, we can compute the log determinant required for evaluating the approximate log density as
\[
\log \, \big| \, \samp{\Sigma}{l} \,\big| = \log\left|\textrm{diag}(\alpha)\right| + 2\log\left|{\tilde{L}}\right|.
\]
This provides a means to efficiently calculate the log density of the sampled point in the approximating distribution as
\begin{equation}\label{eq: log_q_samples}
	\begin{aligned}
		\log \textrm{normal}&(\samp{\mu}{l} + T \cdot v \mid \samp{\mu}{l}, \samp{\Sigma}{l})  \\
		&= -\,\frac{1}{2}
		\left(\log\big|\samp{\Sigma}{l}\big| + (\samp{\mu}{l} + T v - \samp{\mu}{l})^\top (\samp{\Sigma}{l})^{-1}(\samp{\mu}{l} + T v - \samp{\mu}{l}) + N\log(2\pi)\right)\\
		&= -\,\frac{1}{2}\left(\log\big|\samp{\Sigma}{l}\big| + u^\top u + N\log(2\pi)\right) \;.
	\end{aligned}
\end{equation}
In summary, the sampling and log density evaluation algorithm relies on efficiently decomposing the factored form instead of directly Cholesky decomposing $\samp{\Sigma}{l}$ to generate draws and compute log densities of draws. Algorithm~\ref{bfgs-sample.fig} provides the complete pseudocode for the sampling and density evaluation algorithm.

\begin{algorithm}
	\caption{Sample from local approximations
	}\label{bfgs-sample.fig}
	\hspace*{\algorithmicindent} \textbf{Input:} \\
	\hspace*{\algorithmicindent} \hspace*{\algorithmicindent} $\samp{s}{1:l}$: the updates of position upto iteration $l$\\
	\hspace*{\algorithmicindent} \hspace*{\algorithmicindent} $\samp{z}{1:l}$: the updates of gradient ($-\nabla \log p$) upto iteration $l$\\
	\hspace*{\algorithmicindent} \hspace*{\algorithmicindent} $\samp{\theta}{l}$: the position of optimization path at iteration $l$\\
	\hspace*{\algorithmicindent} \hspace*{\algorithmicindent} $\nabla \log p(\samp{\theta}{l}$): the gradient of log density at $\samp{\theta}{l}$ \\
	\hspace*{\algorithmicindent} \hspace*{\algorithmicindent} $\samp{\alpha}{l}$: The diagonal elements of the initial inverse Hessian approximation \\
	\hspace*{\algorithmicindent} \hspace*{\algorithmicindent} $\samp{\xi}{1:l}$:  The indicator of whether the update of the position and gradient are \\
	\hspace*{\algorithmicindent} \hspace*{\algorithmicindent} \hspace*{\algorithmicindent} included in the inverse-Hessian approximation or not.\\
	\hspace*{\algorithmicindent} \hspace*{\algorithmicindent} $J$: 
	size of the history used to approximate the inverse Hessian \\
	\hspace*{\algorithmicindent} \hspace*{\algorithmicindent} $M$: number of draws to return \\
	\hspace*{\algorithmicindent} \textbf{Output:}\\
	\hspace*{\algorithmicindent} \hspace*{\algorithmicindent} $\samp{\phi}{1}, \ldots, \samp{\phi}{M}$: draws from approximate distribution
	($M \times N$ matrix) \\
	\hspace*{\algorithmicindent} \hspace*{\algorithmicindent} $\log q(\samp{\phi}{1}), \ldots, \log q(\samp{\phi}{M})$: log densities of
	draws in the approximate normal distribution \hspace*{\algorithmicindent} \hspace*{\algorithmicindent} \hspace*{\algorithmicindent} ($M$-vector) 
	\begin{algorithmic}[1]
		\Procedure{BFGS-sample}{$\samp{s}{1:l}, \samp{z}{1:l}, \samp{\theta}{l},  \nabla \log p(\samp{\theta}{l}), \samp{\alpha}{l}, \samp{\xi}{1:l}, M$}
		\State find the indexes $\chi$ of the last (at most) $J$ non-zero indicators in $\samp{\xi}{1:l}$ and record the last $J$ \hspace*{\algorithmicindent}updates of positions  $S = \samp{s}{\chi} = \begin{bmatrix} S_1 & \cdots & S_J\end{bmatrix}$ and gradients $Z = \samp{z}{\chi} = \begin{bmatrix} Z_1 & \cdots & Z_J \end{bmatrix}$. \hspace*{\algorithmicindent}The latest update is on the last column
		\State generate the upper-triangular matrix $E$ by \hfill $\bigo{J^2 N}$
		$$E_{i, j} = \begin{cases}
			S_i^{\top} \cdot Z_{j} & \textrm{if} \ i \leq j
			\\
			0 & \textrm{otherwise}
		\end{cases},\; \textrm{and save the diagonal elements of $E$ as $\eta$}
		$$ 
		\State generate $\beta, \gamma$ by \hfill $\bigo{J^2  N + J^3}$ 
		$$\beta = \begin{bmatrix} 
			\textrm{diag}(\samp{\alpha}{l}) \cdot Z & S
		\end{bmatrix}\;,\;
		\gamma = \begin{bmatrix}
			0 & -\,E^{-1}
			\\[2pt] 
			-\, E^{-\top} & E^{-\top} \cdot \left(\textrm{diag}(\eta) + Z^{\top} \cdot \textrm{diag}(\samp{\alpha}{l}) \cdot Z\right) \cdot E^{-1}
		\end{bmatrix}$$
		\State compute the thin QR-factorization ${Q}$ and ${\tilde{R}}$ for $\textrm{diag}({\alpha}^{-\frac{1}{2}}) \cdot \beta$ \hfill $\bigo{J^2  N}$
		\State calculate the Cholesky decomposition ${\tilde{L}}$ of $I + {\tilde{R}} \cdot \gamma \cdot {\tilde{R}}^\top$ \hfill $\bigo{J^3}$
		\State let $\log \left|\Sigma\right| =  \log\left|\textrm{diag}(\alpha)\right| + 2\log\left|{\tilde{L}}\right|$ \hfill $\bigo{N}$
		\State let $\mu = \theta + \textrm{diag}(\alpha) \cdot \nabla \log p(\theta) + \beta \cdot \gamma \cdot \beta^\top \cdot \nabla \log p(\theta)$ \hfill $\bigo{JN + J^2}$
		\For{$m \in 1:M$} \hfill $\bigo{J N M + J^2 M}$
		\State sample $\samp{u}{m} \sim \textrm{multi-normal}(0, \idenI)$ \hfill $\bigo{N}$
		\State let $\samp{\phi}{m} = \mu + \textrm{diag}(\alpha^{\frac{1}{2}}) \{{Q} \cdot ({\tilde{L}} - I) \cdot ({Q}^{\top} \cdot \samp{u}{m}) + \samp{u}{m} \}$
		\hfill $\bigo{J  N + J^2}$
		\State let $\log q(\samp{\phi}{m}) = -\frac{1}{2}\left(\log|\Sigma| + u^{(m)\top} \cdot \samp{u}{m} + N\log(2\pi)\right)$ \hfill $\bigo{N}$
		\EndFor
		\EndProcedure
	\end{algorithmic}
\end{algorithm}


\subsection{Estimating KL divergence from the approximate densities}\label{subsec: est_DIV}

From the sequence $\textrm{multi-normal}(\samp{\mu}{1}, \samp{\Sigma}{1}), \ldots,
\textrm{multi-normal}(\samp{\mu}{L}, \samp{\Sigma}{L})$ of normal approximations along the optimization path, Pathfinder then selects the approximation from step $l^*$ that minimizes Kullback-Leibler divergence to the target density,
\[
l^* = \textrm{arg min}_l \
\textrm{KL}\big[
\textrm{multi-normal}(\theta \mid \samp{\mu}{l}, \samp{\Sigma}{l})
\ \big|\big| \
p(\theta \mid y)
\big].
\]
As usual in variational inference, it is more convenient to define $l^*$ equivalently as the point that maximizes the evidence lower bound (ELBO) 
\citep{wainwright2008graphical}.
With draws from the approximating distribution,
$\samp{\phi}{1}, \ldots, \samp{\phi}{K}
\sim \textrm{multi-normal}(\samp{\mu}{l}, \samp{\Sigma}{l}),
$
the ELBO is straightforward to evaluate with Monte Carlo,
\begin{equation}\label{eq: ELBO_est}
	\begin{array}{l}
		\mbox{ELBO}\big[\textrm{multi-normal}(\samp{\mu}{l}, \samp{\Sigma}{l})
		\, \big|\big| \, p(\theta \mid y) \big]
		\\
		\qquad \approx
		\displaystyle
		\frac{1}{K} \sum_{k = 1}^{K}
		\log p(\samp{\phi}{k})
		- \log (\textrm{multi-normal}(\samp{\phi}{k} \mid \samp{\mu}{l}, \samp{\Sigma}{l})).
	\end{array}
\end{equation}
The pseudocode for the ELBO estimation algorithm is provided in Algorithm~\ref{ELBO-est.fig} in Appendix~\ref{appendix: pseudocode}.

Given the factorization of the L-BFGS covariance described in the previous section, the ELBO can be approximated using $K$ draws from the approximating distribution in $\bigo{N  J^2 + J^3 + J N K + J^2 K}$
operations using only $\mathcal{O}(N J + J^2)$ memory, where $J$ is the history size of L-BFGS, $N$ is the number of dimensions, and $K$ is the number of Monte Carlo draws used to evaluate the ELBO.  Both the history size ($J$) and number of Monte Carlo evaluations ($K$) will be small and fixed, rendering the overall complexity linear in the dimensionality of the target distribution ($N$).

Using a larger number $K$ of Monte Carlo draws will reduce the variance of the ELBO estimate at the cost of more computation.  The ELBO tends to be more stable than other divergence measures when using a finite sample size $K$ \citep{dhaka2021challenges}. We have chosen $K = 5$ in our experiments in Section~\ref{sec: simulation}.  Furthermore, the $K$ samples can be drawn and evaluated for log density in parallel with no synchronization required until they are averaged to produce a final estimate.   We evaluate the sensitivity of our results to the choice of $K$ in Section~\ref{sec: simulation}.

\subsection{Pareto-smoothed importance resampling}\label{subsec: PS-IR}

\begin{algorithm}
	\caption{Pareto-smoothed importance resampling (PS-IR)}\label{ps-sir.fig}
	\hspace*{\algorithmicindent} \textbf{Input:} \\
	\hspace*{\algorithmicindent} \hspace*{\algorithmicindent}
	$\samp{\phi}{1}, \ldots, \samp{\phi}{J}$: draws from proposal distribution $q$ \\
	\hspace*{\algorithmicindent} \hspace*{\algorithmicindent} $\log q(\samp{\phi}{1}), \ldots, \log q(\samp{\phi}{J})$: proposal log densities \\
	\hspace*{\algorithmicindent} \hspace*{\algorithmicindent} $\log p(\samp{\phi}{1}), \ldots, \log p(\samp{\phi}{J})$: target log densities \\
	\hspace*{\algorithmicindent} \hspace*{\algorithmicindent} $R$: number of draws resampled ($R \leq J$) \\
	\hspace*{\algorithmicindent} \textbf{Output:} \\
	\hspace*{\algorithmicindent} \hspace*{\algorithmicindent}
	$\samp{\psi}{1}, \ldots, \samp{\psi}{R}$: importance resampled draws
	\begin{algorithmic}[1]
		\Procedure{PS-IR}{$\samp{\phi}{1:J}, \log q(\samp{\phi}{1:J}), \log p(\samp{\phi}{1:J}), R$}
		\State let $w_1, \ldots, w_{\!J} = \textrm{PSIS}(\log q(\samp{\phi}{1:J}), \log p(\samp{\phi}{1:J}))$ be the Pareto-smoothed \newline importance sampling weights
		\hfill $\bigo{J}$
		\State sample $\samp{\psi}{1}, \ldots, \samp{\psi}{R}$ from $\samp{\phi}{1}, \ldots, \samp{\phi}{J}$ with replacement, with probabilities \newline proportional to $w_{\!j}$
		\hfill $\bigo{R J}$
		\EndProcedure
	\end{algorithmic}
\end{algorithm}

In the final step of multi-path Pathfinder, we employ a Pareto-smoothed importance resampling algorithm to refine the approximation draws based on approximations from $I$ independent runs of Pathfinder.  
Importance resampling is a method for refining a set of draws from an approximating distribution to better approximate draws from a target distribution \citep{rubin1987calculation}.  Importance resampling works by resampling from the original sample with replacement with probabilities proportional to the importance weights. 
Importance sampling estimators weight draws based on their importance ratios but can have high or even infinite variance. \citet{ionides2008truncated} showed that truncating the importance weights improves the efficiency of the resulting Monte Carlo estimator by reducing its variance.  \citet{vehtari2019pareto} introduced a continuous generalization of truncation that fits the importance weights to a generalized Pareto distribution, whose cumulative distribution function is then used to provide an evenly spaced set of importance weights.  Rather than directly using the smoothed weights to calculate expectations, we instead use them as importance resampling weights, leading to Pareto smoothed importance resampling (PS-IR), as listed in Algorithm~\ref{ps-sir.fig}. 
As far as we know, Pareto smoothing has not been previously applied to importance resampling, but only to importance sampling.  We use resampling in order to make it easy to use as an initialization algorithm for MCMC and to simplify expectation and quantile estimation by returning draws rather than weighted draws.


\subsection{Related methods}\label{sec:related}

\textit{Automatic differentiation variational inference} (ADVI) is a method for black-box variational inference with differentiable densities \citep{kucukelbir2017automatic}.  ADVI's variational objective is identical to Pathfinder's, namely Kullback-Leibler (KL) divergence from the approximating distribution to the target distribution.  The difference is that ADVI directly optimizes the variational objective using stochastic gradient descent, whereas Pathfinder optimizes the target density using quasi-Newton optimization and then chooses the point along the optimization path based on the variational objective.

Like Pathfinder, ADVI uses a multivariate normal approximating distribution on an unconstrained parameter space.  Any constrained variables such as scales or covariance matrices or simplexes are transformed to an unconstrained representation in $\mathbb{R}^N$, with appropriate change of variables adjustments.  ADVI's covariance matrix may be taken to be dense or it may be constrained to be diagonal;  \citet{kucukelbir2017automatic} call the former ``full rank'' and the latter ``mean field,'' though both are technically required to have rank $N$.

The Kullback-Leibler (KL) divergence from the normal approximation to the target distribution is evaluated using Monte Carlo methods by taking an average of the log density of draws from the approximating distribution.  This results in a stochastic gradient algorithm, with gradients calculated using automatic differentiation \citep{mohamed2019monte, carpenter2015autodiff}. \citet{dhaka2021challenges} show that Monte Carlo estimates of this KL divergence and its gradient are in general stable, although they may have high variance.

Compared to Pathfinder's direct quasi-Newton optimization of the log density, ADVI is restricted to small step sizes because of the stochastic nature of the gradient calculation and the lack of curvature information.  In evaluations below, we show that ADVI requires one to two orders of magnitude more function evaluations than Pathfinder to find the high probability mass region.  In addition, we show that Pathfinder produces approximations that range from slightly worse to much better than ADVI (with limited computation time) in complex problems as measured by the 1-Wasserstein metric.  Finally, ADVI is intrinsically serial in its evaluation of the KL divergence at each iteration of optimization, whereas Pathfinder is embarrassingly parallel after the relatively fast L-BFGS optimization.

\textit{Early stopping optimization} divides the data in two parts, with one part used for optimization and other part is used to compute out-of-sample performance criterion and the optimization is stopped when the out-of-sample performance starts to decrease \citep{vehtari2000mcmc}. The downside of the approach is that it requires factorizing the likelihood and additional data manipulation to make the data divisions. Pathfinder works also for non-factorized likelihoods as the ``stopping'' is decided by the ELBO estimate (in our implementation the optimization is not stopped early but run to the termination and then ELBO is estimated for each optimization trajectory point, potentially in parallel). Furthermore Pathfinder returns normal approximations instead of just points along the trajectory.

\textit{Early stopping variational inference} generates approximate posterior draws by taking random draws from an initialization distribution, then following an optimization path for a fixed number of steps and taking the result as an approximate posterior draw \citep{duvenaud2016early}.  The number of steps is selected to minimize the KL divergence from the approximating distribution to the target distribution. 
Early stopping variational inference can be viewed as a normalizing flow \citep{rezende2015variational}, which generates an approximate draw from a target distribution by generating a draw from a simple distribution, such as uniform or standard normal, then transforming it. Pathfinder is similar, but it stops at a normal approximation from which we draw a sample. 

Early stopping VI differs from Pathfinder in several substantive ways.  Most importantly, early stopping VI determines a number of optimization steps as the variational parameter, generating a range of values based on the random initialization.  Pathfinder, in contrast, evaluates a variational approximation centered at each point on the optimization path as a variational approximation.  Secondarily, early stopping VI chooses a point on the optimization path, whereas Pathfinder generates a point from a normal approximation located at a point on the optimization path. 
Computationally, early stopping variational inference requires an optimization run for each approximate posterior draw, whereas Pathfinder uses a single optimization run (though multi-path Pathfinder importance resamples among several such paths).

\textit{Invertible flow non-equilibrium sampling} (InFiNE) is another method based on selecting points on a deterministic optimization path with importance resampling \citep{thin2021invertible}.  Unlike Pathfinder and the other systems mentioned so far, InFiNE is asymptotically exact.  It was motivated by the need for efficient estimators for normalizing constants of intractable densities known only up to a constant factor, such as most Bayesian posteriors.  The other motivation mentioned in the paper is in accurately evaluating KL divergence using the evidence lower bound (ELBO).  To achieve these goals, InFiNE uses an iterative importance resampling scheme \citep{andrieu2010particle}. 

As with normalizing flows, InFiNE keeps the Jacobian tractable over the optimization path by using a Hamiltonian flow with a friction term that will cause the flow to come to rest at a local mode (a well in potential energy, which is negative log density) or just keep falling.

\textit{Short parallel MCMC chains.}
\citet{hoffman2020black} demonstrate that HMC can quickly reach the high probability region, which they exploit by generating many short MCMC chains in parallel and using only the small part from the end. In experiments not reported here, we tested starting many parallel chains from the L-BFGS trajectory points and testing for a trend in the log density. A lack of trend in the log density is consistent with the chains having been initialized in the high probability region. Although we have verified this approach works by evaluating short-chain dynamic HMC, the number of log density evaluations required is several orders of magnitude larger than needed by the ELBO estimate used in Pathfinder.


\section{Experiments}\label{sec: simulation}
This section provides experimental evaluations of Pathfinder.  In Section~\ref{subsec: sim_pf_ADVI}, we compare Pathfinder to two popular posterior approximation algorithms, ADVI and an ensemble of short adaptive HMC chains.  ADVI \citep{kucukelbir2017automatic} is the industry standard black-box variational inference algorithm, implemented in Stan \citep{stan2021ref}, PyMC3 \citep{salvatier2016probabilistic}, Pyro \citep{bingham2019pyro}, TensorFlow Probability \citep{dillon2017tensorflow}, JAX \citep{jax2018github}, Turing.jl \citep{ge2018t}, and other differentiable programming languages.  Like Pathfinder, ADVI provides normal approximations of posteriors.  ADVI can be configured to use a dense covariance or restricted to a diagonal covariance matrix.  We evaluate both alternatives in this section.  We treat Stan's no-U-turn sampler \citep{hoffman2014no,betancourt2017conceptual} as a nonparametric posterior approximation algorithm, and, following the conclusions in \citet{hoffman2020black}, we ran many parallel MCMC chains and took samples from the last iteration as approximate draws.  
We evaluate Pathfinder's sensitivity to tuning parameters in Section~\ref{subsec: sim_sens}. In Section~\ref{subsec: case_study_pf}, we investigate the behavior of Pathfinder for difficult posteriors. We evaluate the results of using Pathfinder versus short chains of adaptive HMC for initializing a Gaussian process model in Section~\ref{sec: Birthday}. The code for simulations is available at \url{https://github.com/LuZhangstat/Pathfinder}.



We use 1-Wasserstein distance \citep{Craig_2016, Villani_2009, McCann_1995} between the empirical distribution of the approximate samples and the target posterior distribution to evaluate how well we are taking independent draws from the posterior. We provide an introduction to 1-Wasserstein distance and its computation in our simulation studies in Appendix~\ref{appendix: 1_wasserstein}. In all of the evaluations presented in this section, the model parameters are transformed to the unconstrained scale, with corresponding change of variables adjustments.  The resulting support on all of $\mathbb{R}^N$ matches the support of the multivariate normal approximations used by ADVI and Pathfinder and allows the algorithms to avoid dealing with boundaries. Moreover, we assume that the posterior distribution is closer to normal in the unconstrained space. Hence, for both ADVI and Pathfinder, the Gaussian approximation is made in the unconstrained space. And in all of our experiments, we compare the approximation performance through samples in the unconstrained space. In practice, these samples will be (inverse) transformed back to the constrained space. Details of the transformations, inverse transforms, and their log absolute Jacobian determinants are provided by \citet[Chapter~10]{stan2021ref}.

\subsection{Evaluating Pathfinder as variational inference}\label{subsec: sim_pf_ADVI}

In this section, we compare Pathfinder with ADVI and Stan's default phase I warmup (dynamic HMC in the form of the no-U-turn sampler) through experiments. We evaluate Pathfinder using the 20 models and data sets from \texttt{posteriordb} \citep{mans2021posteriordb}, each of which is supplied with reference posteriors in the form of 10,000 roughly independent draws. The set of models evaluated includes
\begin{itemize}
	\item generalized linear models: \verb|nes|, 
	\verb|earnings|, \verb|dogs|, \verb|diamonds| \verb|sblrc|,
	\item hierarchical meta-analysis models: \verb|eight_schools| (centered and non-centered),
	\item Gaussian processes: \verb|gp_pois|,
	\item mixtures: \verb|low_dim_gauss_mix|,
	\item differential equation dynamics models: \verb|hudson_lynx_hare|, \verb|one_comp_MM_elim|,
	\item hidden Markov models: \verb|bball_drive|, and
	\item time-series models: \verb|arma|, \verb|arK|, and \verb|garch|.
\end{itemize}
For each model in \texttt{posteriordb}, we run single-path Pathfinder with 100 different random initializations, using our proposed default settings:
\begin{itemize}
	\item maximum L-BFGS iterations ($\rmsup{L}{max} = 1000$),
	\item relative tolerance for L-BFGS convergence ($\tau^{rel} = 10^{-13}$),
	\item size of L-BFGS history to approximate inverse Hessian ($J = 6$),
	\item number of Monte Carlo draws to evaluate ELBO ($K = 5$), and
	\item number of draws per run ($M = 100$).
\end{itemize}
For multi-path Pathfinder, we again take 100 approximate draws, but use a larger number of intermediate runs,
\begin{itemize}
	\item number of single-path Pathfinder runs ($I = 20$),
	\item number of draws returned by each single-path Pathfinder run ($M = 100$), and
	\item number of draws per run ($R = 100$).
\end{itemize}
For both single-path and multi-path Pathfinder, we repeat the entire process 100 times. We use Wasserstein distance from the approximate draws in each run to the reference posterior draws to determine how well Pathfinder achieves its goal of producing approximate posterior samples.

In addition to single-path and multi-path Pathfinder, we also evaluate 100 runs by 
\begin{itemize}
	\item Stan phase I adaptation: adaptive Hamiltonian Monte Carlo with Stan's no-U-turn sampler (unit metric, step size adaptation, and a maximum tree depth of 10, keeping the last of 75 iterations),
	\item dense ADVI:  automatic differentiation variational inference with a dense covariance matrix (Stan default settings, return 100 approximate draws), and
	\item mean-field ADVI: automatic differentiation variational inference with a diagonal covariance matrix (Stan default settings, return 100 approximate draws).
\end{itemize}
Each of these procedures uses random
initialization values, generated from a $\textrm{uniform}(-2, 2)$ distribution, which is the default for Stan. 


The expressive power of Pathfinder's low-rank plus diagonal covariance approximation is between that of ADVI's diagonal and dense choices for covariance. In summary, we generated 100 runs of 100 draws for each of our candidate approximate posterior distribution algorithms, including Pathfinder, multi-path Pathfinder, mean-field ADVI and dense ADVI, and we generated 100 approximate samples using Stan's phase I sampler.

\paragraph{Assessing the quality of approximations.}

\begin{figure}[t!]
	\centering{
		\includegraphics[width=0.9\textwidth, keepaspectratio]{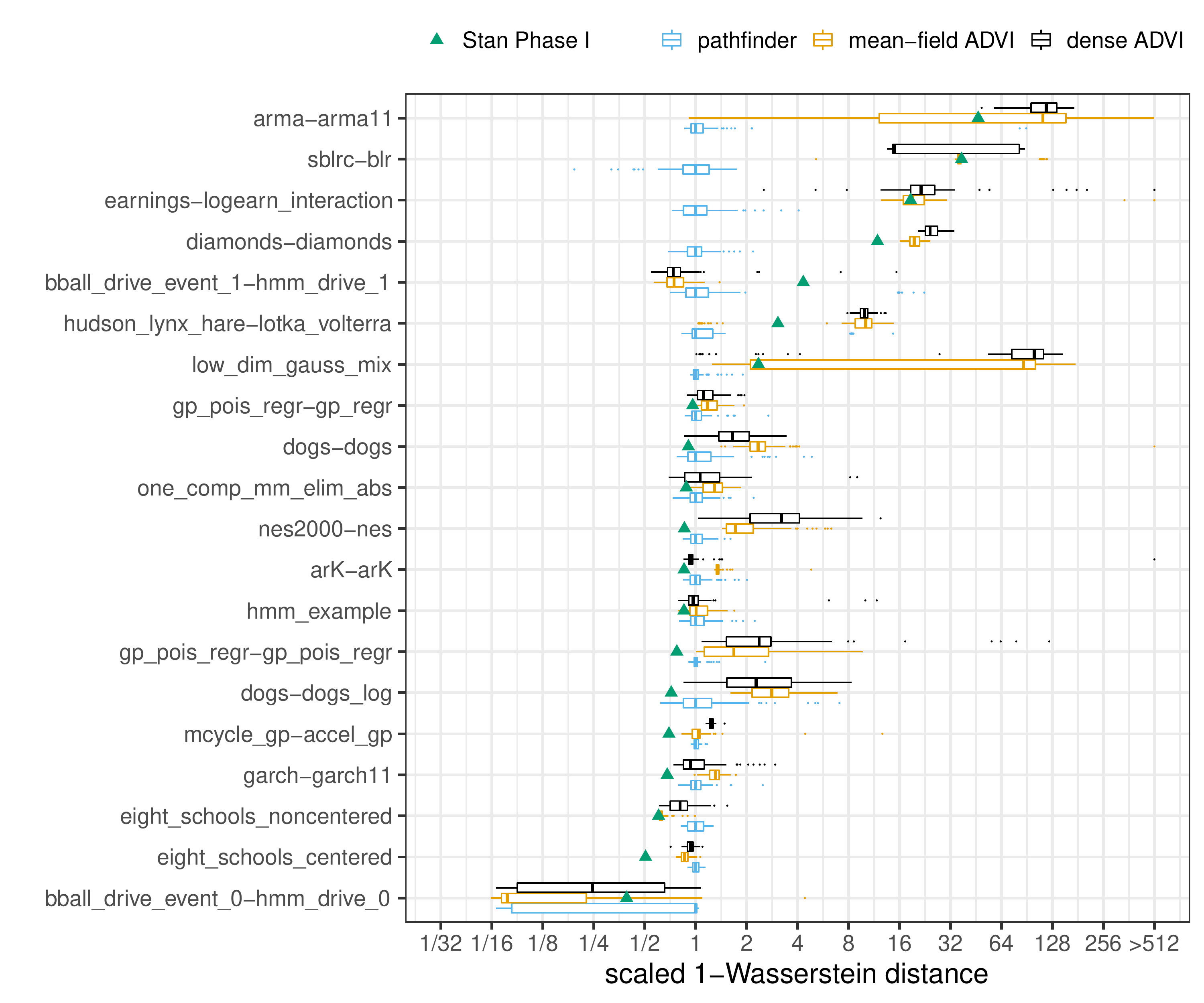}}
	\caption{\it Box plots of 1-Wasserstein distances between the reference posterior samples and approximate draws from single-path Pathfinder and ADVI for the 20 models in \texttt{posteriordb}. Each box plot displays 1-Wasserstein distances of 100 independent runs of (single-path) Pathfinder, mean-field ADVI, and dense ADVI. We calculate the 1-Wasserstein distance with 100 approximate draws from the last iteration of 100 runs of Stan's phase I warmup (adaptive HMC). Distances for each model are scaled by the median of the 1-Wasserstein distances for single-path Pathfinder.}\label{fig: pfvsADVI_compar}
\end{figure}

\begin{figure}[t!]
	\centering{
		\includegraphics[width=.9\textwidth, keepaspectratio]{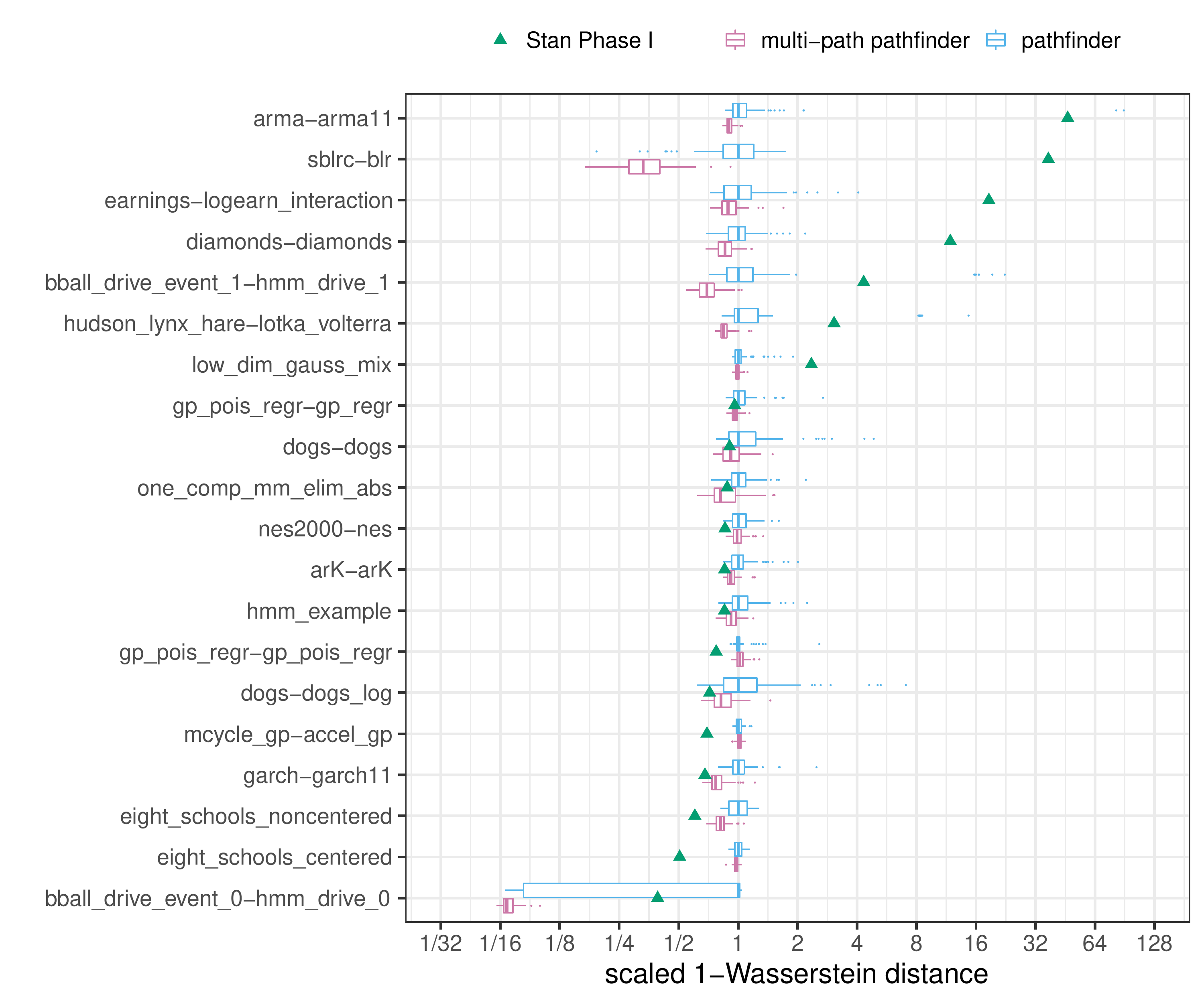}}
	\caption{\it Box plots of 1-Wasserstein distances between the reference posterior samples and approximate draws from single-path Pathfinder and multi-path Pathfinder for the 20 models in \texttt{posteriordb}. Each box plot displays 1-Wasserstein distances of 100 independent runs of single-path or multi-path Pathfinder. We calculate 1-Wasserstein distance with 100 approximate draws from the last iteration of 100 runs of Stan's phase I warmup (NUTS). Distances for each model are scaled by the median of the 1-Wasserstein distances for single-path Pathfinder. }\label{fig: single_vs_multi_pf}
\end{figure}

We provide a comparison of single-path Pathfinder, ADVI and Stan's phase I sampler through 1-Wasserstein distances for all 20 models from \texttt{posteriordb} in Figure~\ref{fig: pfvsADVI_compar}. A comparison of multi-path Pathfinder, single-path Pathfinder and Stan's phase I sampler is in Figure~\ref{fig: single_vs_multi_pf}. 
To adjust for the varying scale of the 1-Wasserstein distances across target densities, we scaled results relative to the median of the 100 1-Wasserstein distances for single-path Pathfinder for each model. This allows us to compare ratios of the 1-Wasserstein distance between Pathfinder's draws and the target posterior and the 1-Wasserstein distance between another system's draws and the target posterior. 

It is clear that single-path and multi-path Pathfinder outperform the ADVI variants for most of the models in \texttt{posteriordb}.  The bar and whisker plots show that over 100 independent runs, multi-path Pathfinder is the most stable, followed by single-path Pathfinder, and then mean-field ADVI, with dense ADVI providing the most variability in 1-Wasserstein distance to the true posterior. The median 1-Wasserstein distance for mean-field ADVI is more than double that of single-path Pathfinder for 8 (of 20) test models. Dense ADVI is worse, with 9 (of 20) test models having double the 1-Wasserstein distance of single-path Pathfinder. 

There is only one model where the 1-Wasserstein distance is much smaller for mean-field ADVI, the hidden Markov model \verb|bball_drive_event_0-hmm_drive_0|, with a median 1-Wasserstein distance that less than one tenth of that for single-path Pathfinder. This particular model has multiple meaningful posterior modes. The noise inherent in the stochastic gradient descent approach used by ADVI allows it to escape minor modes than can trap the L-BFGS optimizer used by Pathfinder. It might be possible to resolve this problem for single-path Pathfinder with a more robust optimization algorithm. Until we find such an algorithm, we note that multi-path Pathfinder eliminates this problem for the \verb|bball_drive_event_0-hmm_drive_0| model, resulting in 1-Wasserstein distances that are comparable to those for mean-field ADVI.

In addition to working better on most posteriors, single-path and multi-path Pathfinder are more stable than Stan's HMC-based phase I adaptation for more challenging posteriors. For 7 (of 20) test models, Stan's phase I warmup produced 1-Wasserstein distances more than double the median distance of single-path Pathfinder.   Except for the \verb|bball_drive_event_0-hmm_drive_0| example, the 1-Wasserstein distances for single-path Pathfinder are at most double that of Stan's phase I warmup.

\paragraph{Assessing the computational cost.}

\begin{figure}[t!]
	\begin{center}
		\subfloat[Log density evaluations ]{\includegraphics[width=0.75\textwidth, keepaspectratio]{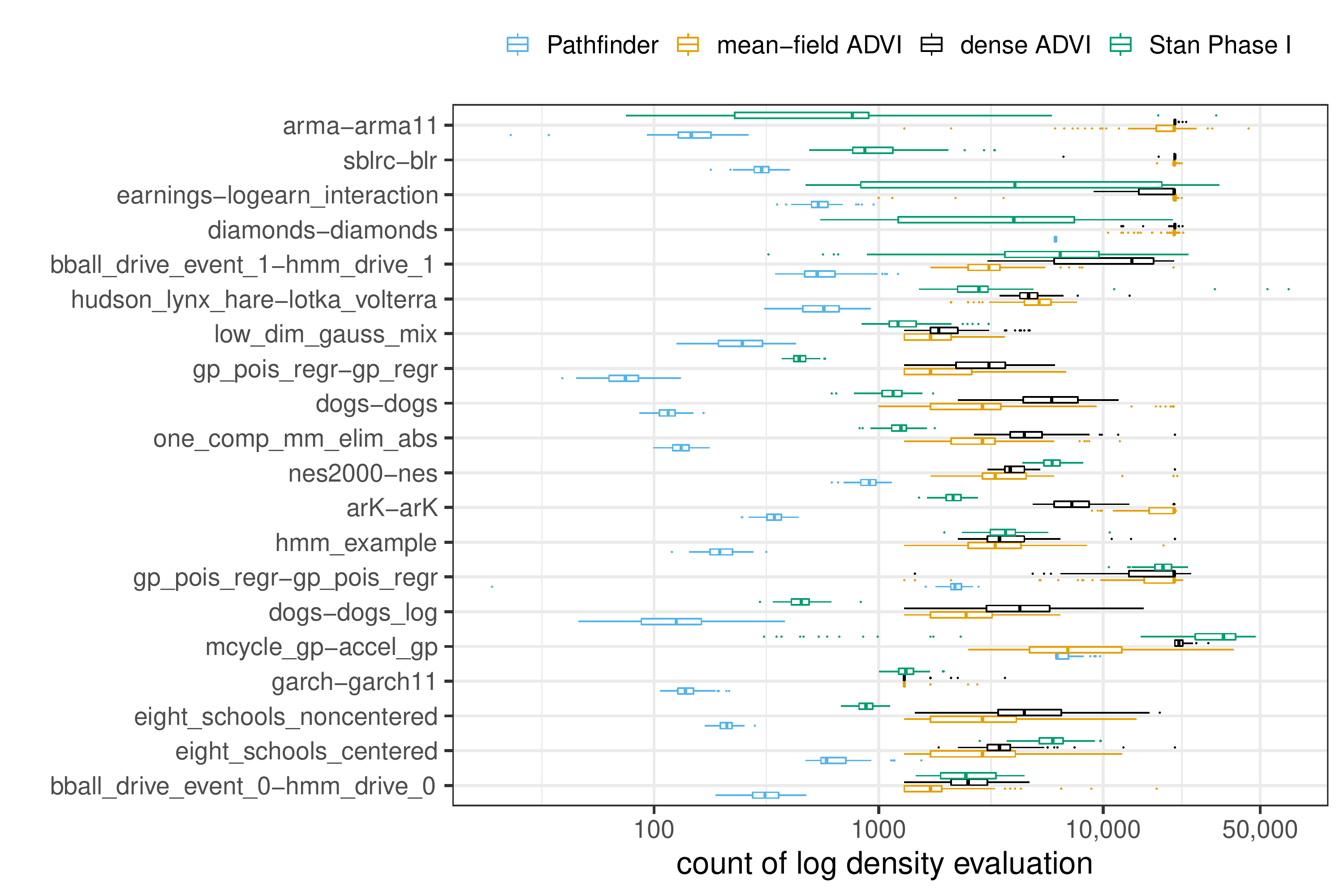}}
		\\
		\subfloat[Gradient evaluations]{\includegraphics[width=0.75\textwidth, keepaspectratio]{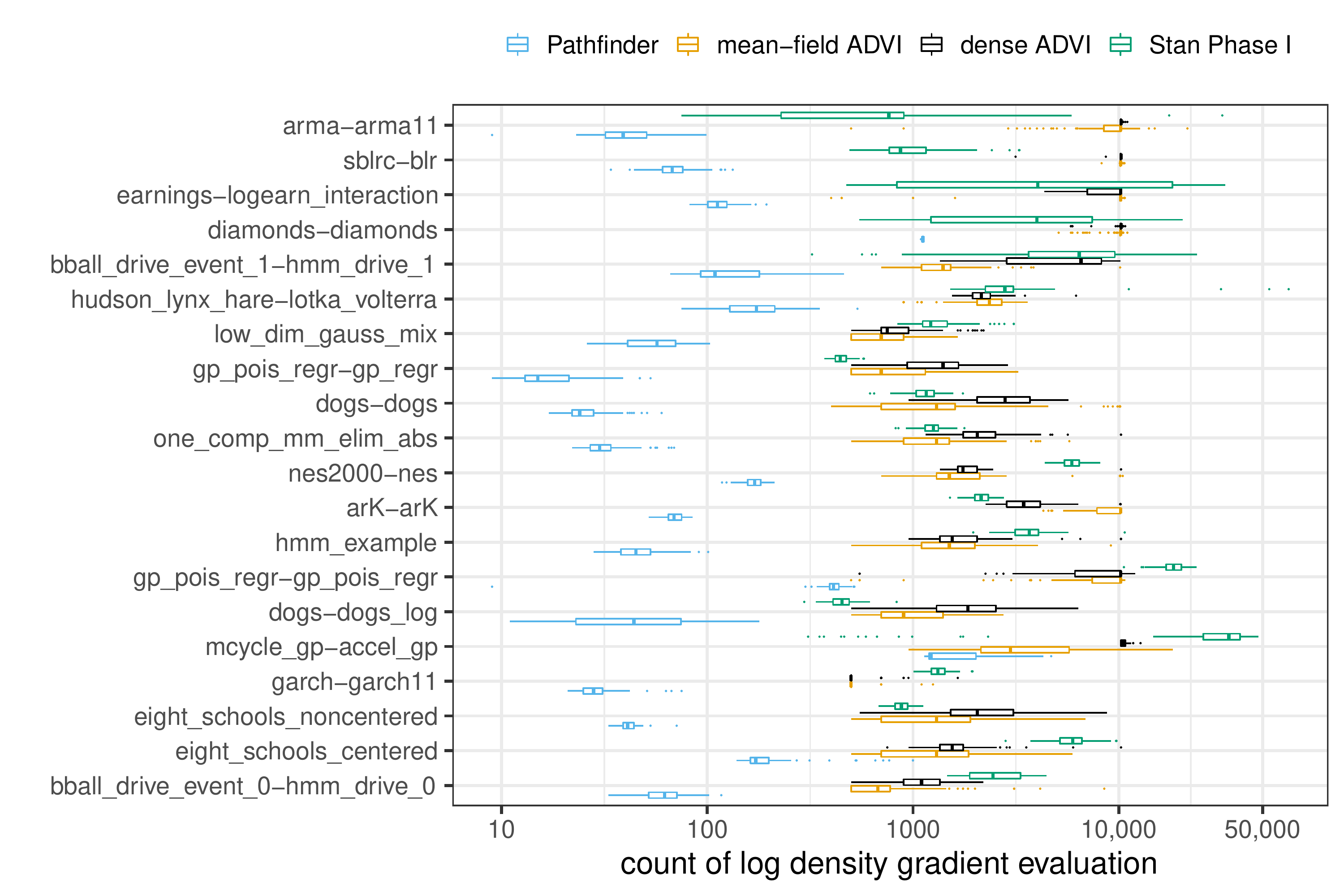}}
		\caption{\it Box plots of the number of (a) log density evaluations and (b) gradient evaluations required by the candidate algorithms.  For each algorithm, we performed 100 independent runs and summarize the results with box plots. Candidate algorithms can abort due to various errors.  We count the log density and gradient evaluations in failed runs.  We do not plot multi-path Pathfinder because its number of evaluations is just a multiple of the single-path results plus a small number of extra approximate and log density evaluations for importance resampling.}\label{fig: cost_lp_compare} 
	\end{center}
\end{figure}
Pathfinder, Stan's phase I sampler, and automatic differentiation variational inference are all dominated computationally by log density and gradient calculations.  Using the number of these operations as a measure of computation conveys the further advantage of being implementation agnostic.  We summarize the cost for 100 repetitions of (single-path) Pathfinder, Stan phase I warmup, and ADVI for each model through box plots in Figure~\ref{fig: cost_lp_compare} (the repetitions are not part of the algorithm, but merely to provide a sense of execution cost variability from run to run).

\begin{table}[t!]
	\begin{center}
		\begin{tabular}{rccc}
			& \multicolumn{3}{c}{Multiple of Pathfinder's evaluations}
			\\
			&  Stan phase I & mean-field ADVI & dense ADVI
			\\
			\hline
			log density & 7.9 & 24 & 28
			\\
			gradient & 34 & 48 & 54
		\end{tabular}
		\caption{\it The values in the table indicate how much more work is required for Stan's phase I warmup and ADVI compared to Pathfinder averaged over all of the test models in \texttt{posteriordb}. The values are ratios of evaluations, so that, for example, mean-field ADVI required 48 times as many gradient evaluations as Pathfinder.  Dense ADVI does additional work beyond gradient evaluations in log density and simulation which are not included. The actual implementation of candidate algorithms can be aborted due to various errors. In this experiment, we count all the log density and its gradient evaluation in the failed trials until success. 
		}\label{cost.table}
	\end{center}
\end{table}
Table~\ref{cost.table} summarizes average costs in terms of log density and gradient operations. Although we summarize both log density and gradient costs for completeness, gradients typically consume closer to 80\% of overall compute cost for Stan phase I and ADVI when calculated with automatic differentiation \citep{carpenter2015autodiff}.  
Because gradients are more expensive with automatic differentiation than log density evaluations, the results indicate that other methods require around 30 to 50 times more operations than Pathfinder. 


These results only consider serial execution.  Multi-path Pathfinder requires 20 (the default number used in the experiments) times as many evaluations as single-path Pathfinder plus the importance resampling step.  Importance resampling is fast, but it does require evaluating a few log densities of each candidate in both the approximating and target density. Thus the wall time for running multi-path Pathfinder could be nearly as fast as the slowest of the runs of single-path Pathfinder.  Compared to short chains of adaptive HMC, there is less variability across runs for Pathfinder, as can be gleaned from the plots in Figure~\ref{fig: cost_lp_compare}.  
These are timings for single runs (averaged over 100 runs).  In practice, we typically initialize multiple Markov chains, so the gap in number of evaluations becomes even wider, though these can be parallelized for all systems.

\subsection{Sensitivity to tuning parameters}\label{subsec: sim_sens}\label{sec: ovarian}

In this section, we provide an analysis of Pathfinder's sensitivity to tuning parameters to evaluate whether we just got lucky with our suggested default settings for Pathfinder.  For both adaptive Hamiltonian Monte Carlo and automatic differentiation variational inference, we use the default settings in Stan, which have proven successful for a broad range of applications \citep{stan2021ref}.

\begin{figure}[t!]
	\centering{
		\includegraphics[width=.9\textwidth, keepaspectratio]{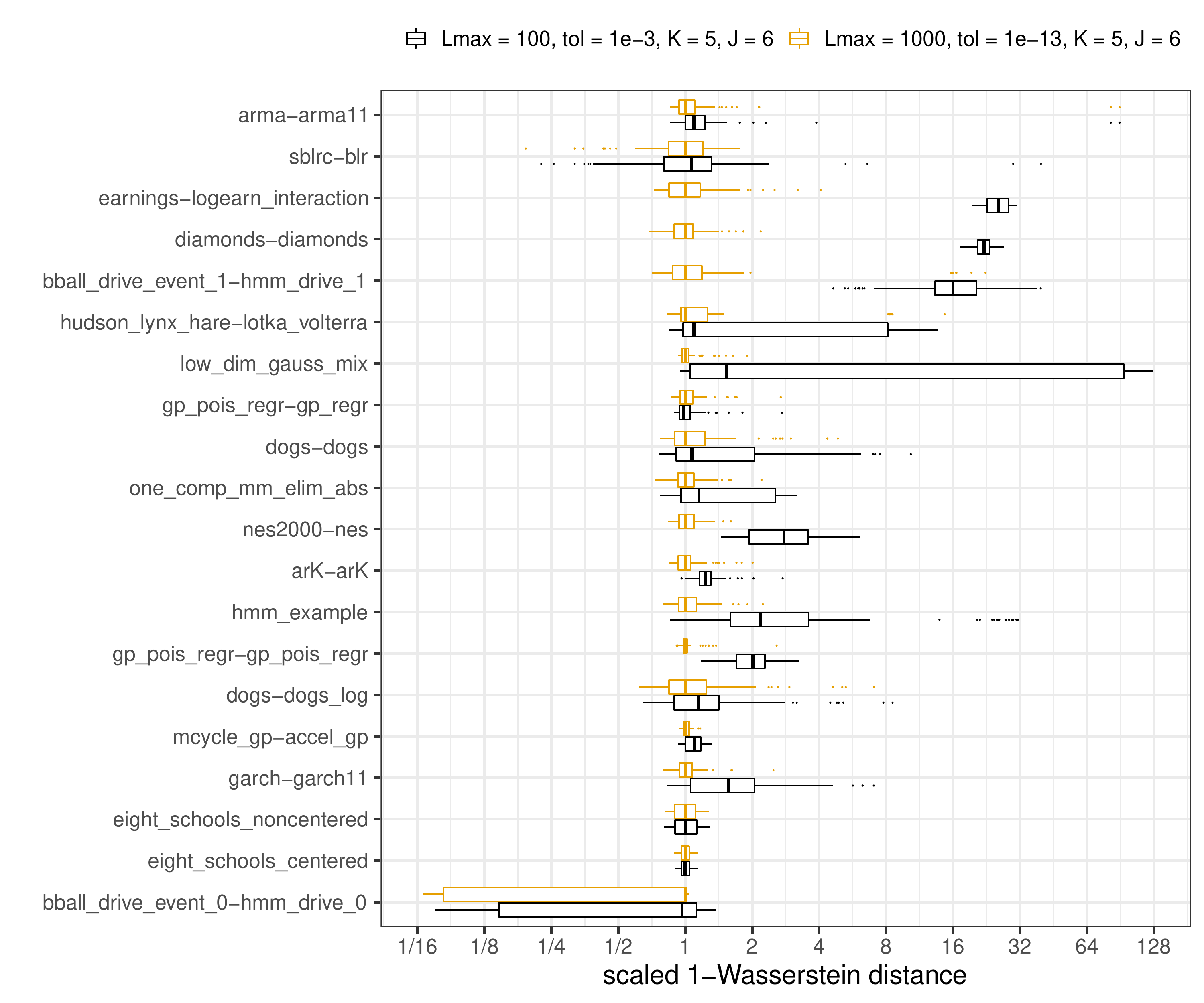}}
	\caption{\it Box plots of 1-Wasserstein distances between the reference posterior samples and approximate draws from Pathfinder for examples in \texttt{posteriordb}.  Each box plot summarizes 1-Wasserstein distances for 100 independent runs of Pathfinder. The two results are for Pathfinder with default settings (orange), and with the number of iterations reduced to $\rmsup{L}{max} = 100$ and the convergence threshold increased to $10^{-3}$ (black). We scaled the 1-Wasserstein distances for the same model by the median distances using the default settings.}\label{fig: sens_test_L}
\end{figure}
\paragraph{Optimization tolerance and maximum number of iterations.} 
To test the performance of Pathfinder under different relative tolerances for convergence $\rmsup{\tau}{rel}$ and maximum iteration for L-BFGS $\rmsup{L}{max}$, we reproduced $100$ runs of Pathfinder with $\rmsup{L}{max}$ reduced to $100$ and $\rmsup{\tau}{rel}$ increased to $10^{-3}$ for each model in \texttt{posteriordb}. The lower cap on number of iterations and more relaxed tolerances should lead to less computation and perhaps less stability in Pathfinder. Following the simulation design in Section~\ref{subsec: sim_pf_ADVI}, we estimate the 1-Wasserstein distance for $100$ approximate draws from each run of Pathfinder. Figure~\ref{fig: sens_test_L} illustrates reduced fidelity and increase in uncertainty with lower $\rmsup{L}{max}$ and higher $\rmsup{\tau}{rel}$ for most of the tested models.  For model \verb|earnings-logearn_interation|, \verb|diamonds-diamond| and \verb|bball_drive_event_1-hmm_drive_1| 1-Wasserstein distances increased more than a factor of 8. None of the models show improved performance under these alternative settings. We thus prefer to keep a larger $L$ and a smaller $\rmsup{\tau}{rel}$, as they do not add much computation for simpler models, for which optimization terminates before the maximum iteration threshold.

\begin{figure}[t!]
	\centering{
		\includegraphics[width=.9\textwidth, keepaspectratio]{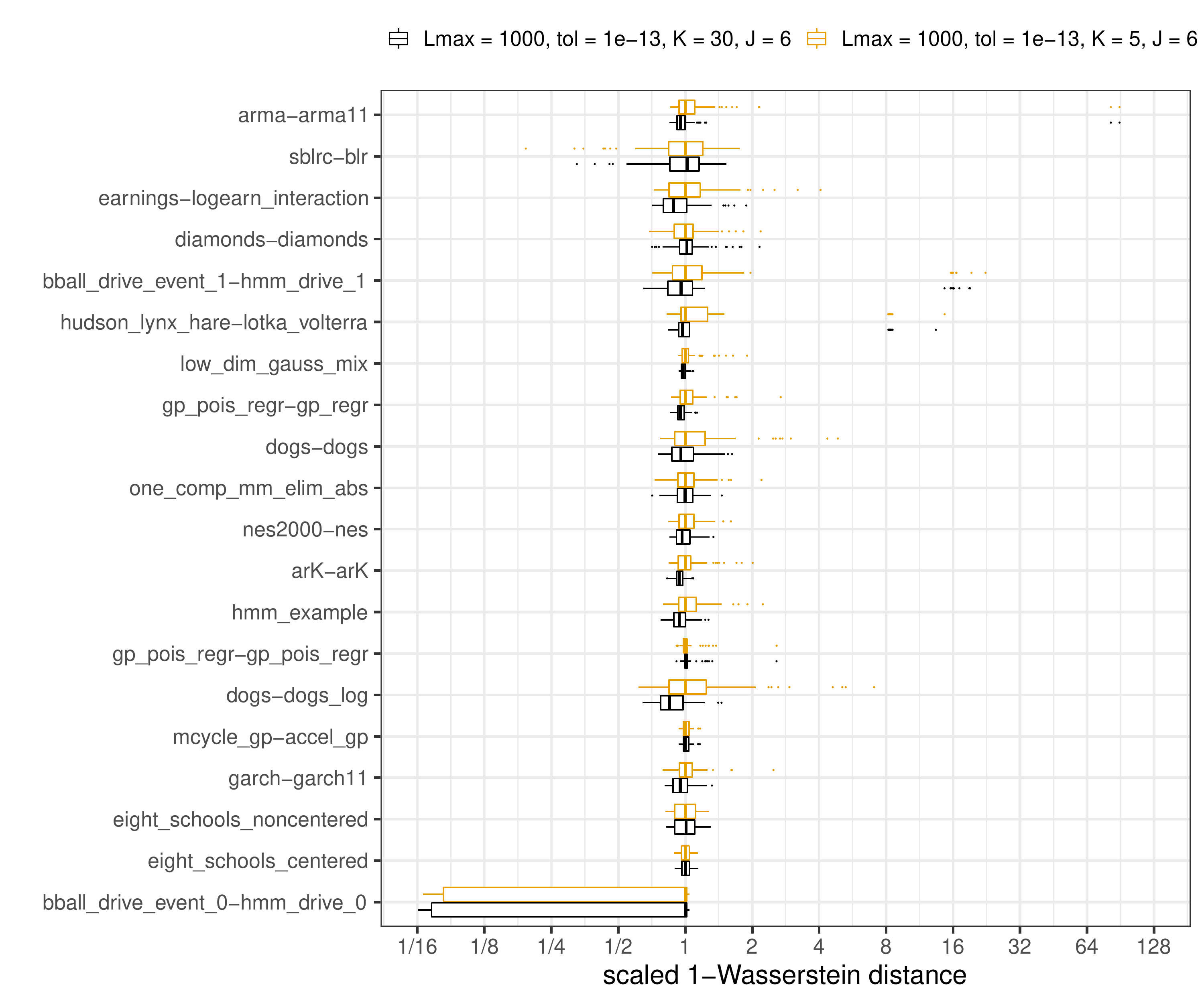}}
	\caption{\it
		Box plots of 1-Wasserstein distances between the reference posterior samples and approximate draws from Pathfinder for examples in \texttt{posteriordb} when using $K=5$ (default, in orange) and $K=30$ (black) Monte Carlo draws to evaluate the evidence lower bound.  The results are scaled by the median of 1-Wasserstein distances for Pathfinder in the default setting.}\label{fig: sens_test_K}
\end{figure}
\paragraph{The number of Monte Carlo draws to estimate ELBO.}
Figure~\ref{fig: sens_test_K} reports our sensitivity test for the number of Monte Carlo draws used to evaluate the evidence lower bound in Pathfinder (tuning parameter $K$).  In particular, it compares 1-Wasserstein distances using the default $K = 5$ draws for evaluating the ELBO with the result of $K = 30$ draws. This increases the number of log density evaluations and the number of random numbers generated, but not the number of gradient evaluations, and reduces the scale of Monte Carlo error by around $(1 - \sqrt{5} / \sqrt{30})$, or 60\%.
Increasing $K$ consistently improves the performance of Pathfinder, but not by much.  The median 1-Wasserstein distances for all models were reduced by only about 3.2\% on average, with a maximum reduction of only 15.1\%.  The cost is about 4.9 times as many log-density evaluations on average.  Based on the tradeoff between the quality of approximation and cost, a smaller $K$ is a better choice when using Pathfinder, especially multi-path Pathfinder, to quickly find a handful of draws close to high probability mass region. On the other hand, when using Pathfinder to generate an approximate posterior, a larger $K$ may be warranted.  Using $K = 30$ also reduces the variability in 1-Wasserstein distance by a little bit, as can be seen in the narrow 50\% intervals and less extreme tail behavior.

\begin{figure}[t!]
	\centering{
		\includegraphics[width=.9\textwidth, keepaspectratio]{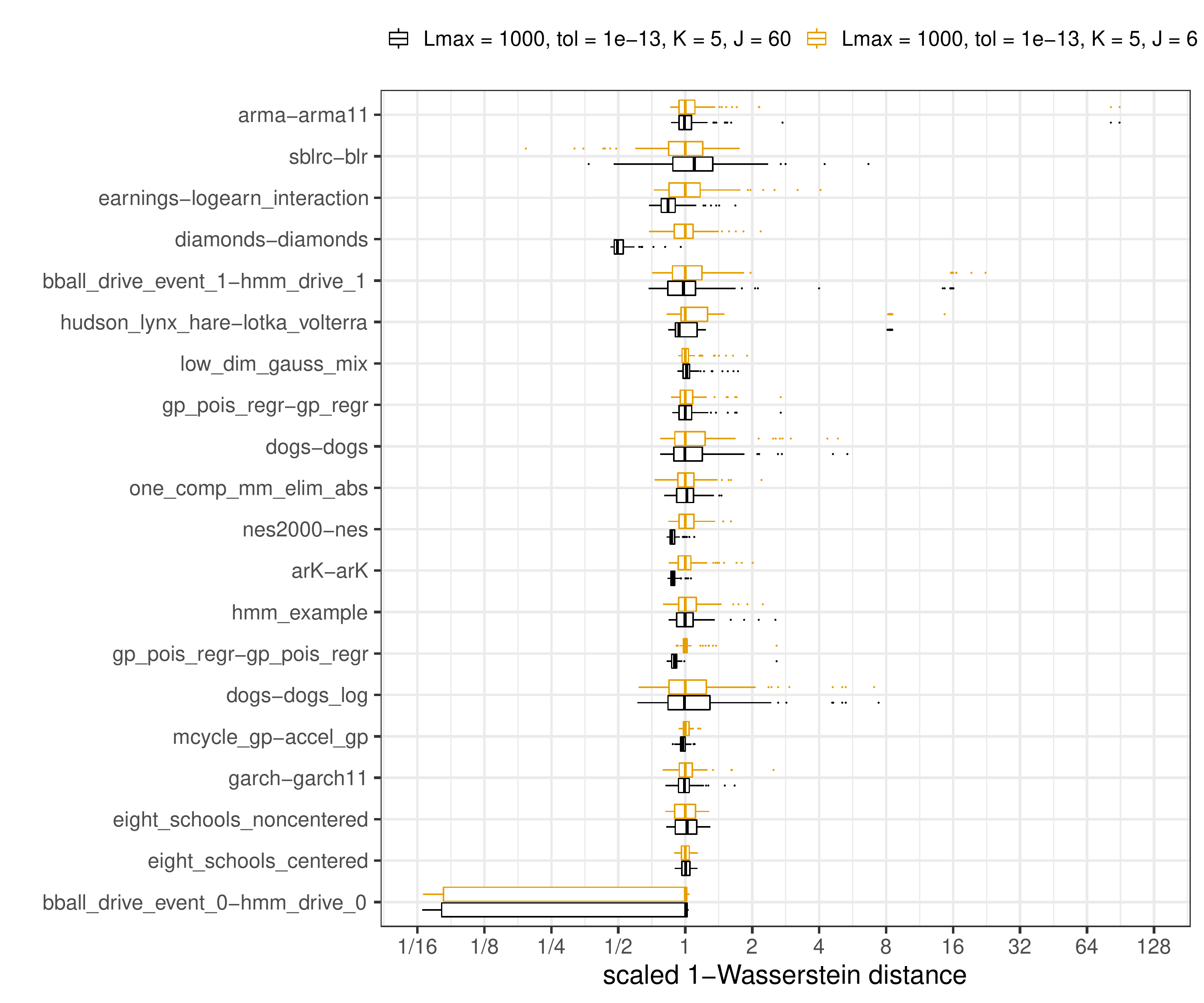}}
	\caption{\it
		Box plots of 1-Wasserstein distances between the reference posterior samples and approximate draws from Pathfinder for examples in \texttt{posteriordb} with the default of $J = 6$ (orange) gradients used to approximate the inverse Hessian and a much larger value of $J = 60$ (black) history elements.  The distances are scaled by the median of the default behavior ($J = 6$).}\label{fig: sens_test_J}
\end{figure}
\paragraph{History size of L-BFGS.}
Figure~\ref{fig: sens_test_J} provides a plot evaluating the difference between the default history size of $J = 6$ for L-BFGS to estimate an inverse Hessian with the much longer history size of $J = 60$.  As before, we report on a comparison of 100 different runs of each system.  Sensitivity to L-BFGS history size ($J$) varies across models.  For the majority of models, Pathfinder works better with longer histories.  There is substantial improvement for \verb|diamonds-diamonds|, whereas performance for \verb|sblrc-blr| declines with larger $J$. This is because in some cases we need more history to get a higher-rank estimate of covariance, whereas in other cases, it can hurt local adaptation to keep a longer history. The example \texttt{diamonds-diamonds} has 27 parameters and the posterior distribution exhibits high correlations among 25 parameters. Hence we observed a great improvement of Pathfinder with a larger $J$. We encourage a larger $J$ for approximation when the target distribution is expected to have high dependencies among a large number of parameters. Meanwhile, a smaller $J$ requires less memory and computational cost, which is more efficient for finding draws from the high probability mass region.
For the other 19 models in \texttt{posteriordb}, median 1-Wasserstein distances were reduced an average of 0.7\%
when moving from $J = 6$ to $J = 60$.  Among these 19 models, the maximum reduction was 16.3\%
and the maximum increase was 9.8\%.
In summary, a good choice of $J$ depends on the specific problem. We found that, in general, $J = 6$ works well for the models in \texttt{posteriordb}.  Our default history length is in line with defaults used for $J$ in the range of 5--10 used as the default for most software distributions of L-BFGS.  For example, R's \texttt{stats::optim()} function defaults to 5, whereas \texttt{SciPy} uses 10 as the default for its \verb|scipy.optimize.fmin_l_bfgs_b| function \citep{virtanen2020scipy}.

\begin{figure}[t!]
	\centering{
		\includegraphics[width=.9\textwidth, keepaspectratio]{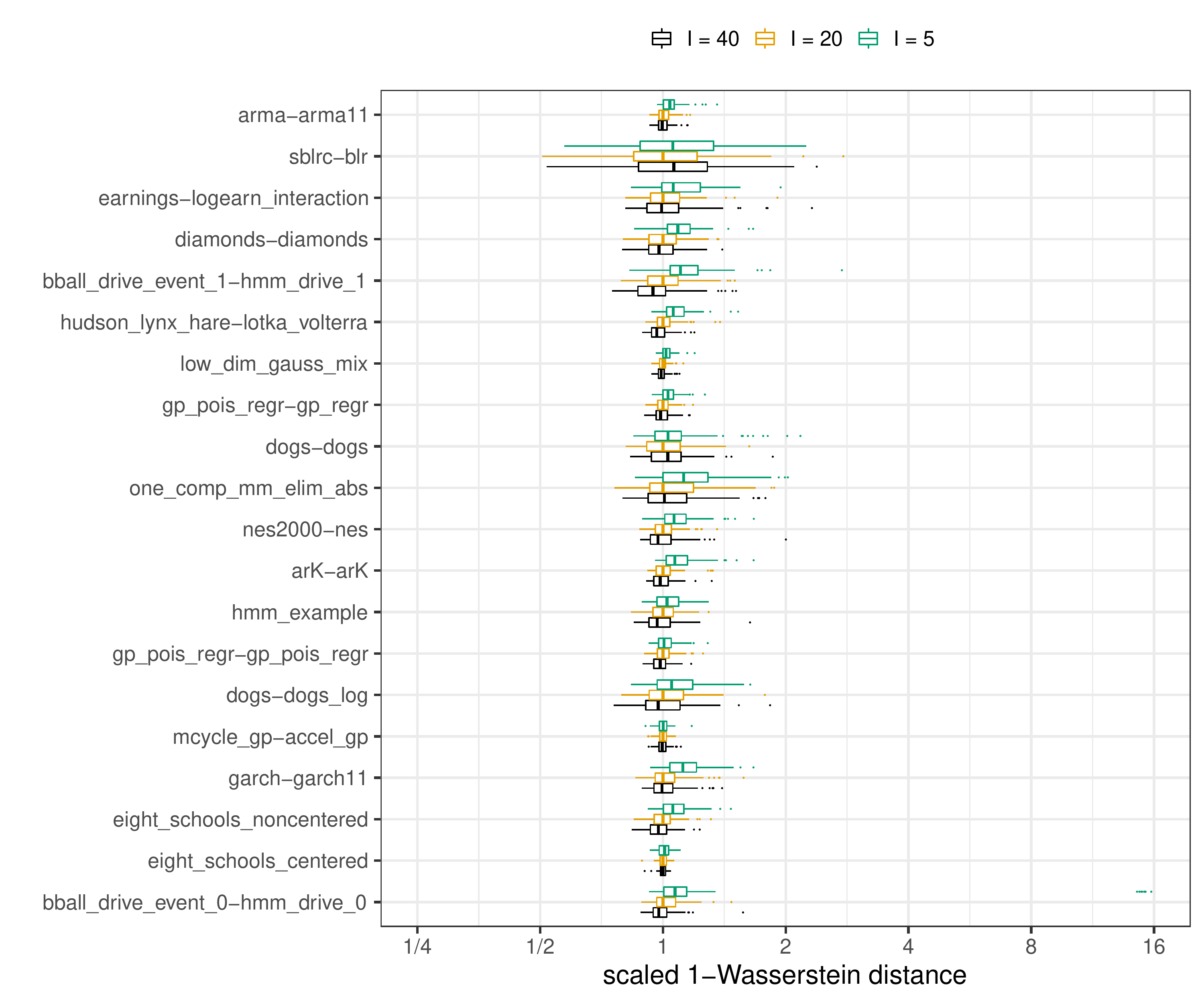}}
	\caption{\it
		Box plots of 1-Wasserstein distances between the reference posterior samples and approximate draws from multi-path Pathfinder for examples in \texttt{posteriordb} when using $I = 5$ (green), 20 (orange, default), and 40 (black) independent Pathfinder runs.  The results are scaled by the median of 1-Wasserstein distances for multi-path Pathfinder with $I = 20$.}\label{fig: sens_test_I}
\end{figure}

\paragraph{Number of parallel runs.}
To evaluate how sensitive multi-path Pathfinder is to the number of single-path Pathfinder runs used ($I$), we reproduce 100 runs of multi-path Pathfinder with $I \in \{ 5, 20, 40 \}$, generating $R = 100$ approximate draws from each run (20 is our suggested default setting). We left all other tuning parameters at their proposed default values. We estimate the 1-Wasserstein distance for each run of multi-path Pathfinder to compare the performance of multi-path Pathfinder under different values of $I$.  Figure~\ref{fig: sens_test_I} shows that the change of $I$ does not have much impact on the approximate performance of multi-path Pathfinder.   When decreasing $I$ from $20$ to $5$, the median 1-Wasserstein distances only increase 5.4\%
on average, with a minimum increase of 0.1\% and maximum increase of 12.5\%.
When increasing $I$ from $20$ to $40$, the median 1-Wasserstein distances are reduced by only 1.1\% on average, with a maximum reduction of 5.5\% and maximum increase of 6.4\%.
Increasing $I$ from $5$ to over $20$ eliminates the extreme outcomes for the \texttt{bball\_drive\_event\_0-hmm\_drive\_0} example, which we further investigate in the next section.

\subsection{Pathfinder for posteriors with challenging geometry}\label{subsec: case_study_pf}

In this section, we consider four problems that challenge optimizers and MCMC samplers and thus might be expected to challenge Pathfinder.

\paragraph{Neal's funnel.}
\citet{neal2003slice} presents a model with funnel-like posterior geometry, defined by
\begin{equation}\label{eq: neal_funnel}
	p(\tau, \beta_1, \ldots, \beta_N)
	= \textrm{normal}(\tau \mid 0, 3)
	\cdot \prod_{n = 1}^{N} \textrm{normal}(\beta_n \mid 0, \exp(\tau / 2)).
\end{equation}
For values of $\tau > 0$, the $\beta_n$ are relatively free (the ``mouth'' of the funnel), whereas for $\tau < 0$, they are constrained to be near 0 (the ``neck'' of the funnel).  This density is problematic for at least two reasons.  First, the density grows without bound (i.e., has no maximum) as $\tau \rightarrow -\infty$ and $\beta_n \rightarrow 0$.  Second, the condition number of the inverse Hessian (ratio of largest to smallest eigenvalue) grows quickly as $\tau$ moves away from 0, which bounds the efficacy of gradient-based updates.  Furthermore, there is no way to globally precondition this distribution, as the curvature changes direction as $\tau$ moves away from 0 to the positive and negative side.  The poor conditioning is endemic to hierarchical models, even with data \citep{betancourt2015hamiltonian}.  The \texttt{posteriordb} package includes the eight schools model of \citet{rubin1981estimation}, in both a centered and non-centered parameterization;  see \citet{papaspiliopoulos2003non, stan2021users, betancourt2015hamiltonian} for more information on these parameterizations.

\begin{figure}[t!]
	\centering{
		\subfloat[optimization paths, $\textrm{uniform}(-2,2)$ inits\label{subfig: 8_tr_22}]{\includegraphics[width=0.3\textwidth, keepaspectratio]{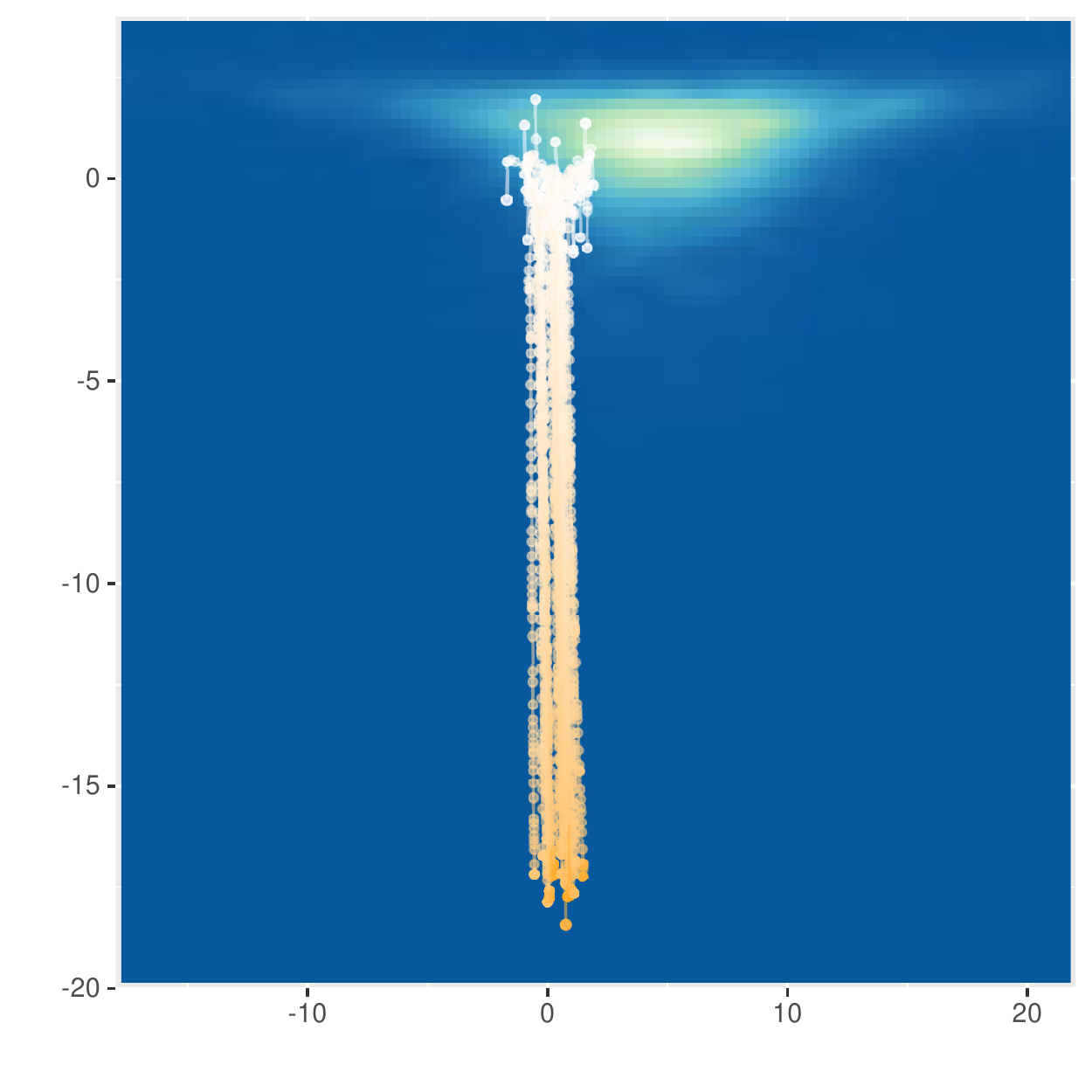}}
		\qquad \qquad
		\subfloat[approximate draws from multi-path Pathfinder \label{subfig: 8_points_22}]{\includegraphics[width=0.3\textwidth, keepaspectratio]{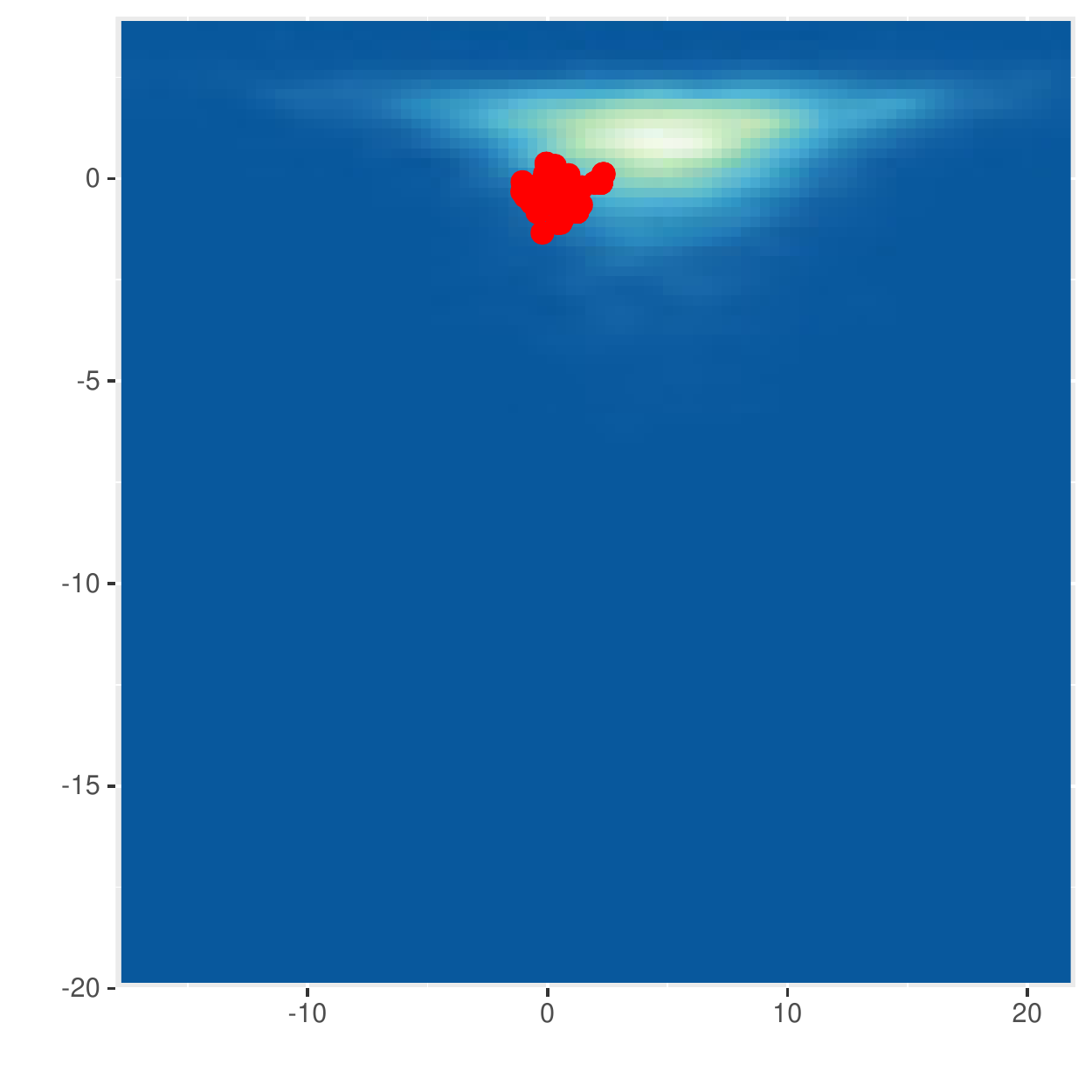}}\\
		
		\subfloat[optimization paths, $\textrm{uniform}(-15, 15)$ inits \label{subfig: 8_tr}]{\includegraphics[width=0.3\textwidth, keepaspectratio]{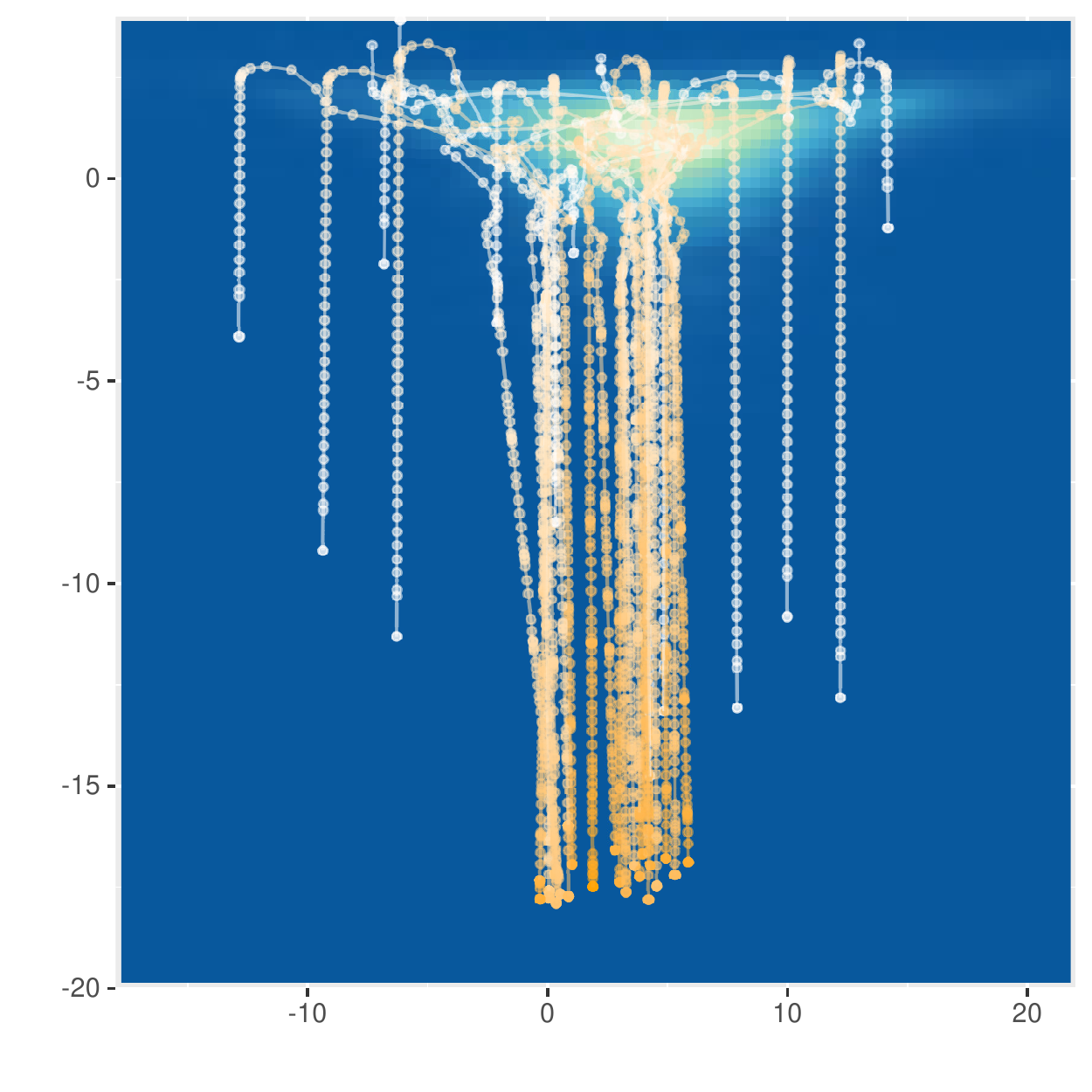}}
		\qquad \qquad
		\subfloat[approximate draws from multi-path Pathfinder \label{subfig: 8_points}]{\includegraphics[width=0.3\textwidth, keepaspectratio]{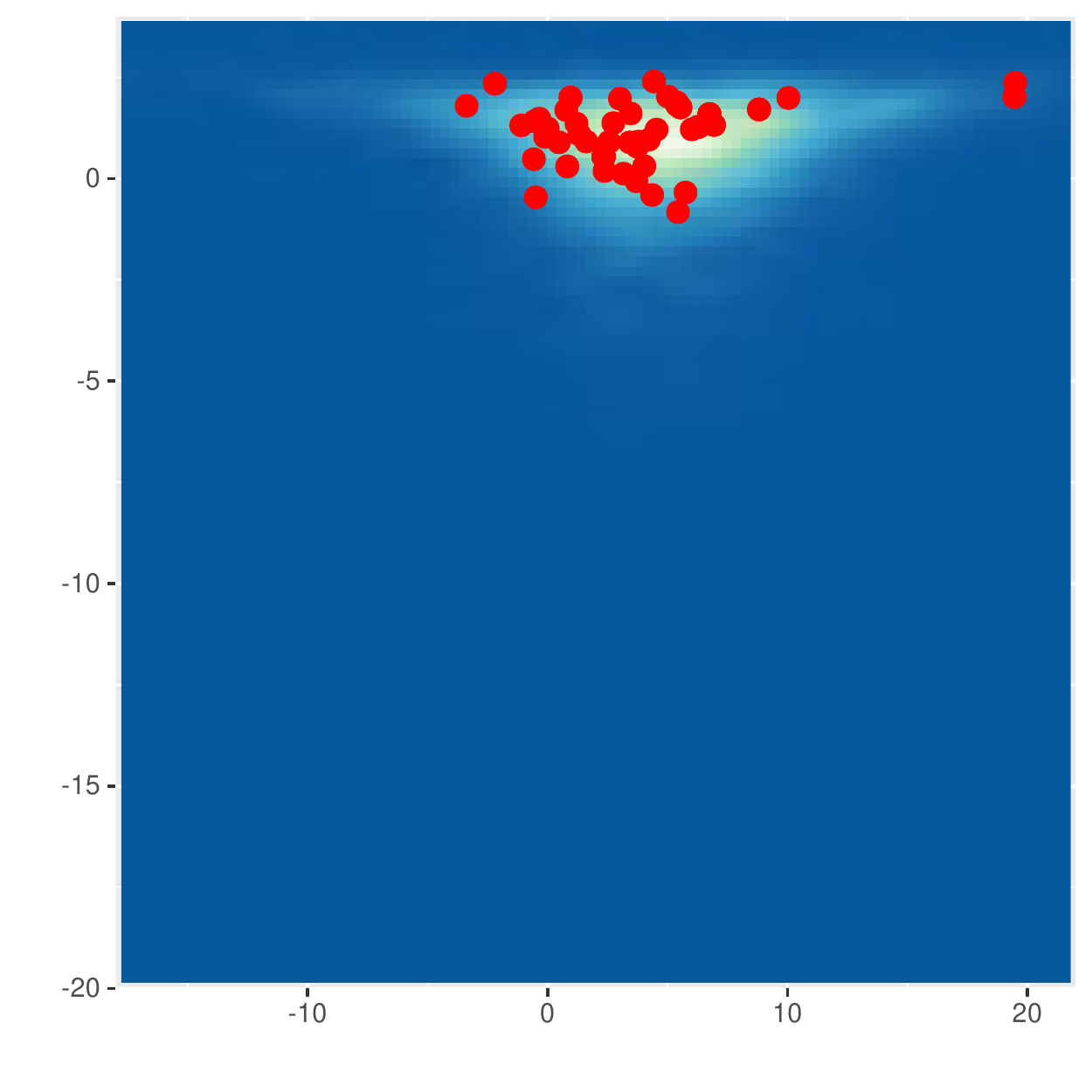}}
	}
	\caption{\it Results of multi-path Pathfinder with varying initial distribution for the eight-schools model with a centered parameterization. Optimization paths (left column) are displayed next to the points points selected by pathfinder (right column), with $\textrm{uniform}(-2,2)$ initialization (top row) and $\textrm{uniform}(-15, 15)$ initialization (bottom row). Multi-path Pathfinder used all 20 optimization paths. The density plots use lighter color for higher density.  The optimization paths are plotted in the left panels using white for initial points and orange for later points in the optimization trajectories;  later points dive into the neck of the funnel where density increases without bound but volume and hence probability mass is low. }\label{fig: eight-school}
\end{figure}
In Figure~\ref{fig: eight-school}, we plot the behavior of the centered parameterization of the eight schools model as an example of funnel-like behavior, where the optimization paths and approximate draws by multi-path Pathfinder when initials of Pathfinder are randomly generated from $\textrm{uniform}(-2, 2)$ (our default) or $\textrm{uniform}(-15, 15)$ distributions in the unconstrained parameter space. We observe that even though the optimization paths correctly follow an optimization trajectory down the neck of the funnel, Pathfinder successfully identifies points in the high probability mass (not high density) region, which is evenly split above and below $\tau = 0$.  The figure also shows that Pathfinder is sensitive to the choice of initial distribution $\pi_0$. When the initial distribution is concentrated in a region of high target density, the optimization paths may not pass through all regions of high probability mass, resulting in approximate draws clustered in a relatively small region as shown in the right-hand side plot in Figure~\ref{fig: eight-school}. Meanwhile, ADVI tends to be less sensitive to the initials in this example, outperforming Pathfinder in both 1-Wasserstein distance and ELBO, as shown in Appendix~\ref{appendix: ELBO_compar}.

\begin{figure}[t!]
	\centering{
		\subfloat[optimization paths with $\textrm{uniform}(-2, 2)$ initialization\label{subfig: funnel_tr_22}]{\includegraphics[width=0.3\textwidth, keepaspectratio]{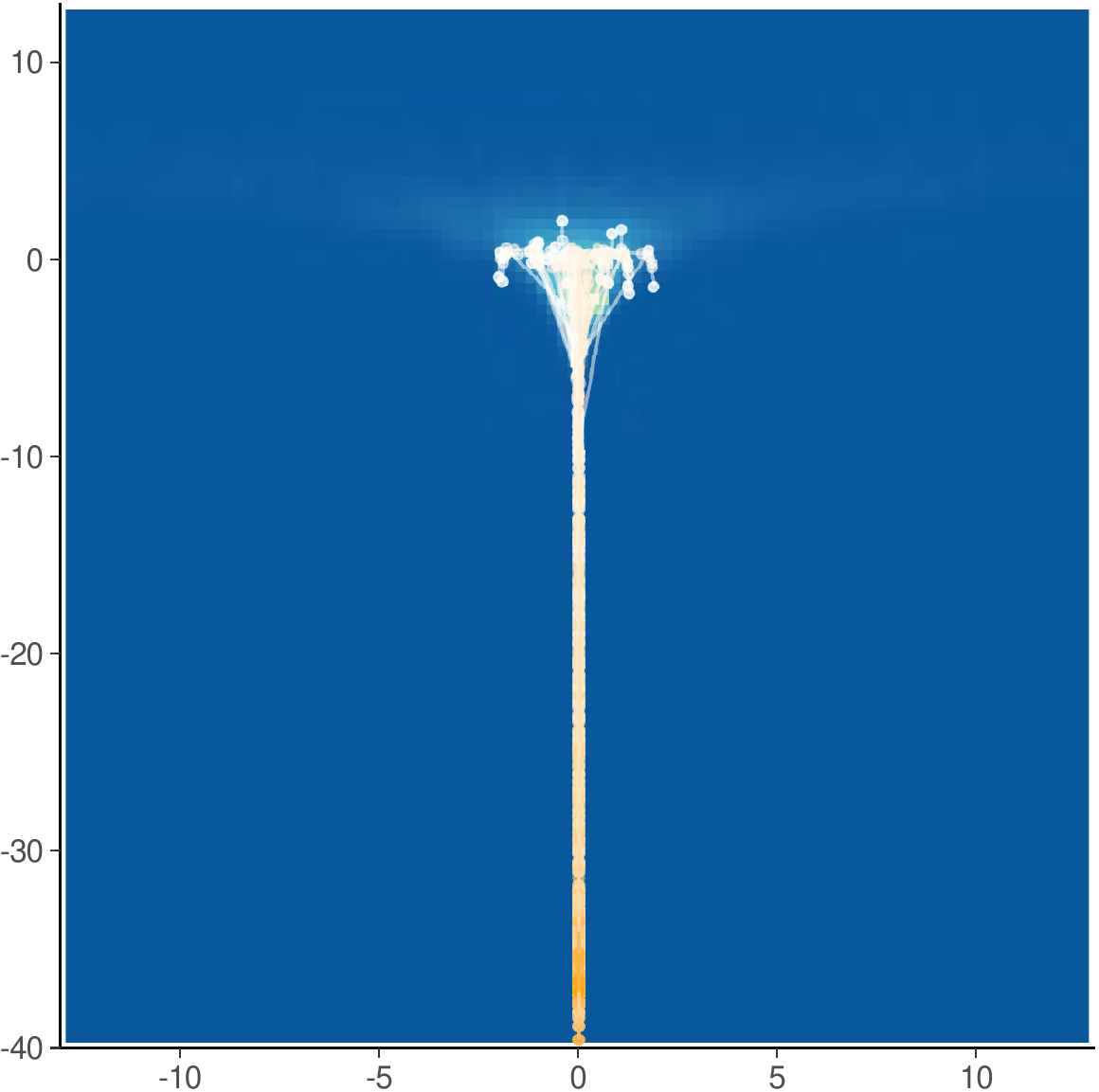}}
		\qquad \qquad
		\subfloat[approximate draws from multi-path Pathfinder \label{subfig: funnel_points_22}]{\includegraphics[width=0.3\textwidth, keepaspectratio]{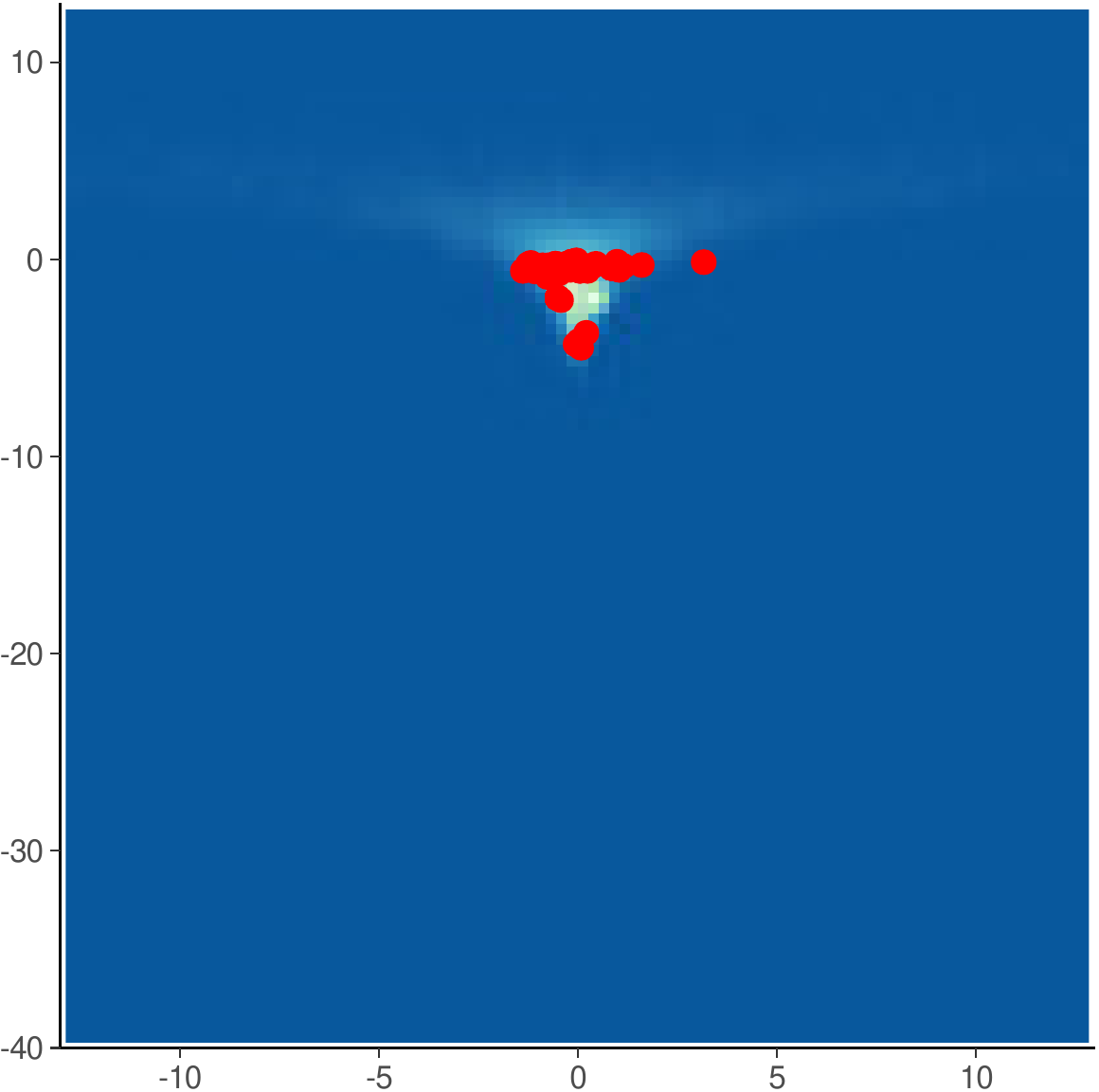}}\\
		\subfloat[optimization paths with $\textrm{uniform}(-10, 10)$ initialization\label{subfig: funnel_tr}]{\includegraphics[width=0.3\textwidth, keepaspectratio]{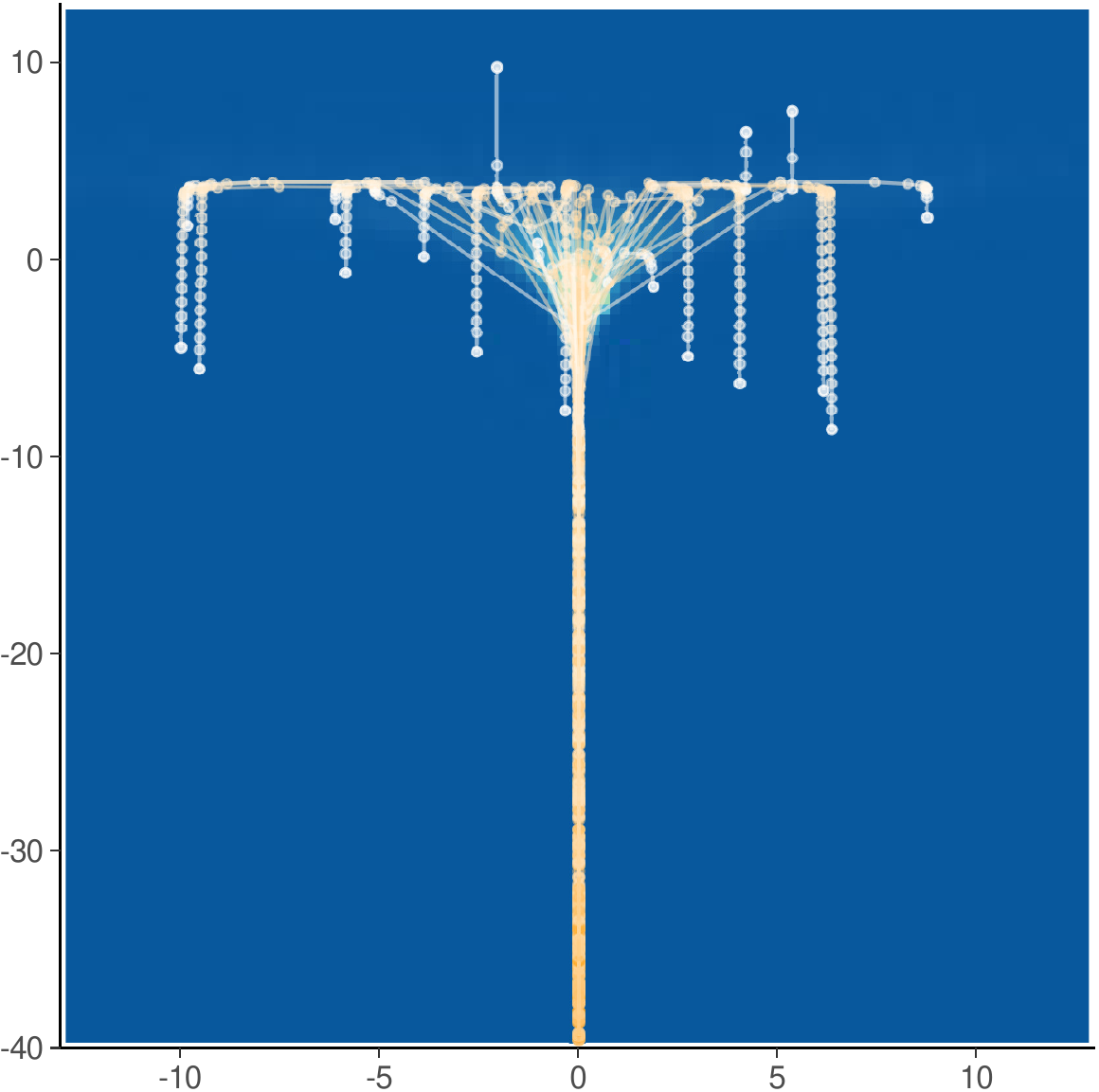}}
		\qquad \qquad
		\subfloat[approximate draws from multi-path Pathfinder \label{subfig: funnel_points}]{\includegraphics[width=0.3\textwidth, keepaspectratio]{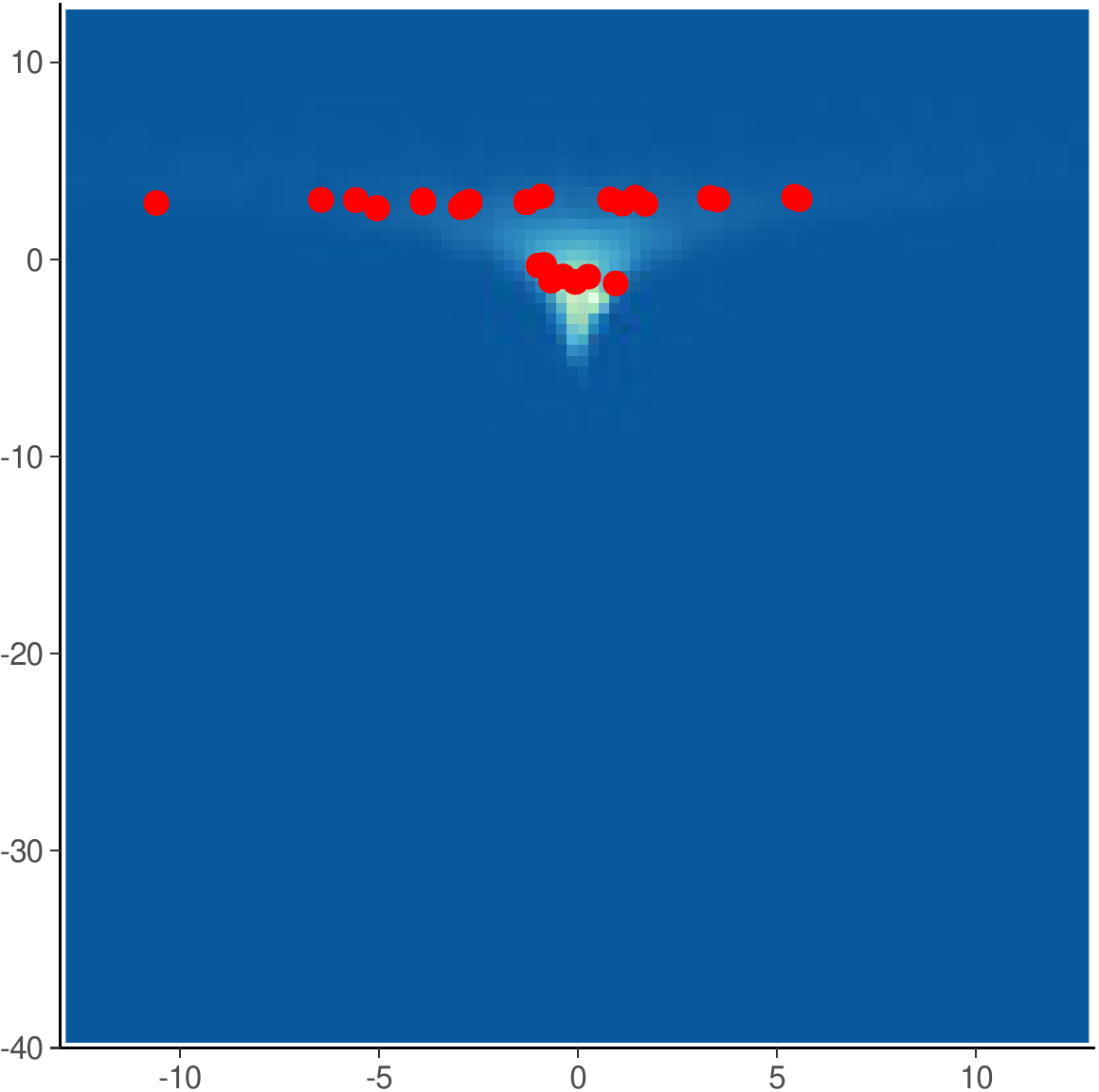}}
	}
	\caption{\it Results of multi-path Pathfinder with varying initial distribution for the 100-dimensional Neal's funnel model. The optimization paths (left column) are displayed next to the points points selected by pathfinder (right column), with $\textrm{uniform}(-2,2)$ initialization (top row) and $\textrm{uniform}(-10, 10)$ initialization (bottom row). Multi-path Pathfinder used all 20 optimization paths. The density plots illustrate region with higher probability with lighter color.  The optimization paths are plotted in the left panels using white for initial points and orange for later points in the optimization trajectories. The reference samples for density plots are generated by 4 MCMC chains fitted with \texttt{cmdstanr}, with an adaptation period of 100,000 iterations, 850,000 saved iterations, and a thinning rate of 300.}\label{fig: funnel100}
\end{figure}
We will next consider a medium-dimensional instance of Neal's funnel. Letting $N$ in \eqref{eq: neal_funnel} be $99$, we fit the 100-dimensional Neal's funnel model with multi-path Pathfinder using our proposed default settings.  As in the eight-schools example, Pathfinder successfully locates the high probability mass region along the optimization paths as illustrated in Figure~\ref{fig: funnel100}. With Pathfinder's default initialization, the approximate draws are underdispersed when Pathfinder is started at the region of high posterior density, and the approximation can be greatly improved with a more diffuse initialization distribution.

\paragraph{Non-normal posteriors.}

\begin{figure}[t!]
	\centering{
		\subfloat[optimization paths \label{subfig: 18_tr}]{\includegraphics[width=0.3\textwidth, keepaspectratio]{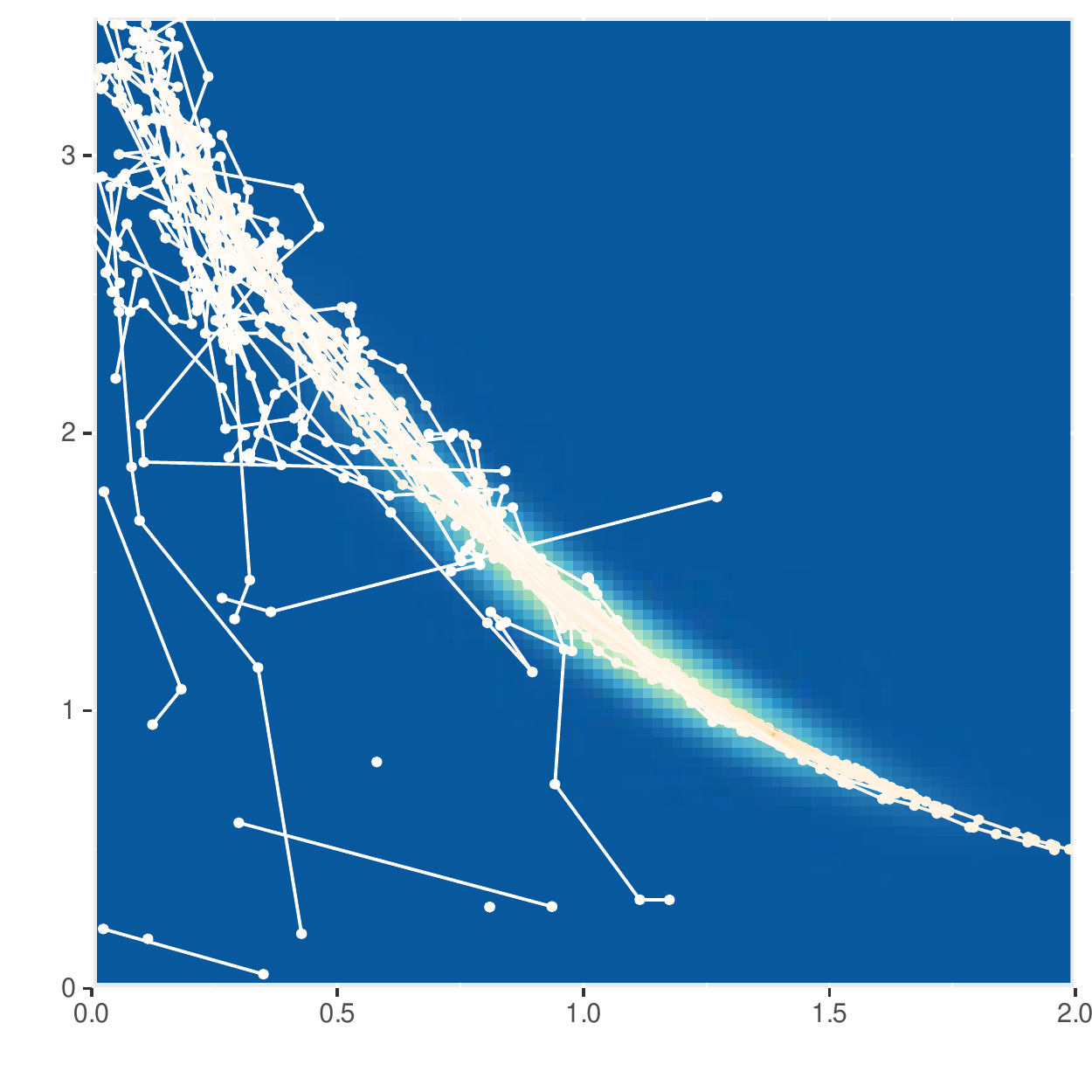}}
		\qquad \qquad
		\subfloat[draws from multi-path Pathfinder \label{subfig: 18_points}]{\includegraphics[width=0.3\textwidth, keepaspectratio]{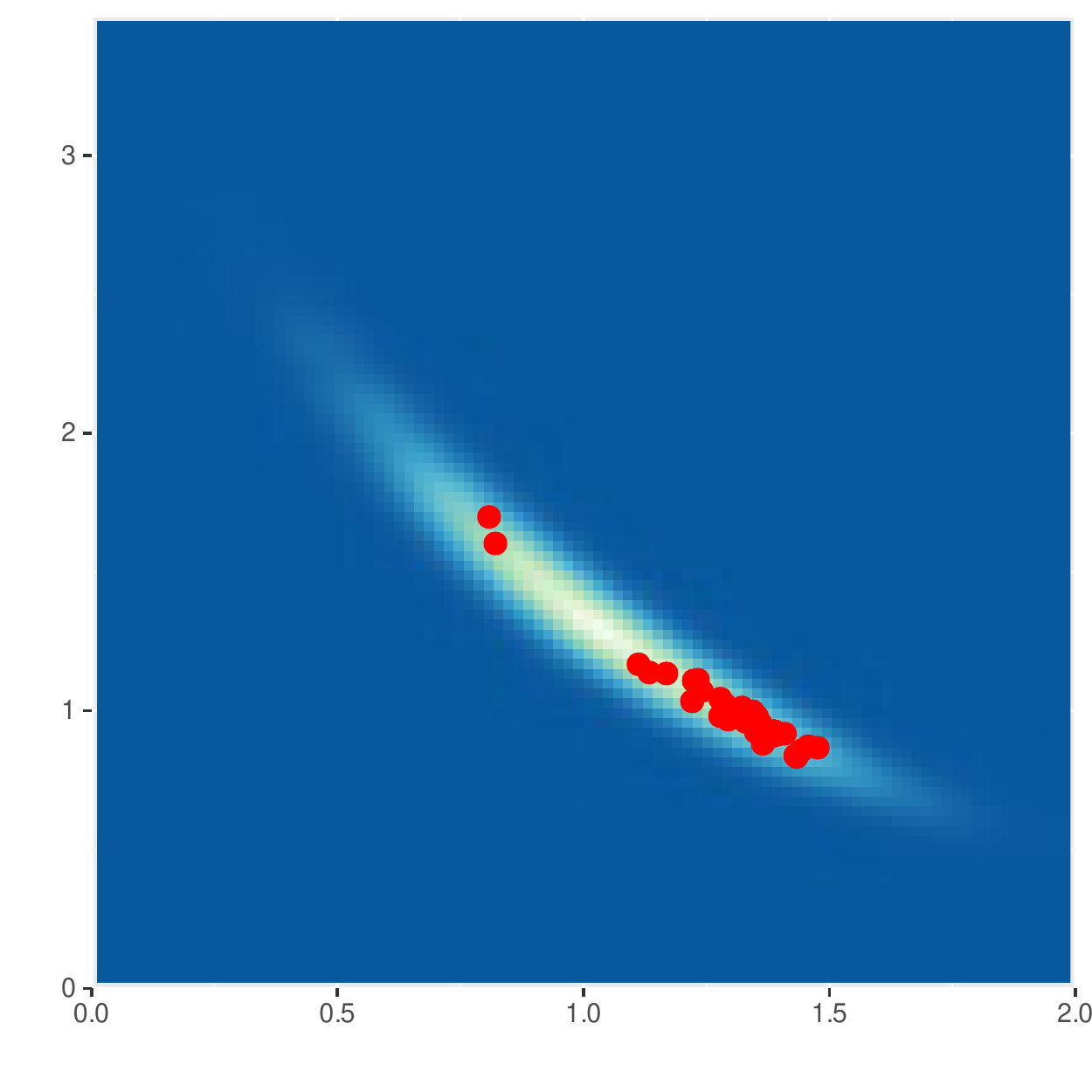}}}
	\vspace{-.1in}
	\caption{\it (a) 20 optimization paths and (b) the 100 approximate draws of multi-path Pathfinder, overlaid on the density plot of two selected parameters for the \texttt{gp\_pois\_regr-gp\_pois\_regr} example.  Multi-path Pathfinder used all 20 optimization paths. Lighter color in the density plot indicates higher density.  The optimization paths are shaded from white (initial) to orange (final).}\label{fig: 18-gp_pois_regr}
\end{figure}
Given that Pathfinder and ADVI rely on normal approximations to generate approximate draws, we are interested in the behavior of Pathfinder for posteriors that are far from normal.  We focus here on the Gaussian process Poisson regression model \verb|gp_pois_regr-gp_pois_regr| to explore the performance of Pathfinder in approximating non-Gaussian posteriors. As shown in Figure~\ref{fig: 18-gp_pois_regr}, the high probability mass region for this model projected down to two selected dimensions has the shape of a waning moon.  The approximate draws generated by Pathfinder concentrate in a relatively small region within the high probability mass region. This phenomenon is expected, because Pathfinder selects an approximate normal distribution based on minimizing KL divergence to the target density, which favors more concentrated approximations that fall within the bulk of the target probability mass.  Therefore, Pathfinder tends to make a conservative guess on the high probability region when the posterior cannot be well approximated by a normal distribution.  

\paragraph{Multimodality.} Multimodal posteriors have more than one local optimum.  In some cases, such as high-dimensional mixture models or neural networks, the multimodality can be so extreme that it defeats Monte Carlo methods. In other cases, the posterior has one, or maybe a few major modes, with other modes having negligible probability mass.  Off-the-shelf MCMC sampling can work in these cases if it's possible to move among the modes, but this is usually too difficult, and specialized samplers for multimodal problems need to be employed, such as bridge sampling \citep{meng1996simulating}.  If there is only one major mode, MCMC will succeed if it's initialized near that mode or if chains initialized near minor modes can escape.  

\begin{figure}[t!]
	\centering{
		\subfloat[100 draws from Stan phase I warmup \label{subfig: 3_phI}]{\includegraphics[width=0.33\textwidth, keepaspectratio]{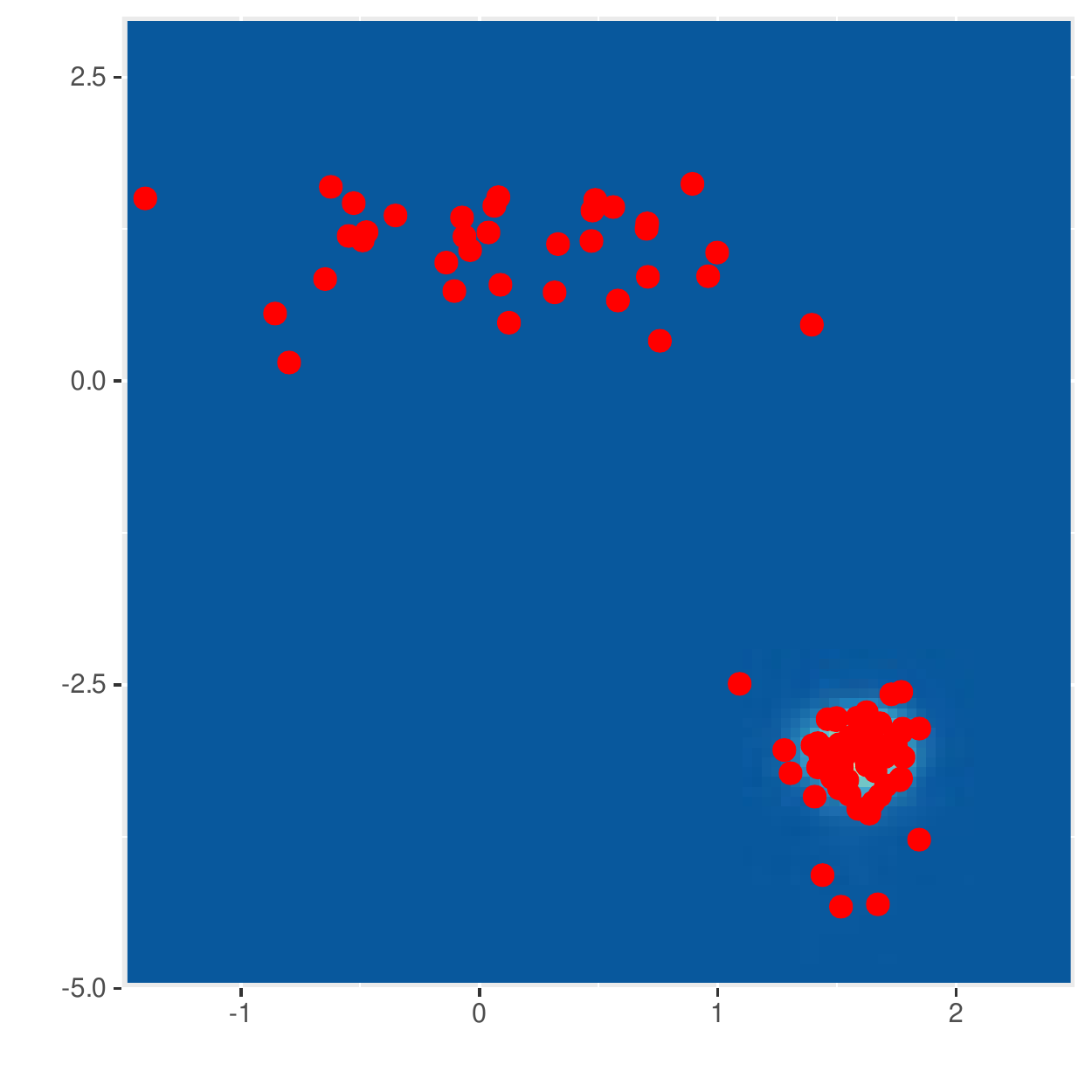}}
		\hfill
		\subfloat[20 optimization paths \label{subfig: 3_tr} ]{\includegraphics[width=0.33\textwidth, keepaspectratio]{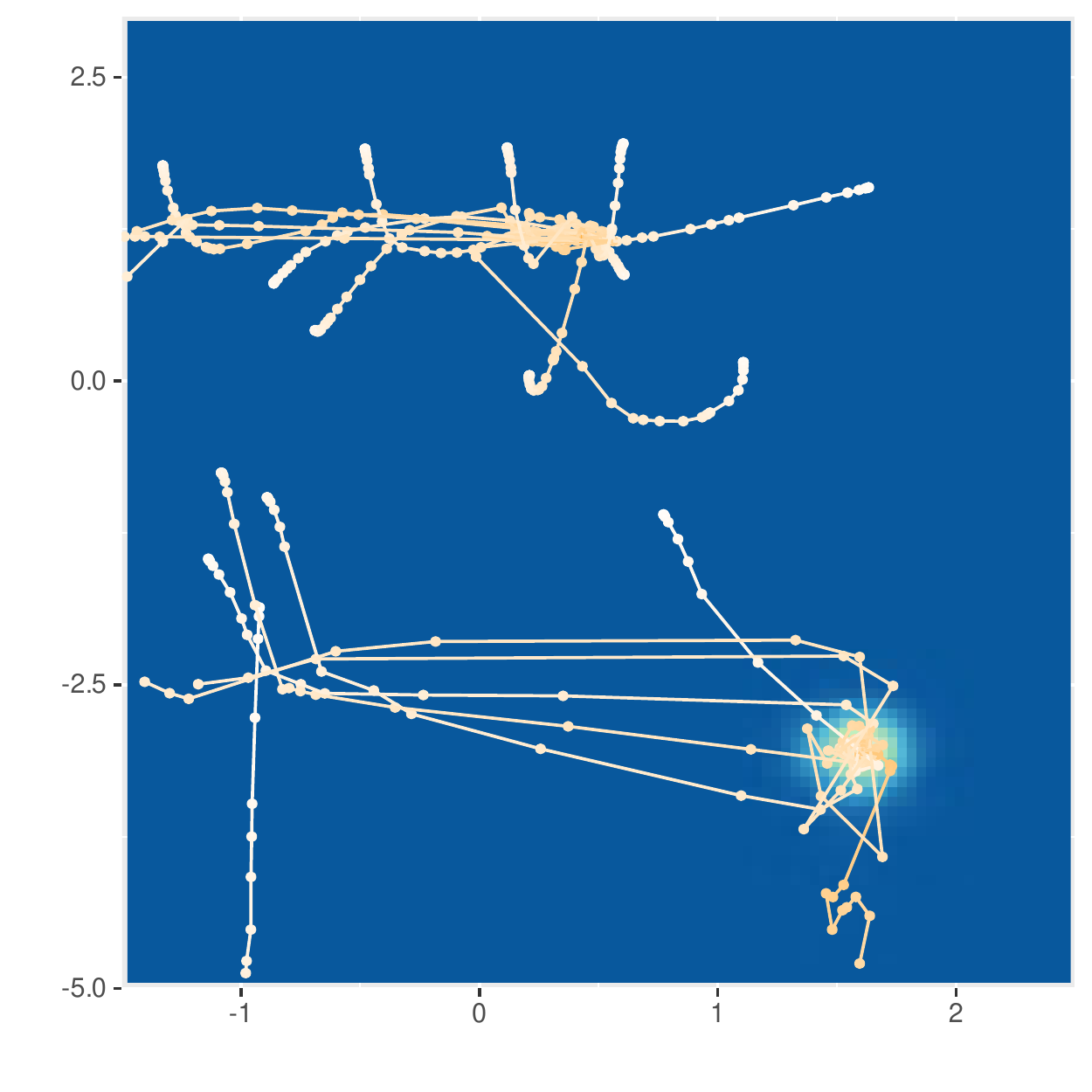}}
		\hfill
		\subfloat[100 approximate draws from multi-path Pathfinder \label{subfig: 3_points}]{\includegraphics[width=0.33\textwidth, keepaspectratio]{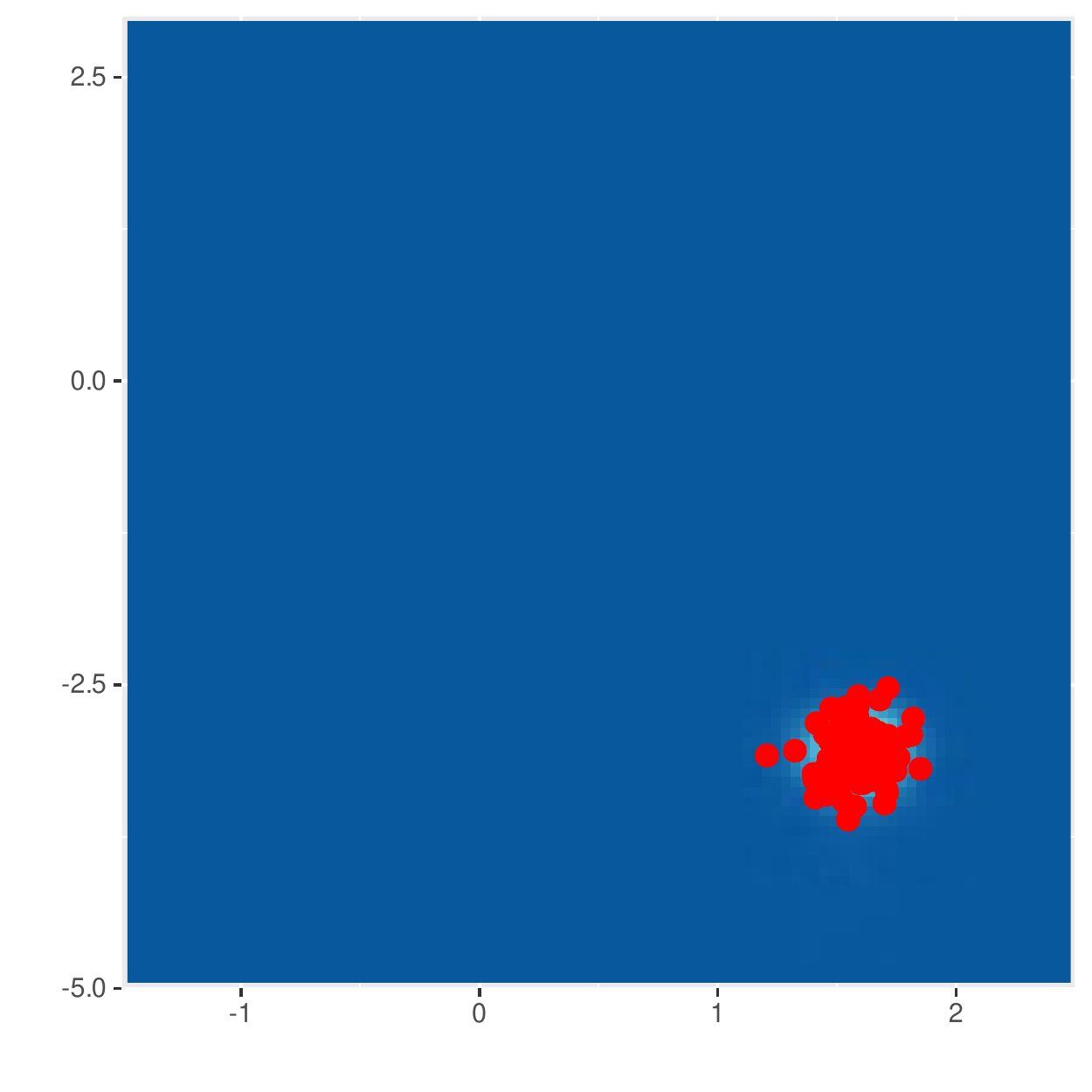}}}
	\caption{\it (a) 100 initializations generated by Stan phase I warmup, (b) 20 optimization paths, and, and (c) the 100 approximate draws of multi-path Pathfinder, overlaid on the density plot of two selected parameters for the \texttt{bball\_drive\_event\_0-hmm\_drive\_0} example.  The density plot illustrates the region with higher probability with lighter color. In (b), the optimization paths are plotted from white (initialization) to orange (later iterations).}\label{fig: multimodal}
\end{figure}

We observed severely biased MCMC sampling due to the existence of minor modes for 4 of the 49 test models in \texttt{posteriordb}. In Figure~\ref{fig: multimodal} we illustrate 100 approximate draws generated by Stan phase I sampler, optimization paths for 20 runs of Pathfinder, and 100 approximate draws by multi-path Pathfinder using all 20 runs of Pathfinder for the \verb|bball_drive_event_0-hmm_drive_0| example in \texttt{posteriordb}.  This posterior has multiple modes as shown in Figure~\ref{subfig: 3_tr}. Figure~\ref{subfig: 3_phI} reveals how Stan's phase I adaptation can get stuck near this minor mode.  The importance resampling step of multi-path Pathfinder is able to filter out points around minor modes as shown in Figure~\ref{subfig: 3_points}.  Of course, when all optimization paths are trapped in minor modes, importance resampling cannot recover. As a result, multi-path Pathfinder is more robust with more single-path Pathfinder runs from which to importance resample.

\begin{figure}[t!]
	\vspace{-.2in}
	\centering{
		\includegraphics[width=0.8\textwidth, keepaspectratio]{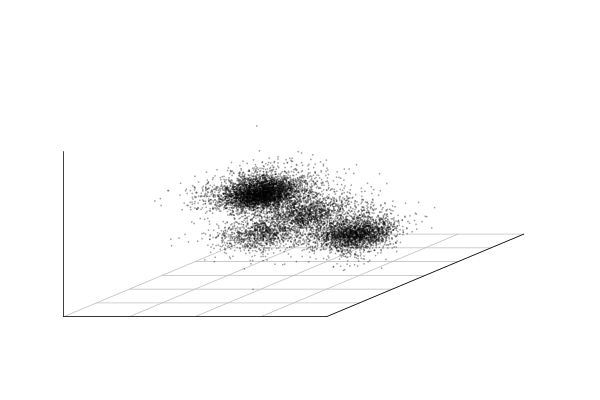}}
	\vspace{-.6in}
	\caption{\it Illustration of multimodality for the \texttt{ovarian-logistic\_regression\_rhs} example. The 3-d scatterplot illustrates the reference posterior samples for three selected parameters. We can see at least four major modes in this 3-d marginal posterior. Overall, we estimate the joint posterior to have hundreds of modes with non-negligible masses. 
	}\label{fig: ovarian_mode}
\end{figure}
We now turn to \verb|ovarian-logistic_regression_rhs|, a challenging high-dimensional example from \texttt{posteriordb}. The example works on ovarian cancer data containing gene expression measurements of tissues. The example fits a hierarchical logistic regression for the purpose of discriminating between tissues from different classes (e.g., tumor and normal samples).  The model has 3075 parameters and a horseshoe prior that performs soft variable selection \citep{piironen2017sparsity}.  The prior induces multiple major modes, one of which corresponds to all unconstrained coefficients being very close to zero.  
\citet[][Section~4.2]{piironen2017sparsity} provide a detailed discussion about the challenging features in this example.
This interesting example is not included in the tests in Sections~\ref{subsec: sim_pf_ADVI} and \ref{subsec: sim_sens} because the reference samples are not available in \texttt{posteriordb}. 
We produce a reference posterior sample by running 4 adaptive HMC chains in \texttt{cmdstanr} with 10,000 warmup iterations, 25,000 saved iterations, and a thinning rate of 10.  Figure~\ref{fig: ovarian_mode} provides a three dimensional scatterplot illustrating the multimodality of the reference draws.

\begin{figure}[t!]
	\centering{
		\subfloat[$\textrm{uniform}(-2, 2)$ initialization of multi-path Pathfinder\label{subfig: ovarian_1}]{\includegraphics[width=0.3\textwidth, keepaspectratio]{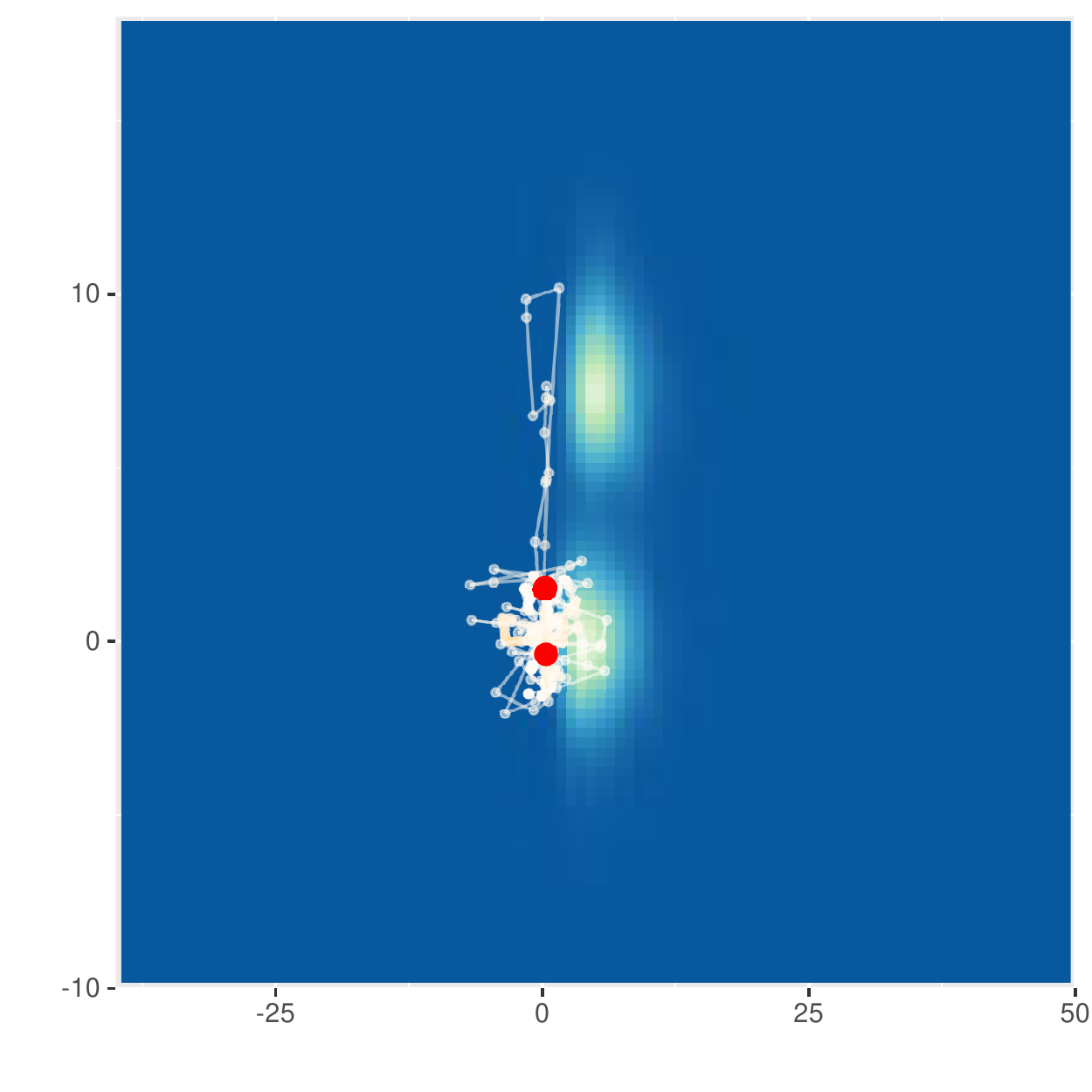}}
		\qquad \qquad
		\subfloat[random initialization from reference draws of multi-path Pathfinder\label{subfig: ovarian_3}]{\includegraphics[width=0.3\textwidth, keepaspectratio]{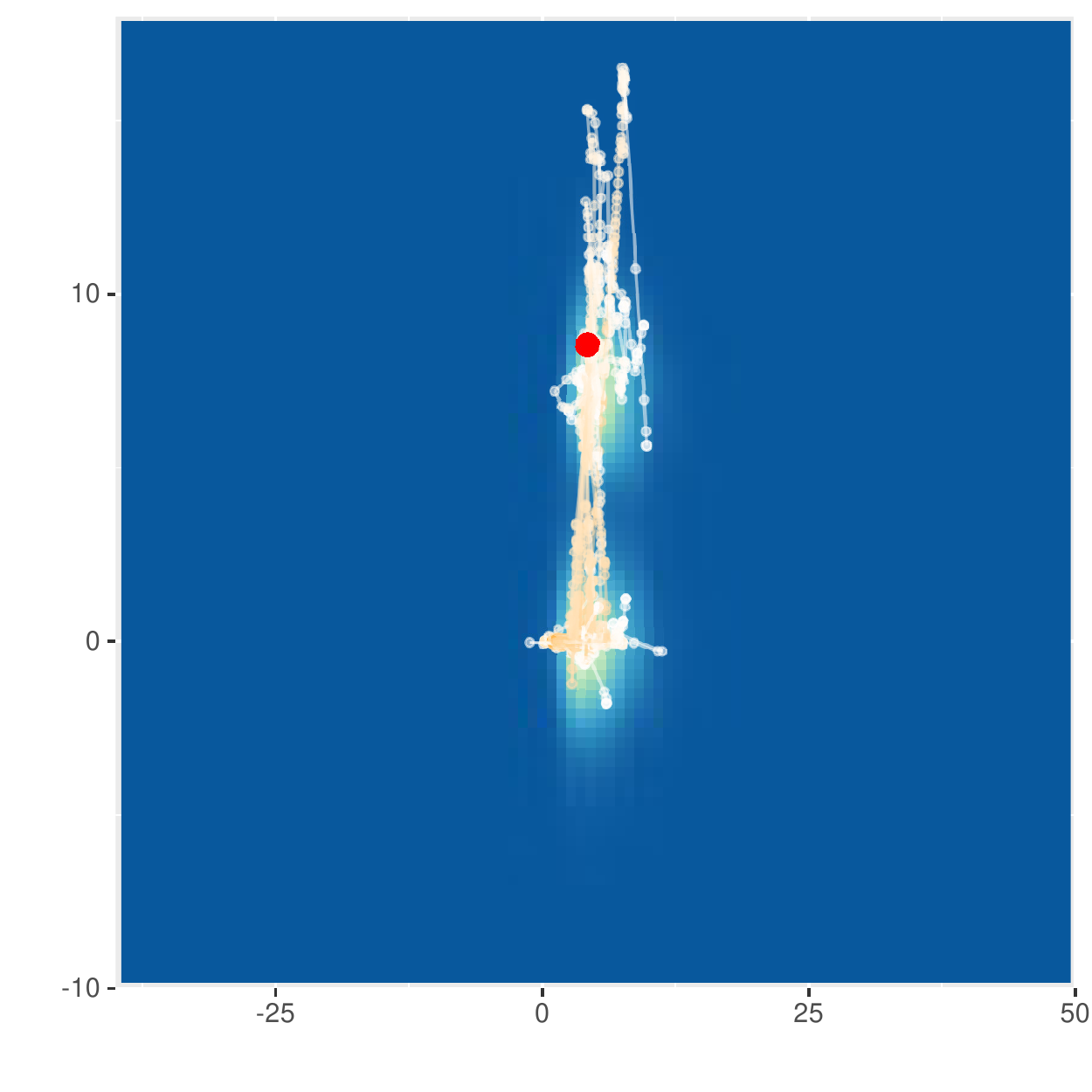}} \\
		\subfloat[predictive performance with $\textrm{uniform}(-2, 2)$ initialization\label{subfig: ovarian_1_p}]{\includegraphics[width=0.3\textwidth, keepaspectratio]{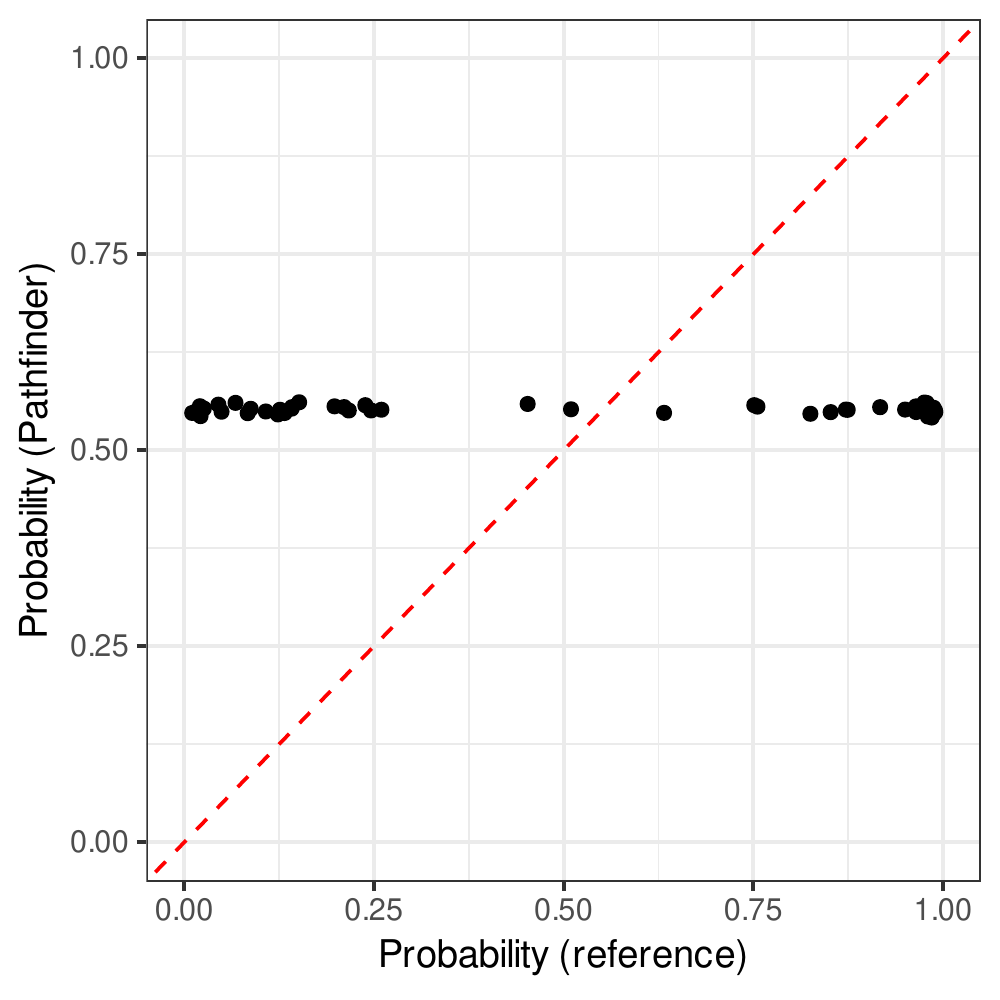}}
		\qquad \qquad
		\subfloat[predictive performance with reference initialization \label{subfig: ovarian_3_p}]{\includegraphics[width=0.3\textwidth, keepaspectratio]{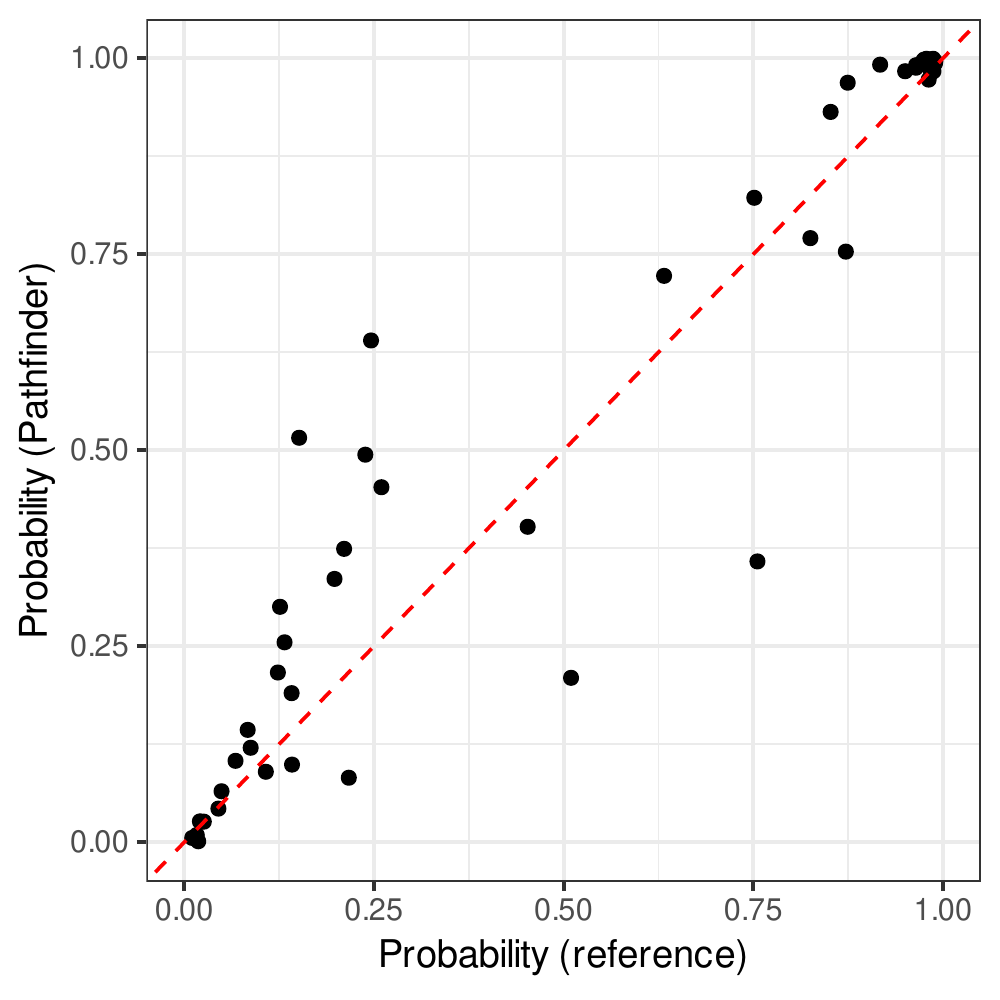}}} 
	\caption{\it Results of multi-path Pathfinder with varying initial distribution for the \texttt{ovarian-logistic\_regression\_rhs} example.  With $\textrm{uniform}(-2, 2)$ initialization (top left), multi-path Pathfinder does not explore both modes and predictive performance is poor (bottom left).  When initialized with draws from the reference sample (top right), multi-path Pathfinder explores both modes and performance improves (bottom right), but is still poor compared to the reference draws.  The density plots illustrate region with higher probability with lighter color.  The optimization paths are plotted from white (initialization) to orange (later iterations). The approximate draws of Pathfinder are indicated by red dots.  The density plots show two major modes for the two selected parameters (top row, background). Predictive performance is the estimated probability of a tumor, with the $x$-axis defined by the reference draws (bottom row).}\label{fig: ovarian}
\end{figure}
We fit multi-path Pathfinder with varying initial distributions and illustrate the results in Figure~\ref{fig: ovarian}. The starting points of Pathfinder determine the optimization paths and, therefore, the performance of Pathfinder. Meanwhile, we observe a small number of distinct samples returned by multi-path Pathfinder even when the optimization paths succeed in finding the high probability mass region. The small number of distinct samples shows that importance resampling fails to select appropriately diffuse samples, which provides evidence that the approximation found by Pathfinder is not good enough. 

Due to the high dimensionality and the shape of the posterior, the optimization from random initial values tends to find the major mode around the origin, which is consistent with all coefficients being close to zero.  This concentration around the origin leads to very poor predictive performance as shown in Figure~\ref{fig: ovarian}c. On the other hand, if Pathfinder could somehow be initialized within or near the minor modes, optimization will find them as shown in Figure~\ref{fig: ovarian}b. Thus, even though Pathfinder is finding a single region of high probability, in high dimensions it can fail to find other relevant high probability modes.

\begin{figure}[t!]
	\centering{
		\subfloat[default behavior of multi-path Pathfinder  \label{subfig: 32_12}]{\includegraphics[width=0.33\textwidth, keepaspectratio]{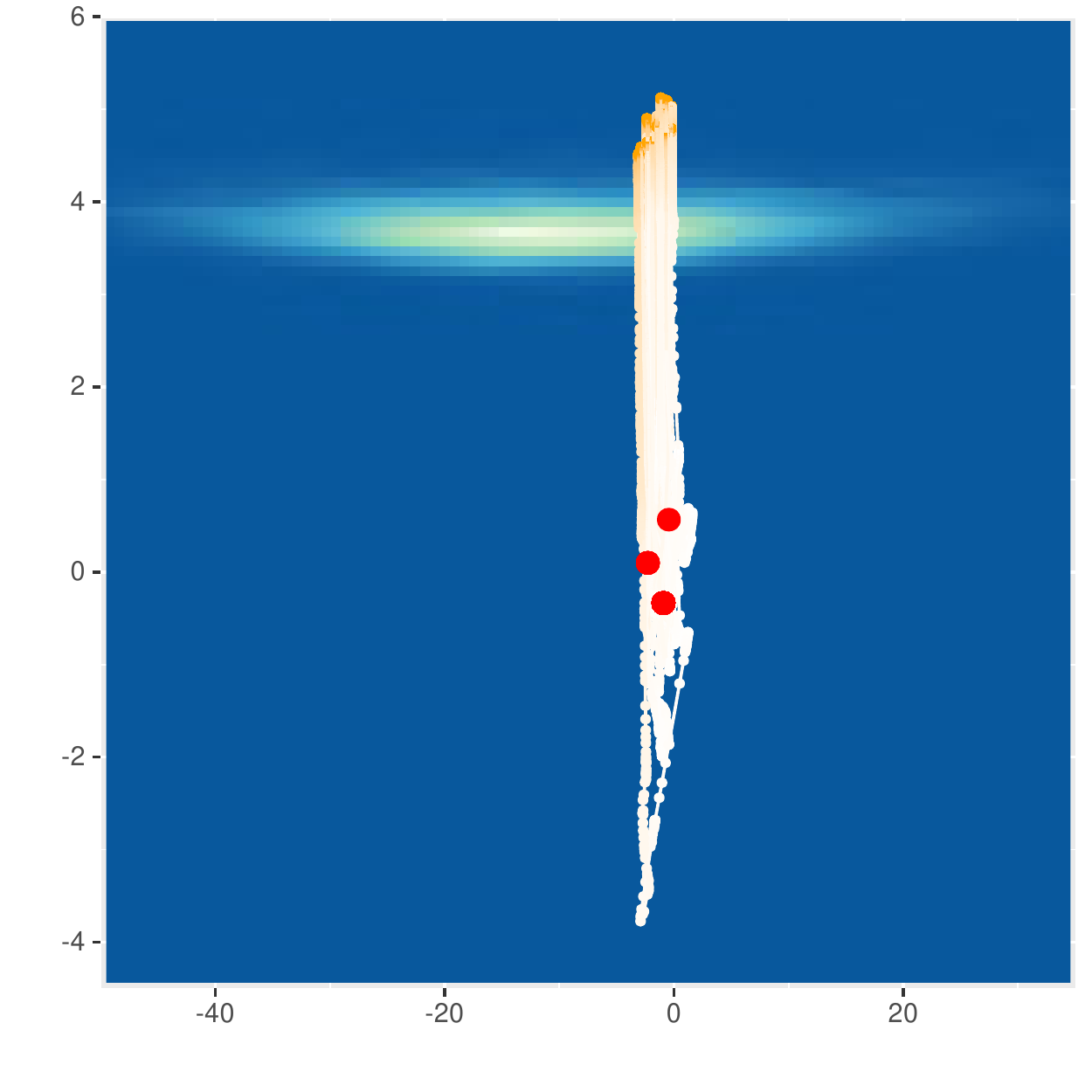}}
		\subfloat[central region of plot (a)\label{subfig: 32_12_s}]{\includegraphics[width=0.33\textwidth, keepaspectratio]{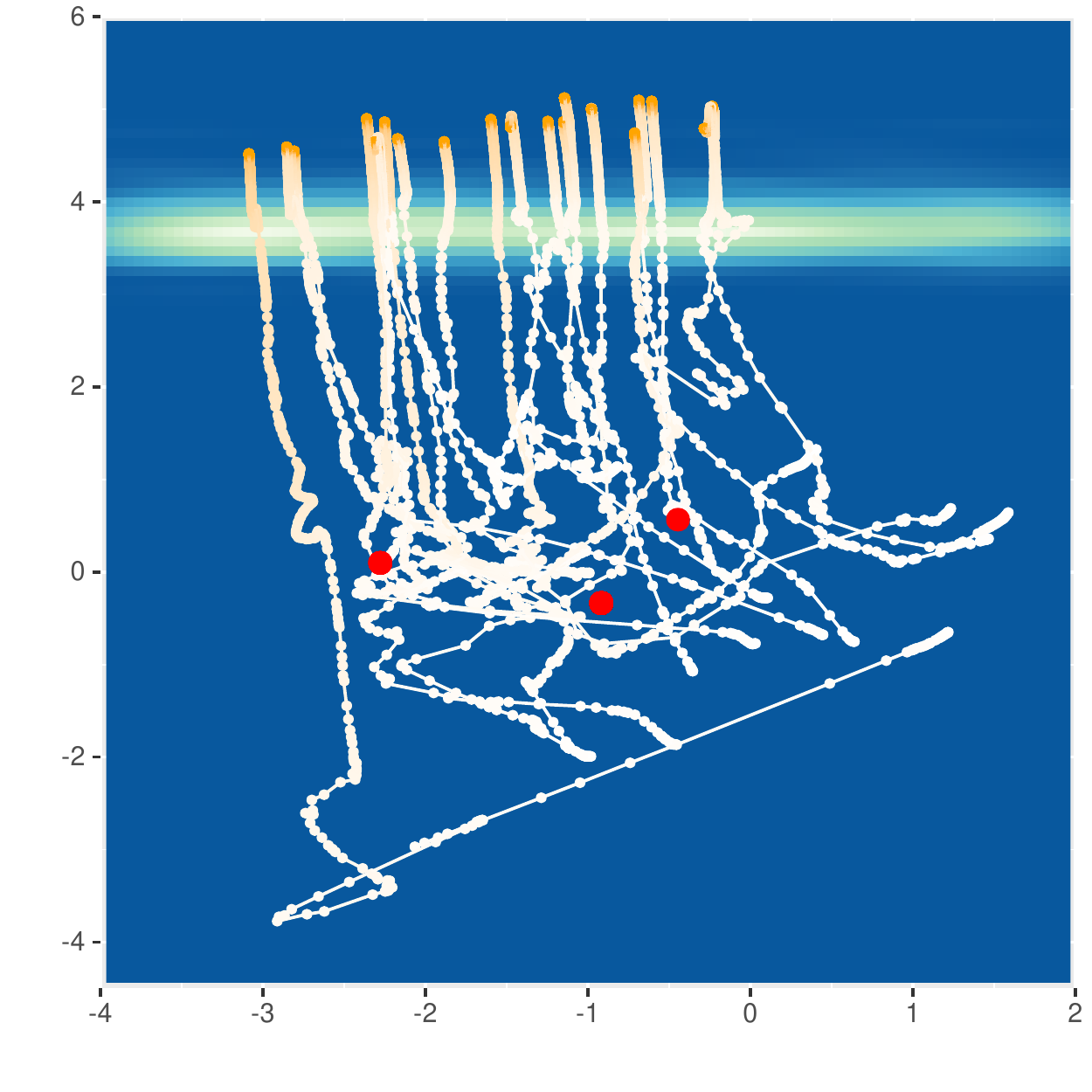}}
		\subfloat[result of multi-path Pathfinder with specialized initials \label{subfig: 32_12_2}]{\includegraphics[width=0.33\textwidth, keepaspectratio]{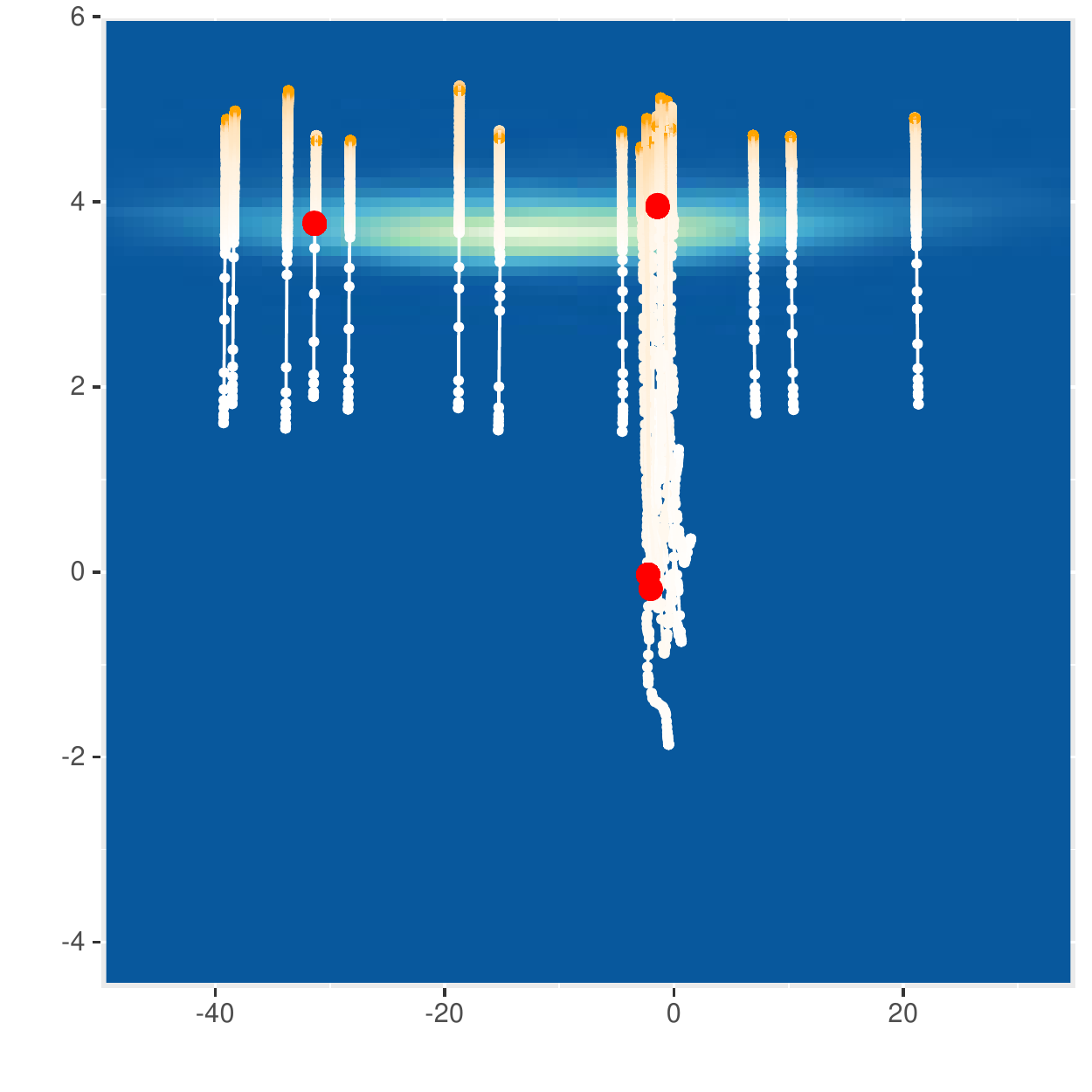}}\\
		\subfloat[Stan phase I sampler draws \label{subfig: 32_12_phI}]{\includegraphics[width=0.33\textwidth, keepaspectratio]{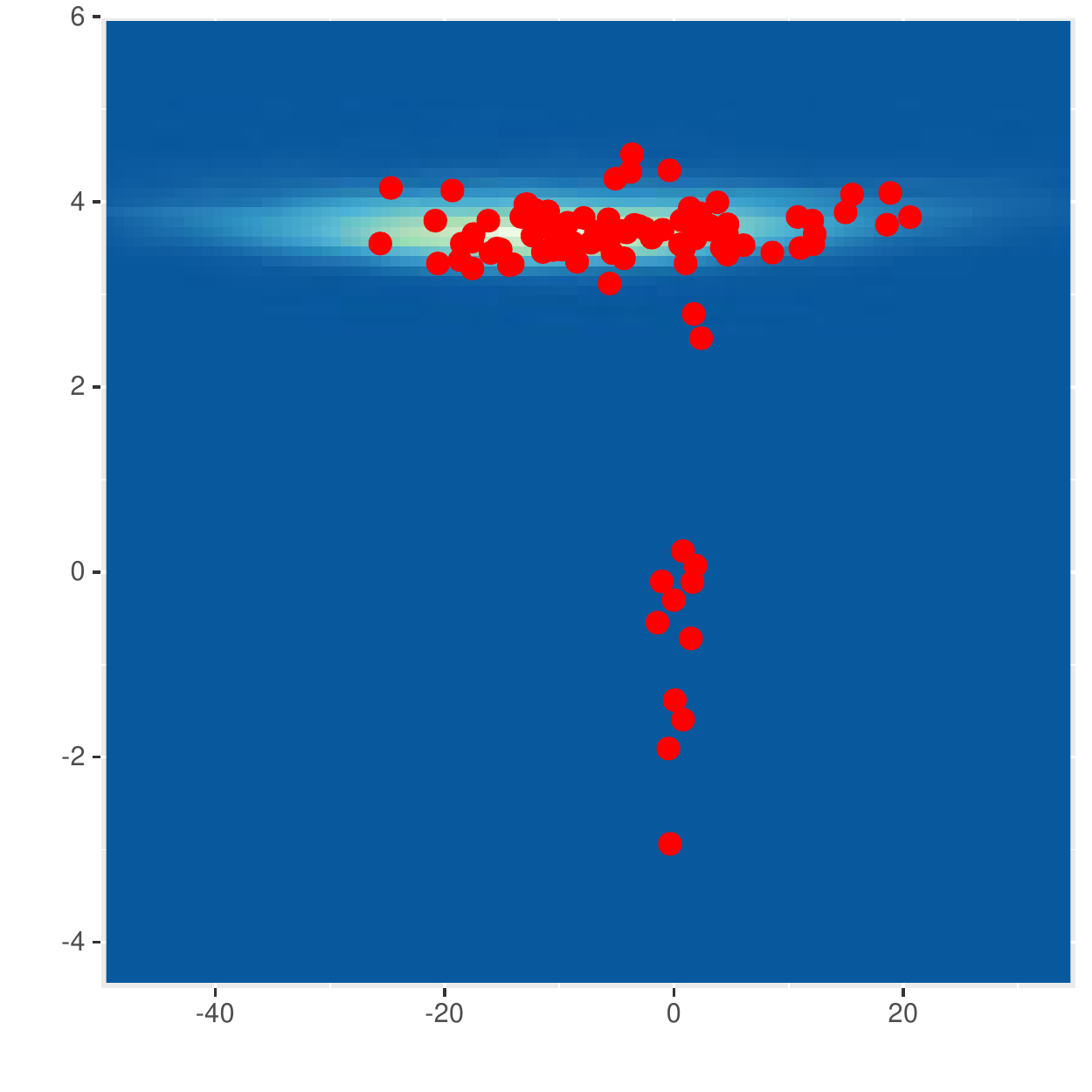}} 
		\subfloat[reproduce (b) with $J = 100$ \label{subfig: 32_12_100h}]{\includegraphics[width=0.33\textwidth, keepaspectratio]{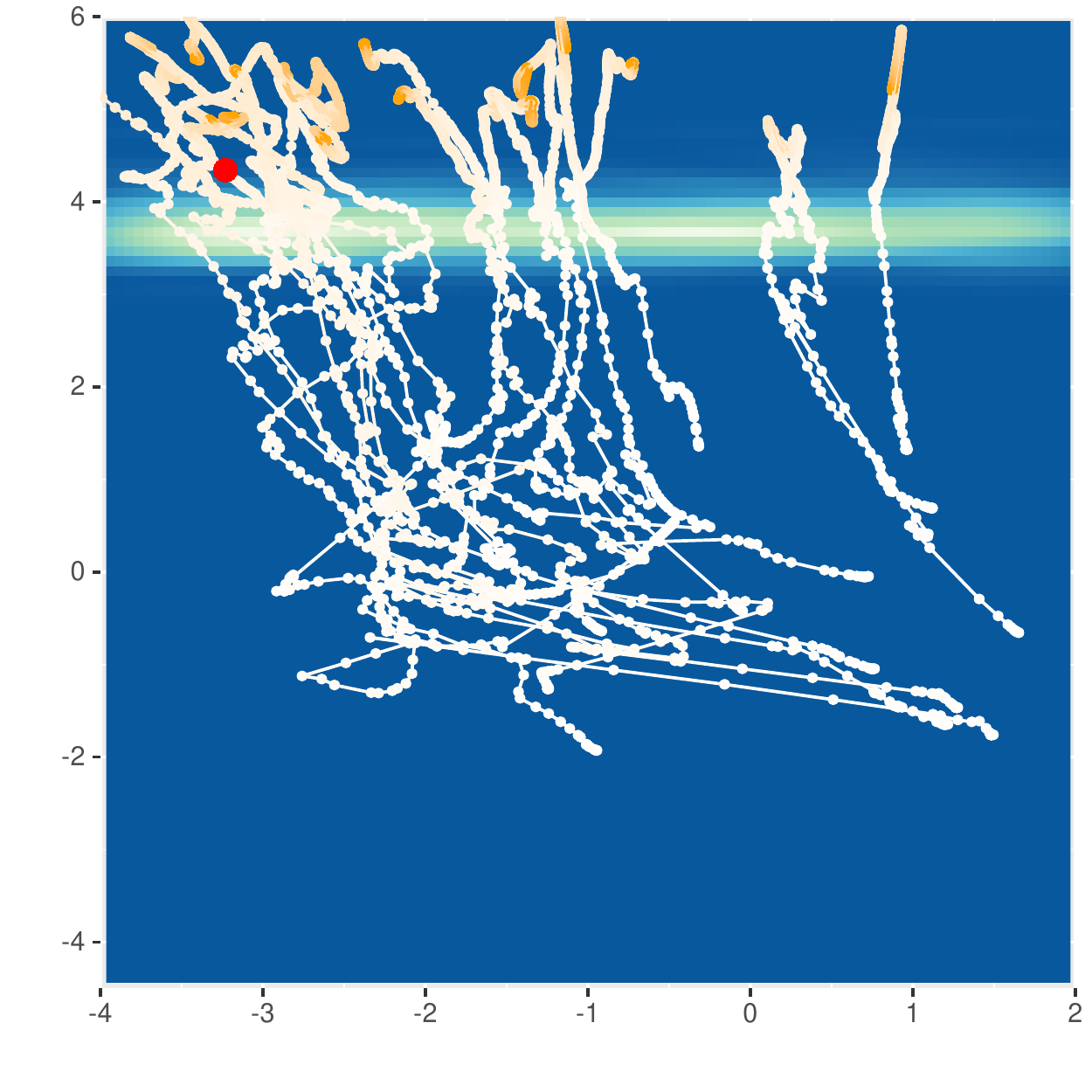}}
		\subfloat[compare expectation of $f$ of 133 observations \label{subfig: 32_12_prediction}]{\includegraphics[width=0.33\textwidth, keepaspectratio]{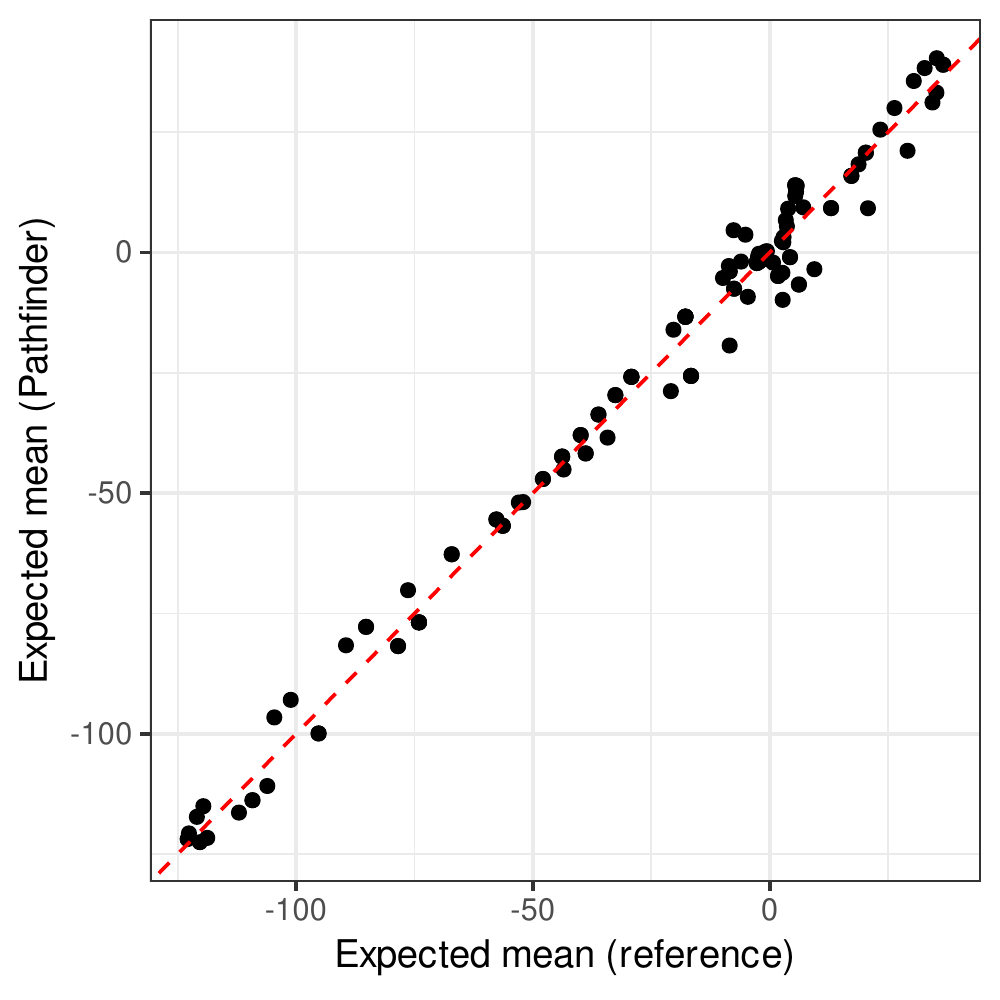}} 
	}
	\caption{\it Illustration of weakly identified posteriors for two parameters in the \texttt{mcycle\_gp-accel\_gp} example.  In (a), we see the result of Pathfinder with default initializations in $\textrm{uniform}(-2, 2)$; in (b), we zoom in the $x$-axis of plot (a).  The optimization paths are plotted from initialization in white to later iterations in orange.  In (c), we initialize parameters on the $x$ axis between $-40$ and $20$, resulting in L-BFGS declaring convergence at various points along the $x$ axis as a result of the weak identification. Due to the poor approximations from Pathfinder, the importance resampling step in multi-path Pathfinder only picks 4 distinct draws (recall that sampling is with replacement). Short runs of adaptive HMC perform better than Pathfinder (d), but are quite expensive in terms of density and gradient evaluations. Figure (e) shows that a large history size ($J = 100$ as opposed to our default $J = 5$) improves the local Hessian approximations in optimization. Figure (f) compares posterior expectations of parameters $\mu_f$, the mean of the Gaussian process for $f$, in the reference draws (horizontal axis) and draws from multi-path Pathfinder with history length of $J = 100$ (vertical axis). 
	}\label{fig: varying_curvature}
\end{figure}

\begin{figure}[t!]
	\centering{
		\hfill
		\subfloat[20 optimization paths\label{subfig: 32_tr}]{\includegraphics[width=0.33\textwidth, keepaspectratio]{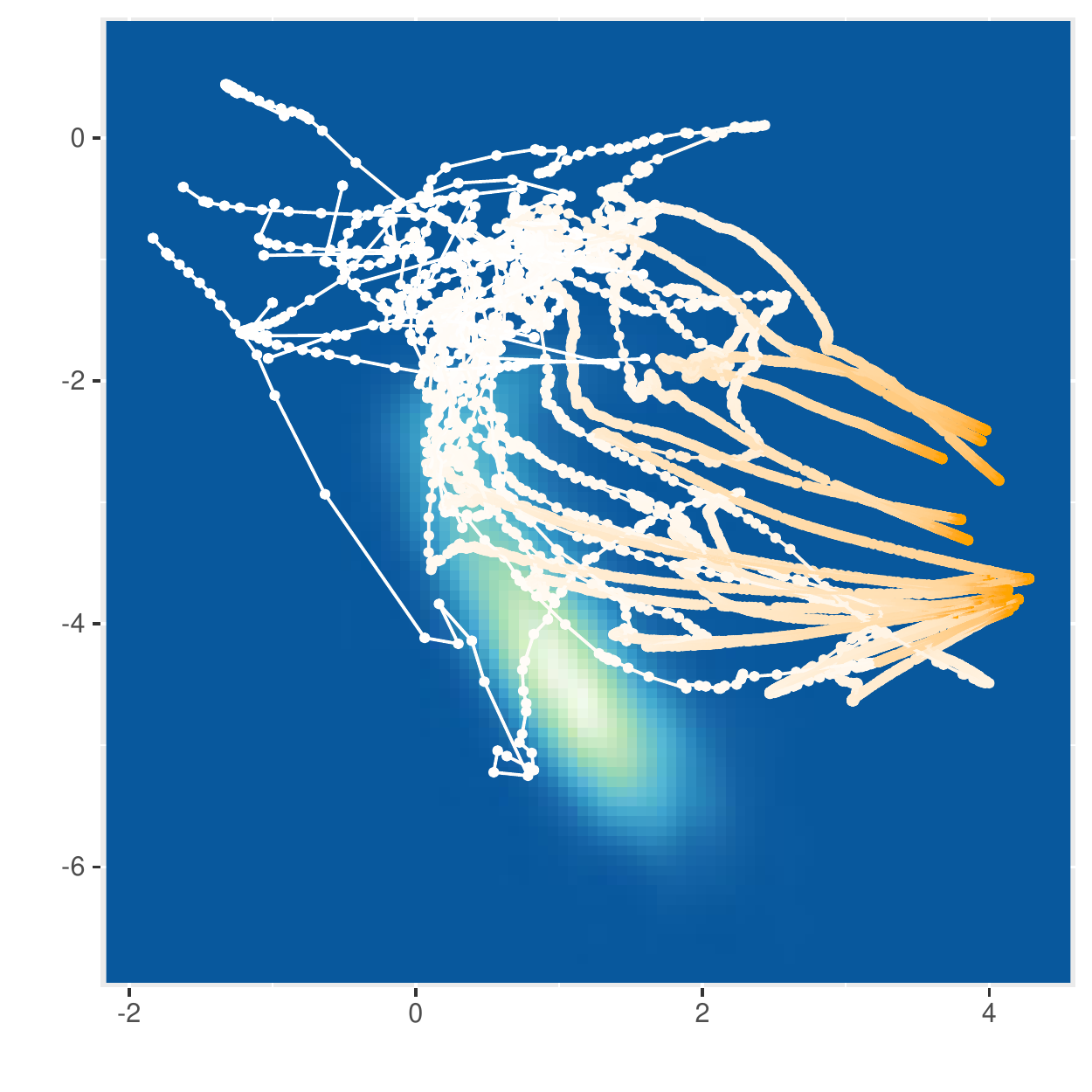}} 
		\hfill
		\subfloat[100 approximate draws from multi-path Pathfinder \label{subfig: 32_points}]{\includegraphics[width=0.33\textwidth, keepaspectratio]{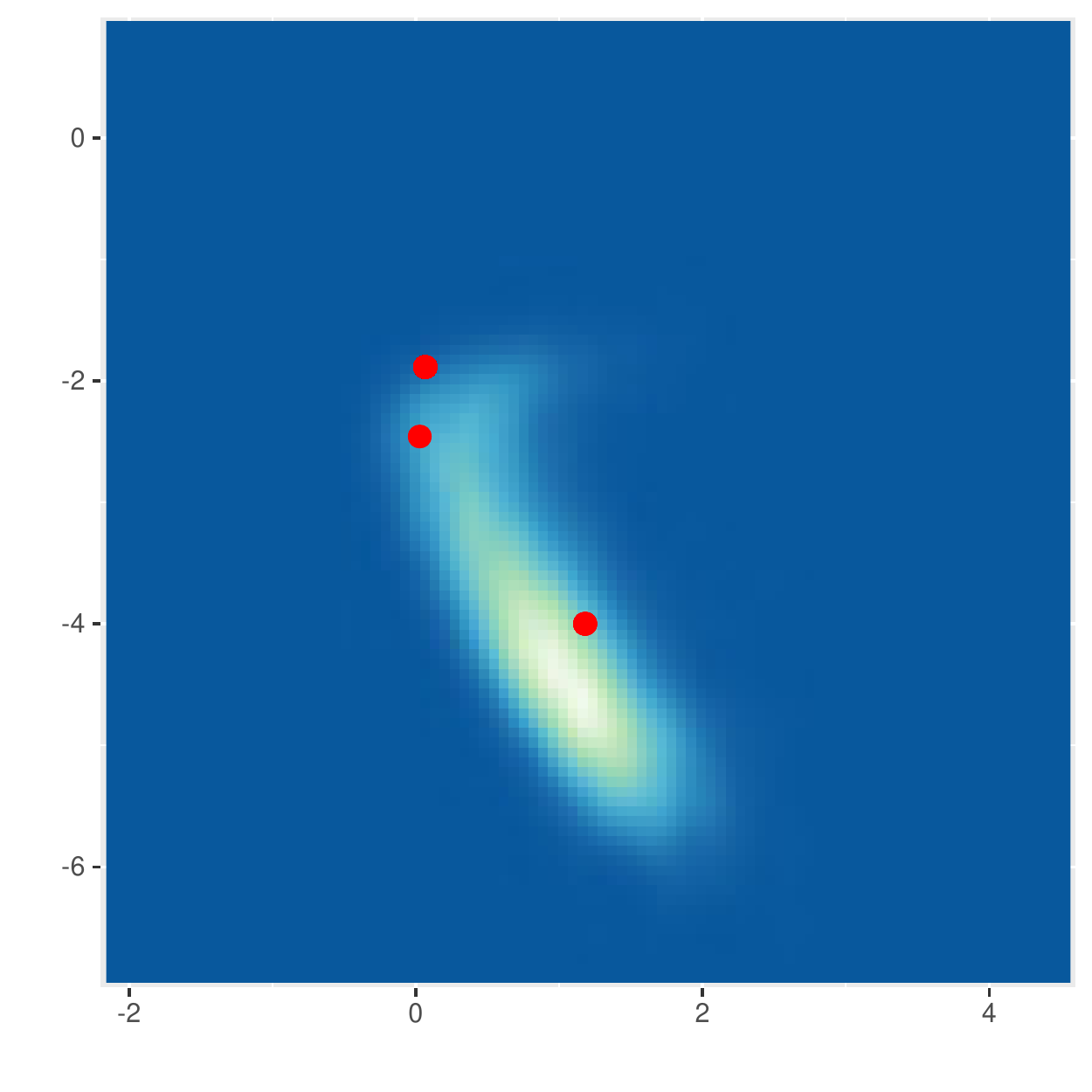}}
		\hfill
	}
	\caption{\it Illustration of non-normality in two additional dimensions of the \texttt{mcycle\_gp-accel\_gp} example. The 20 optimization paths are shown in (a) with later iterations in orange.  The 100 approximate draws selected by multi-path Pathfinder are shown in (b) as red dots overlaid on the target density. The target density is clearly not normal in these two dimensions.  Due to the poor approximations from Pathfinder, importance resampling with replacement in multi-path Pathfinder only generates 4 distinct draws among the 100 approximate draws it chooses because of the high variance of the importance weights in this example.}\label{fig: varying_curvature_2}
\end{figure}

\paragraph{Weak identifiability.} Unlike multimodal priors, which have multiple modes, non-identifiable posteriors do not have any modes. For example, a $\textrm{uniform}(0, 1)$ distribution is flat and does not have a mode. Flat regions in densities must be compact, because flatness in unbounded spaces leads to unnormalizable distributions. In Stan and this paper, we work with unconstrained distributions, where a variable $\alpha$ with a $\textrm{uniform}(0, 1)$ distribution is transformed to $\textrm{logit}(\alpha) = \log \frac{\alpha}{1 - \alpha}$, which has a $\textrm{logistic}(0, 1)$ distribution.  Although we cannot have properly flat unbounded posteriors, they can be nearly flat over large regions, which can cause the same computational problems. For example, we see such behavior with collinear predictors in regressions, which produces high posterior variability and high correlation between their regression coefficients. Such flat regions correspond to plateaus in the density and cause difficulties with convergence for adaptive HMC, ADVI, and Pathfinder.

We observe pathological flatness in the posterior of model \verb|mcycle_gp-accel_gp| from \texttt{posteriordb}, which is one of the most expensive models in \texttt{posteriordb} according to the cost evaluation presented in Figure~\ref{fig: cost_lp_compare}. This example models measurements of head acceleration in a simulated motorcycle accident.  Specifically, the observation $y$ is modeled with a normal distribution having Gaussian process prior on mean function $f$ and log standard deviation function $g$,
\[
y \sim \normal(f, \exp(g)), \; f \sim \mbox{GP}(\mu_f, K_1), \; g \sim \mbox{GP}(\mu_g, K_2),
\]
where $K_1$ and $K_2$ denote covariance functions, and $\mu_f$ and $\mu_g$ model the mean of the priors of $f$ and $g$. Both Gaussian processes $f$ and $g$ are modeled with Hilbert-space approximate basis functions  \citep{solin2020hilbert,riutort2020practical}. There are in total 66 parameters. In Figure~\ref{fig: varying_curvature}, we illustrate 100 draws from multi-path Pathfinder, which are based on resampling 20 optimization paths, for two selected dimensions of \verb|mcycle_gp-accel_gp|. Figures~\ref{subfig: 32_12}~and~\ref{subfig: 32_12_s} show that the optimization paths are almost parallel to each other when passing the high probability region. The gradients of the log density along the $x$-axis fail to guide optimization paths toward the high probability region of the posterior. Figure~\ref{subfig: 32_12_2} illustrates that with specialized initials, the positions of the maxima found by L-BFGS span from $-40$ to $20$ on the $x$-axis, a strong sign of weak posterior identifiability. Since Pathfinder relies on estimating curvature along optimization paths to estimate the covariance of normal approximations, Pathfinder fails to provide good covariance estimates for normal approximations centered near the high probability mass region in this example. 

Apart from weakly identified parameters, we find that several parameters exhibit correlation in the posterior. Therefore, L-BFGS with a small history size ($J$), which uses a low-rank plus diagonal matrix to approximate local Hessian, cannot precisely capture local curvature information.  Figure~\ref{subfig: 32_12_100h} shows how a large history size can improve the performance of Pathfinder in this example. Pathfinder successfully identifies the high probability mass region among the optimization paths, though the approximate draws are concentrated in a small region due to the weak identifiability problem. Figure~\ref{subfig: 32_12_prediction} compares the expectation of $f$ for 133 observations estimated by reference samples and approximate samples from multi-path Pathfinder with $J = 100$.   We conclude that approximate inference from Pathfinder is reasonable in this case because expectations are close between Pathfinder and the reference draws as indicated by the 45-degree line where $y = x$.

In addition to weakly identified and correlated parameters, \verb|mcycle_gp-accel_gp| has other features that frustrate Pathfinder. In Figure~\ref{fig: varying_curvature_2}, we see that the high probability region in the other two dimensions is shaped like a boomerang. Because the posterior distribution is far from normal, none of the parametric VI algorithms find good approximations. The approximate draws by multi-path Pathfinder on the two selected dimensions are shown in Figure~\ref{subfig: 32_points}. 
Although Pathfinder fails to provide a good normal approximation, the draws from Pathfinder are close to the target region for most of the parameters in this example.

For difficult high-dimensional examples such as the Gaussian process \verb|mcycle_gp-accel_gp| and the hidden Markov model \verb|bball_drive_event_0-hmm_drive_0|, the approximate draws are concentrated in a small region, but they can be useful starting points for more elaborate and more costly algorithms.  


\section{Using Pathfinder to initialize MCMC}\label{sec: Birthday}


Markov chain Monte Carlo methods, including adaptive Hamiltonian Monte Carlo, are typically initialized randomly and then allowed to evolve until multiple chains have ``converged.''  This informal notion of convergence can mean several things.  In the strictest sense, it requires running the chains long enough to fully forget their initial positions, as measured, for example, by coupling \citep{jacob2020unbiased}.  Such forgetting is necessary to avoid estimation bias due to the initialization.  A less strict criterion only requires adequate mixing of the chains, which may be measured through within- and between-chain variances \citep{gelman1992inference,vehtari2021rank}. The most generous notion of convergence provides the modest goal of using Pathfinder for initialization, namely finding a point, or several points, that look as if they might have been drawn from the posterior. This is the point at which the transient bias due to not initializing with a draw from the target is mostly eliminated and estimation error starts to become dominated by sampling error \citep[Section~2.2.4]{angelino2016patterns}.  


Initialization to remove most of the transient bias of MCMC can succeed by producing draws within the high probability volume of the posterior without covering that posterior---it only requires draws to look reasonable. Besides, even if initial points are underdispersed, if they are close enough to the high probability mass region, the total variance would still likely increase faster than the within chain variance. As there will be additional warmup after Pathfinder, the probability of having high between-chain variance, would be negligible. Therefore, the convergence diagnostic based on within- and between-chain variances is still reliable with underdispersed initials in high probability mass region. We can diagnose problems with Pathfinder's distributional approximation by using the Pareto-$k$ statistic  \citep{yao2018yes}. In such cases, Pathfinder can still be valuable as a starting distribution for the adaptation phase of MCMC, improving current practice whereby adaptation can be slow when starting points are not chosen well. 


\paragraph{The birthday problem Gaussian process.} 
We consider the challenging problem of modeling the time series of the number of births by day, where the default no-U-turn Hamiltonian Monte Carlo sampler in Stan tends to get stuck in local minor modes. This time series shows periodic trends at weekly and annual scales, as well as longer-term trends and date- and holiday-based behavior. This is an attractive example for Bayesian inference and computation because national-level data are publicly available, sample sizes are large enough that fine-grained patterns can be detected with careful analysis, the underlying scientific questions are accessible and of general interest, and the repetition at different time scales is not exact (for example, there is a general pattern of more births in the summer than in the winter, but that contrast changes over time and varies by country).  Together, these features motivate the use of Gaussian processes, a class of models that support flexible patterns of ad hoc, trend-based, and periodic patterns at different time scales.

We follow \cite{gelman2013bda} in analyzing the number of births per day in the United States from 1969 through 1988 with Gaussian processes (GP). Applying GPs at this scale is challenging because of the need to work with an $N \times N$ covariance matrix, where $N$ is the number of days in the series. The generic form of the likelihood is
\begin{equation}
	y_n \sim \textrm{normal}(f(x_n), \sigma),
\end{equation}
where $y$ is the time series of logarithm of number of births by day. (The counts are large enough that we do not need to worry about their discreteness.) The function $f$ is assigned a Gaussian process prior, which takes into account
a global trend $f_1$, yearly seasonal trend $f_2$,  the effect of weekdays $\beta_{\text{day of week}}$, and the day of year effect $\beta_{\text{day of year}}$,
\begin{align}
	&f = \alpha + f_1 + f_2 + \beta_{\text{day of week}} + \beta_{\text{day of year}}, \\
	&\alpha \sim \textrm{normal}(0, 1),
	f_1 \sim \textrm{GP}(0, K_1), \;f_2 \sim \textrm{GP}(0, K_2), \nonumber \\
	&\beta_{\text{day of week}} = 0 \; \text{ if day of week is Monday}, \nonumber \\
	&\beta_{\text{day of week}} \sim \textrm{normal}(0, 1) \text{ if day of week is not Monday, and} \nonumber \\
	&\beta_{\text{day of year}} \sim \text{normal}(0, 0.1), \nonumber 
\end{align}
where the first GP uses the exponentiated quadratic covariance function $K_1$ and the second a periodic covariance function $K_2$. Most years have 365 calendar days and every four years (during the data range) there are 366 day, and thus we simplify and use period of 365.25 for the periodic component. The Gaussian process priors for $f_1$ and $f_2$ are modeled by the Hilbert space approximate basis function approximation of Gaussian processes \citep{solin2020hilbert,riutort2020practical}. This model includes a total of $429$ parameters.

\begin{figure}[t!]
	\centering{
		\subfloat[][draws from Stan phase I 
		
		(1-W = 9.65) \label{subfig: birthday_pt}]{\includegraphics[width=0.25\textwidth, keepaspectratio]{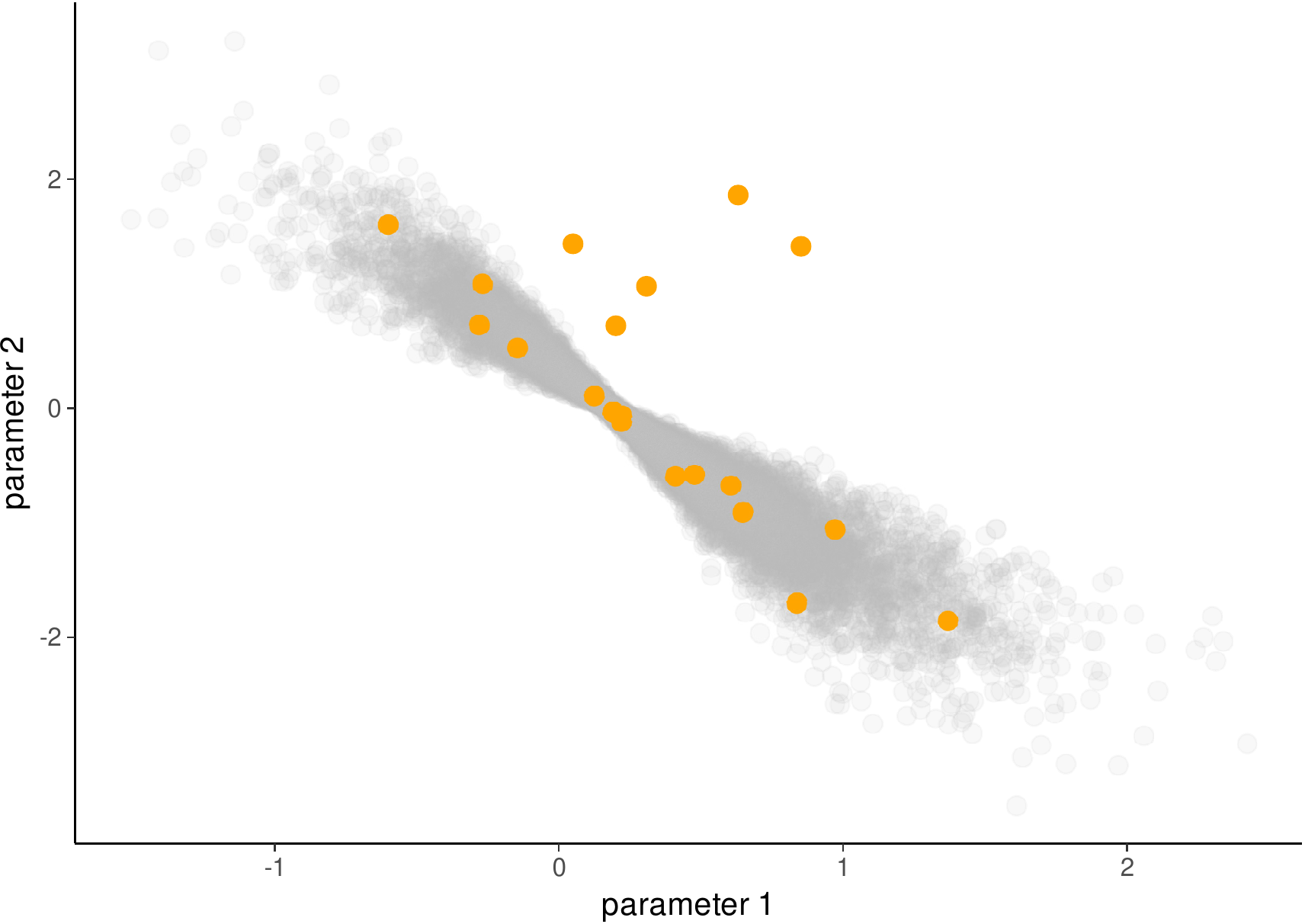}}
		\hfill
		\subfloat[][draws from multi-path Pathfinder 
		(1-W = 5.45) \label{subfig: birthday_pt_pf2} ]{\includegraphics[width=0.25\textwidth, keepaspectratio]{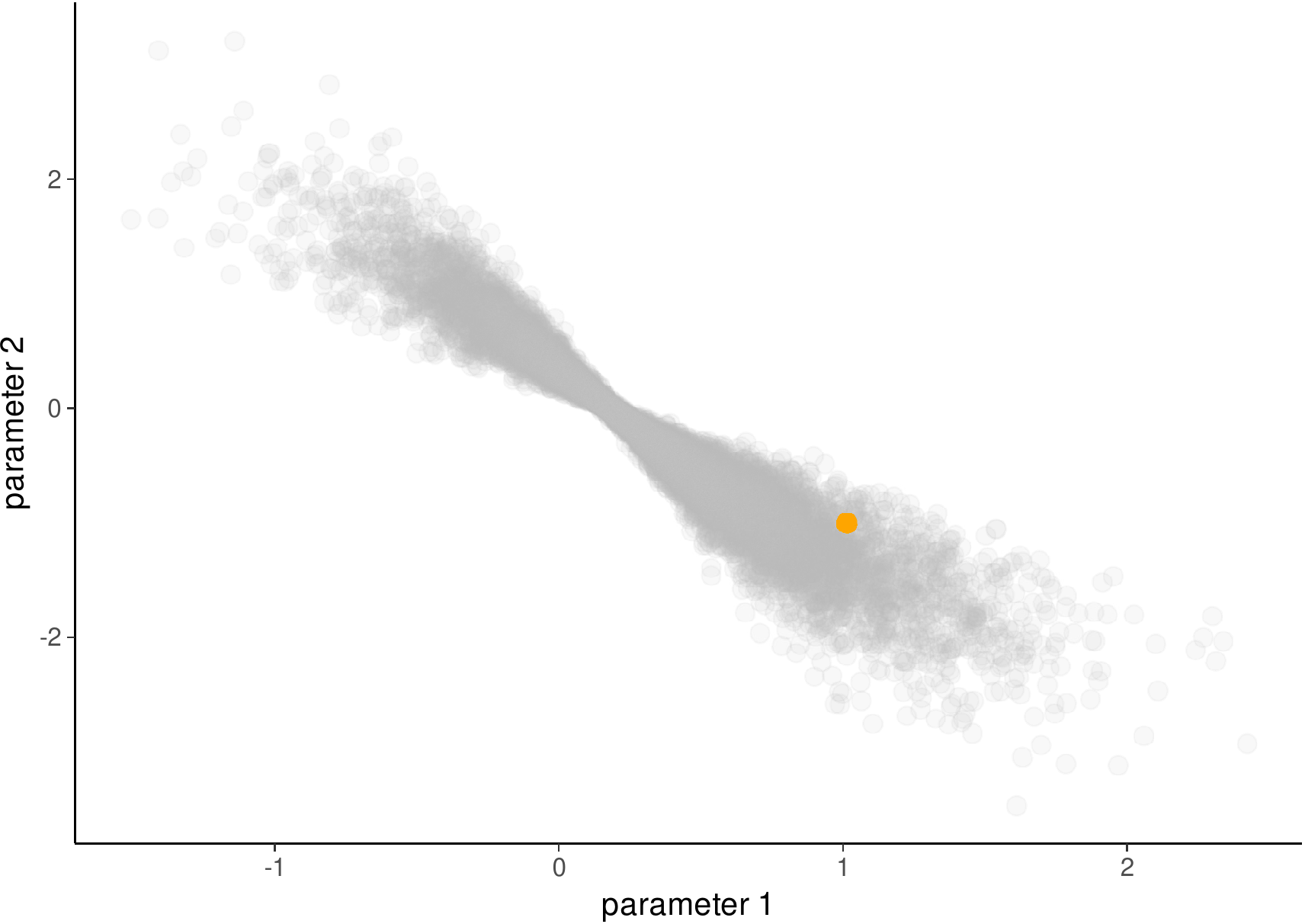}}
		\hfill
		\subfloat[][draws from Pathfinder + Stan phase I 
		(1-W = 5.32) \label{subfig: birthday_pt_pf} ]{\includegraphics[width=0.25\textwidth, keepaspectratio]{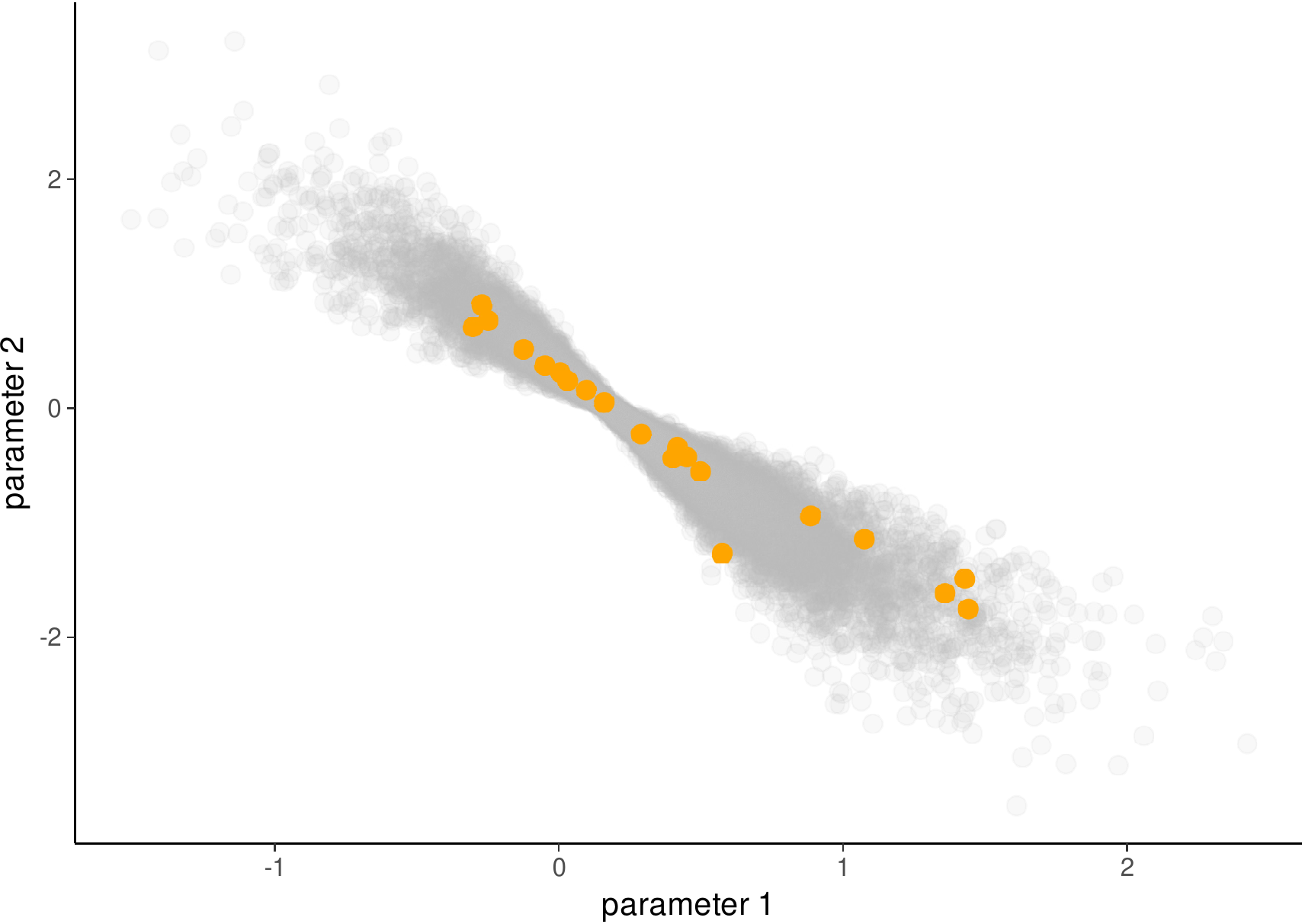}}
		\\
		\hfill
		\subfloat[ Stan phase I traceplots with random inits from $\textrm{uniform}(-2, 2)$ \label{subfig: birthday_phI}]{\includegraphics[width=0.3\textwidth, keepaspectratio]{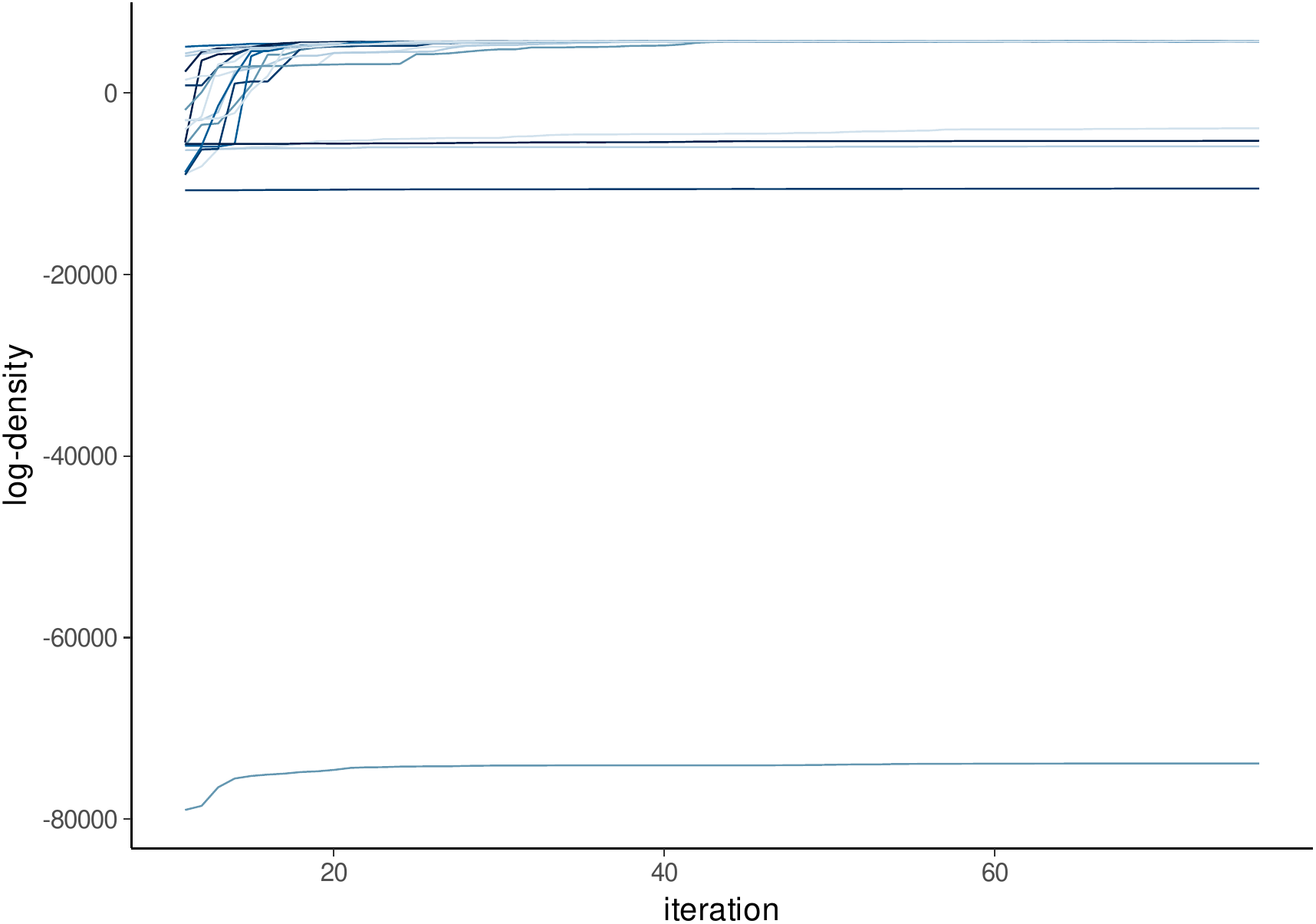}}
		\hfill
		\subfloat[ Stan phase I traceplots with Pathfinder inits \label{subfig: birthday_phI_pf} ]{\includegraphics[width=0.3\textwidth, keepaspectratio]{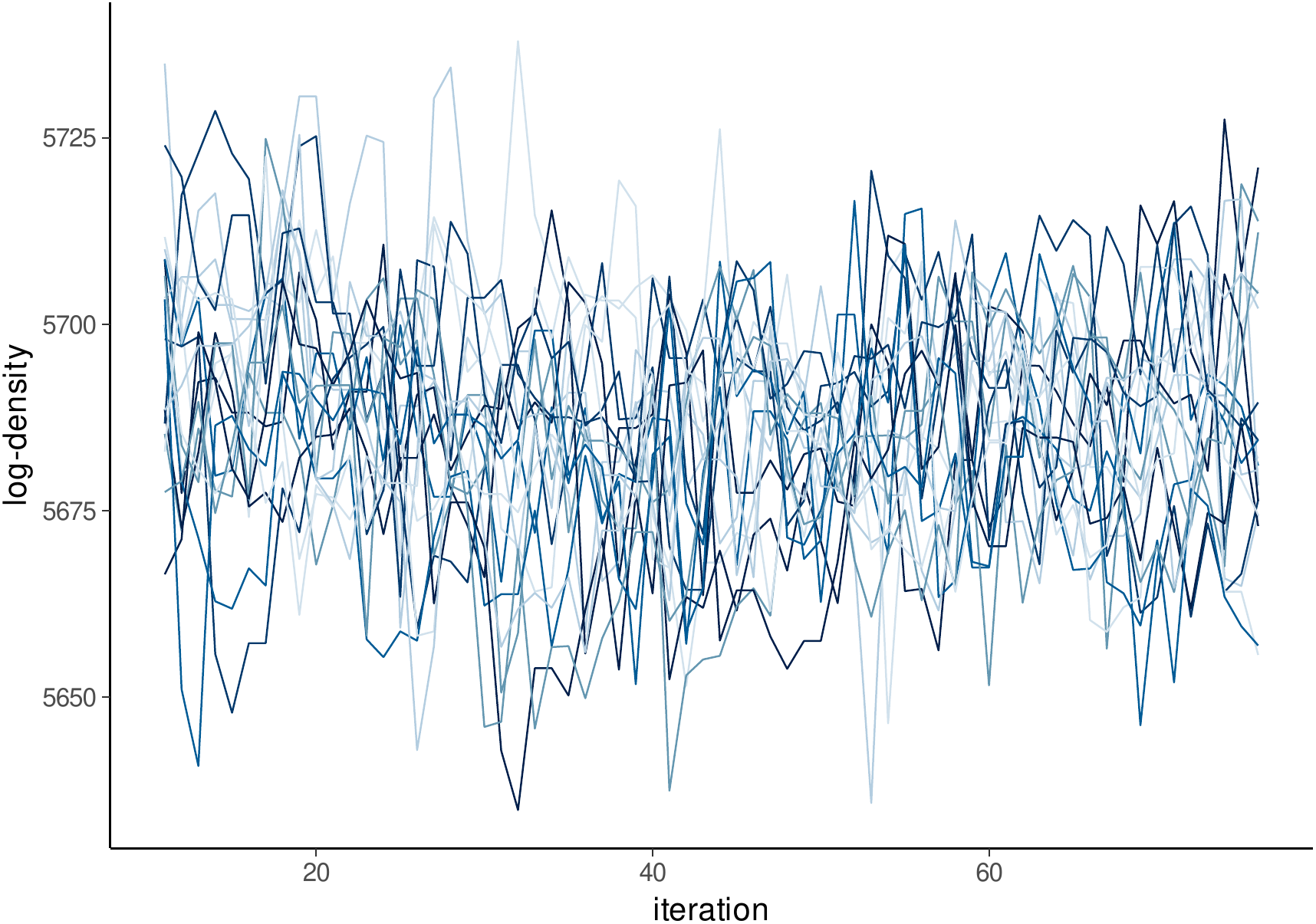}}
		\hfill \null
		\\
		\subfloat[ expected births, reference vs.\ final Stan phase I \label{subfig: birthday_pred_phI}]{\includegraphics[width=0.25\textwidth, keepaspectratio]{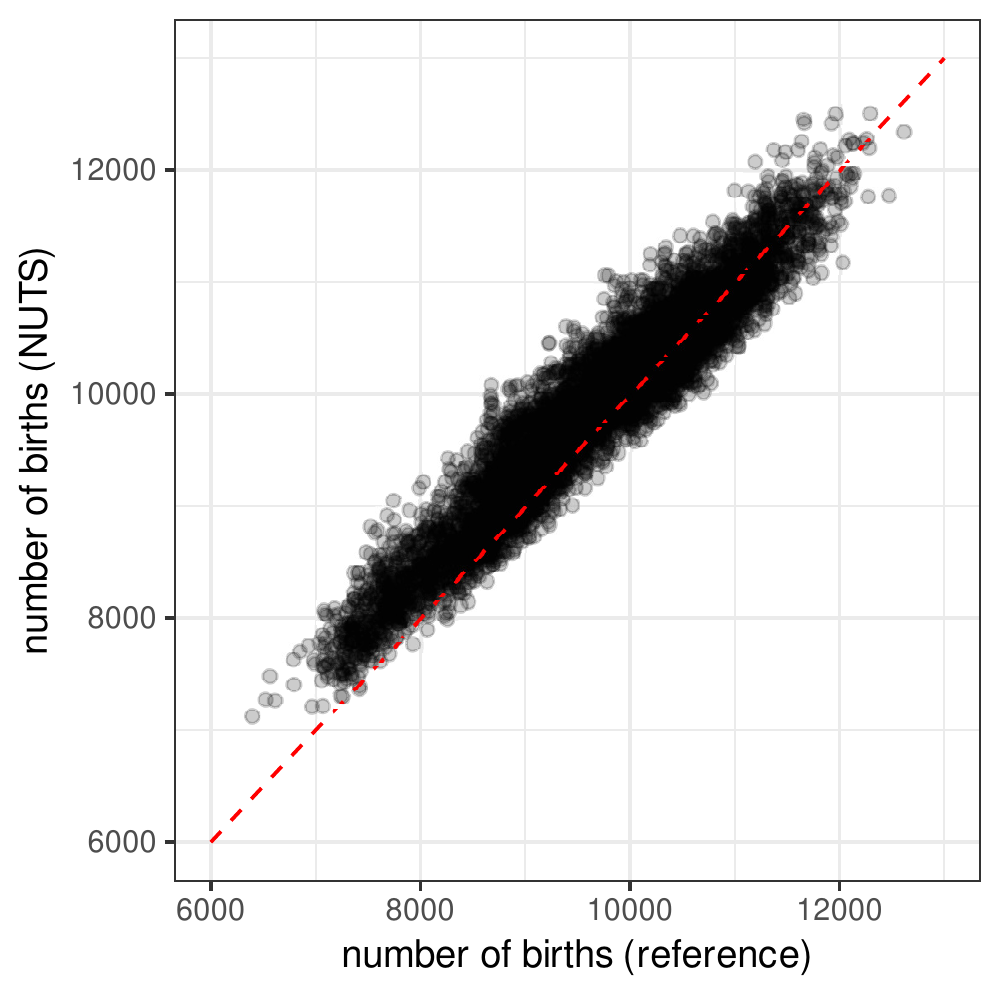}}
		\hfill
		\subfloat[expected births, reference vs.\ multi-path Pathfinder \label{subfig: birthday_pred_pf} ]{\includegraphics[width=0.25\textwidth, keepaspectratio]{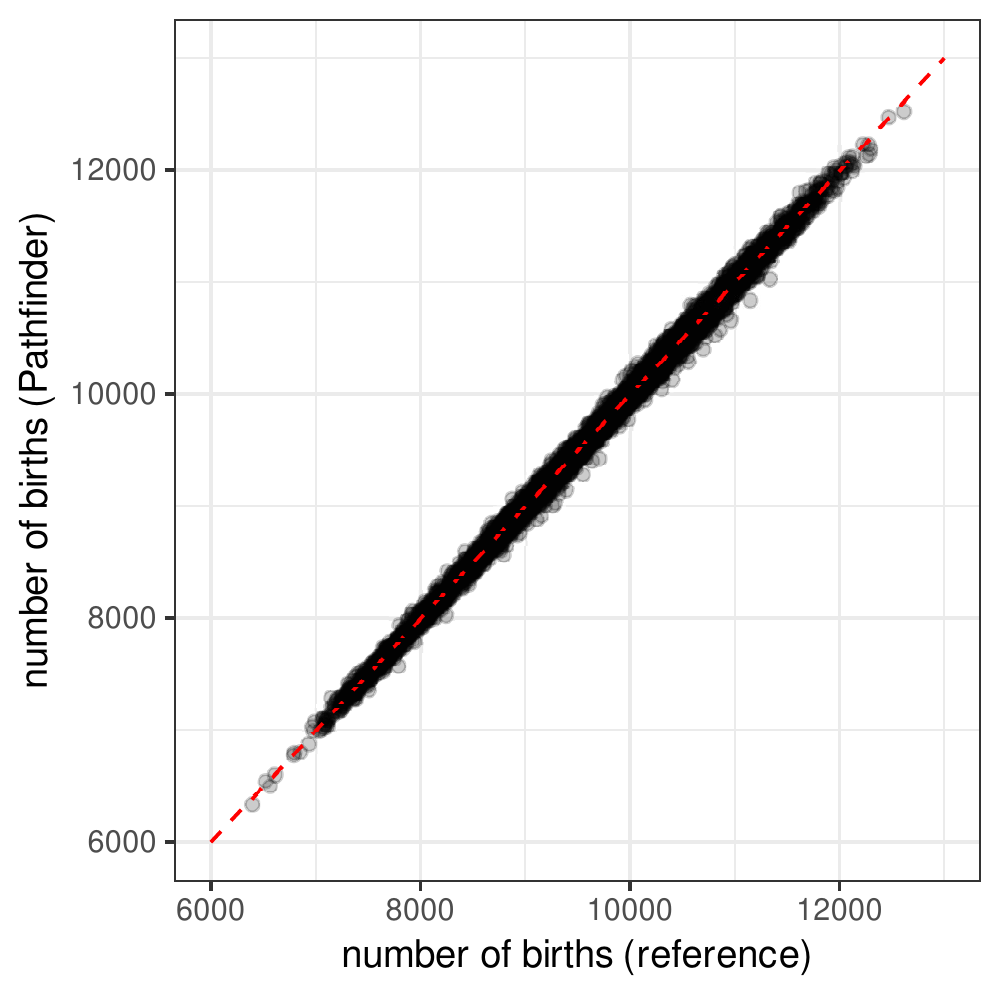}}
		\hfill
		\subfloat[ expected births, reference vs. Pathfinder + Stan phase I 
		\label{subfig: birthday_pred_pf_hmc} ]{\includegraphics[width=0.25\textwidth, keepaspectratio]{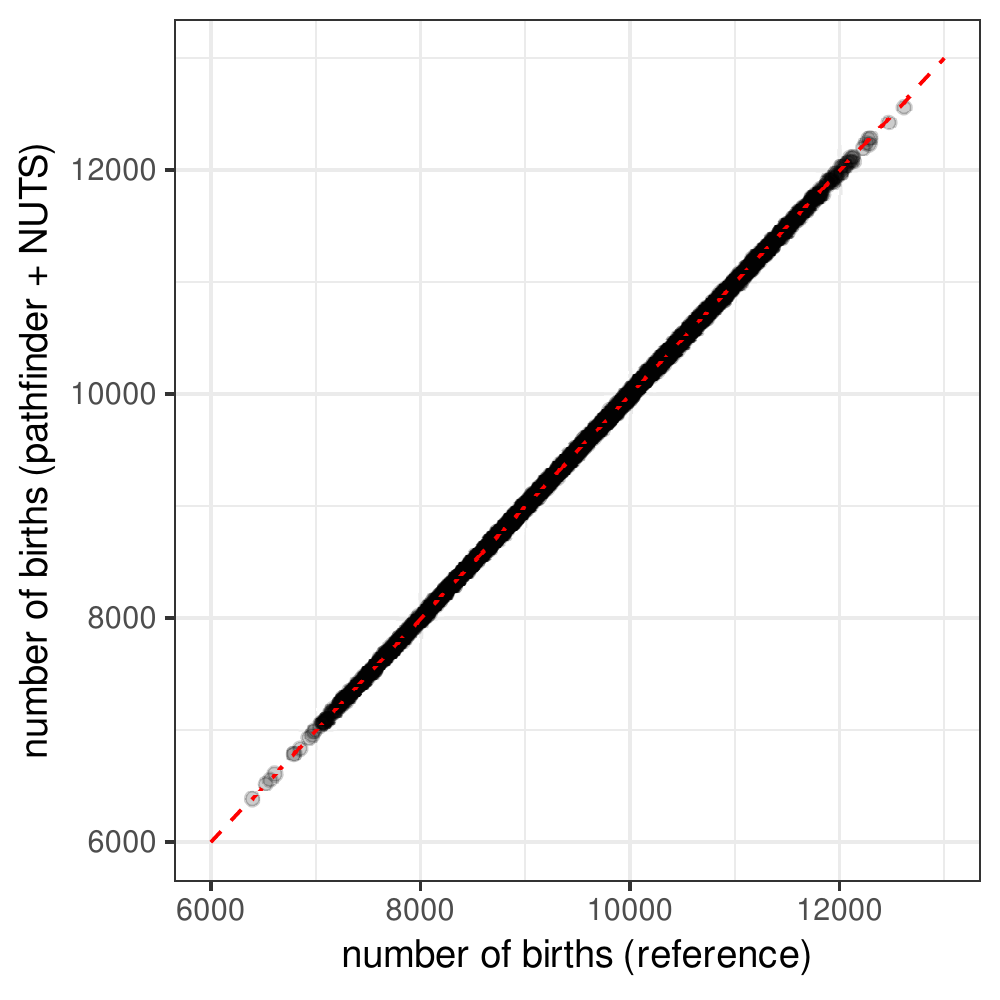}}}
	\caption{\it 
		Comparison of 75 iterations of adaptive HMC (Stan Phase I adaptation), multi-path Pathfinder, and HMC initialized by multi-path Pathfinder for the birthday Gaussian process. Top row: draws in orange against the target density of two selected parameters for (a) draws from 20 adaptive HMC chains initialized with $\textrm{uniform}(-2, 2)$, (b) 20 approximate draws from multi-path Pathfinder, and (c) draws from 20 adaptive HMC chains initialized by multi-path Pathfinder. Draws from multi-path Pathfinder are closer to the target than 75 iterations of adaptive HMC.  Middle row: traceplots of 75 adaptive HMC iterations initialized randomly (d) and with Pathfinder (e). Bottom row: (f) expected number of births by day estimated by reference samples and draws from adaptive HMC, (g) multi-path Pathfinder, and (h) adaptive HMC initialized by multi-path Pathfinder. Pathfinder-initialized HMC provides inference closest to the reference.}\label{fig: birthday}
\end{figure}
To evaluate the performance of Pathfinder, we compare the approximate draws obtained by adaptive Hamiltonian Monte Caro (Stan's phase I sampler), multi-path Pathfinder, and adaptive HMC initialized with draws from multi-path Pathfinder. For multi-path Pathfinder, we generate $20$ approximate draws using $20$ optimization paths and the default setting of tuning parameters. For Stan's phase I sampler, we run 20 adaptive HMC chains with 75 iterations for each, and take the samples at the last iteration as the approximate draws. The trace plots of log density plotted in Figure~\ref{fig: birthday} indicate multiple modes in the posterior distribution. Initializing an MCMC algorithm with Pathfinder avoids wasting computation time sampling within minor modes with negligible posterior mass. To compute a reference for comparison, we ran 4 MCMC chains with an adaptation period of 50,000 iterations, 300,000 saved iterations, and a thinning rate of 100 using \texttt{cmdstanr}. In Figure~\ref{fig: birthday}, we illustrate reference posterior draws and approximate draws of three competitors 
on two selected dimensions. The reference posterior has the shape of an hourglass, which is found by all draws from Pathfinder-initialized adaptive HMC. The 1-Wasserstein distance between the reference posterior draws and draws after 75 iterations of adaptive HMC as well as the comparison of the estimated number of births highlight the benefits of initializing MCMC with draws from Pathfinder.

Figure~\ref{subfig: birthday_pt} show that 5 of the 20 randomly initialized adaptive HMC chains failed to reach the high probability region, while the remaining chains provided reasonable approximate draws.  Figure~\ref{subfig: birthday_phI} shows the 5 non-converging MCMC chains were trapped in minor modes. Convergence diagnostic results are consistent with multimodality. {Therefore, initializing MCMC chains with Pathfinder helps to avoid the local minor modes in the subsequent MCMC sampling. 
	On average, the number of gradient evaluations of an adaptive HMC chain in this example is around 10 times that of single-path Pathfinder.  Pathfinder further requires around 5 times more log-density evaluations than gradient evaluations to estimate ELBOs and generate approximate draws. With our experimental Pathfinder implementation in \texttt{cmdstanr}, the average running time of a single-path Pathfinder is around one-third of that of a single 75 iteration HMC chain (3s vs 9s). In summary, initializing MCMC with Pathfinder can improve the overall sampling efficiency.} 

\section{Discussion}\label{sec: discussion}

The Pathfinder algorithm we present in this paper is similar to automatic differentiation variational inference \citep{kucukelbir2017automatic}, with two key differences.  First, instead of performing stochastic gradient descent directly on the evidence lower bound, we use quasi-Newton methods to directly optimize the objective function.  Second, rather than a diagonal or dense covariance approximation, Pathfinder uses a low-rank plus diagonal factorization that is more expressive than a simple diagonal covariance matrix, but nearly as cheap to employ.  Direct optimization is much more stable because we can use cheap L-BFGS estimates of inverse Hessians to locally condition our optimization steps \citep{nocedal1980updating}.  Pathfinder can evaluate the ELBO of normal approximations built upon L-BFGS estimates of inverse Hessians along this path (in parallel), and we know from the intermediate value theorem that if starting from the tail of the distribution, the optimization trajectory will move from the tail, through the body of the distribution, to the mode or pole.  In our examples, Pathfinder requires one to two orders of magnitude fewer log density and gradient evaluations than using automatic differentiation variational inference or using dynamic Hamiltonian Monte Carlo to warm up.  Pathfinder provides approximations slightly inferior to that of adaptive step-size Hamiltonian Monte Carlo, but better than mean-field or dense ADVI as measured by 1-Wasserstein distance.  

Multi-path Pathfinder improves on the robustness of single-path Pathfinder by running multiple single-path instances, then importance resampling.  We have shown that this can both filter out minor modes where the algorithm might otherwise get stuck, and improve fidelity in representing irregular posteriors.  In this usage for ``black-box'' variational inference, Pathfinder is similar to using multiple short MCMC paths for inference \citep{hoffman2020black}. In most cases, the approximation provided by multi-path Pathfinder is comparable to or better than that of short chains of Hamiltonian Monte Carlo. Although multi-path Pathfinder requires many log density and gradient evaluations, these can be parallelized, unlike the necessarily serial evaluation of Markov chain Monte Carlo. 

For posteriors with high computational cost, it might be useful to use Bayesian optimization \citep{shahriari2015taking} to find the ELBO maximizing point along the L-BFGS optimization trajectory. Such optimization could also consider parameter values between L-BFGS iterations, as L-BFGS might sometimes take a big step over the optimal point. Multi-path Pathfinder is more robust than single-path Pathfinder to the variability due to the small number of Monte Carlo draws used to evaluate the ELBO, because importance resampling will favor the best approximate distributions from multiple single-path runs.

The performance of Pathfinder can be sensitive to the initial values if they happen to be underdispersed and close to the mode. To make Pathfinder more robust to initial values, it would be possible to create the first approximation with any initial values and then use that approximation to generate additional initial values from a normal distribution with much larger scale than the scale of the first approximation (e.g., 10 times larger scale). For example, in the ovarian cancer example in Section~\ref{sec: ovarian}, this approach would be likely to find additional modes beyond the mode at the origin.

Pathfinder can be part of a more effective computational workflow, starting with fast multivariate optimization, moving to Pathfinder's distributional approximation, and then if necessary moving to fully stochastic MCMC algorithms. Even biased approximate inference can be useful if it can produce one reasonable draw quickly or several in parallel. If such draws are unreasonable, there is a good reason to believe the model is misspecified or has an error in its code.  This allows us to fail fast during model development, a perspective from software engineering \citep{shore2004fail} that we recommend applying to statistical workflow \citep{gelman2020bayesian, gabry2019visualization}.

Perhaps the most obvious extension to consider is using optimizers other than L-BFGS.  For example, it might be possible to improve scalability by subsampling long data sets and using stochastic gradient descent \citep{robbins1951stochastic} or one of its modern adaptive variants such as Adam \citep{kingma2014adam} on the objective (not on the evidence lower bound).  The overall algorithm should be tolerant to data subsampling because we do not need high tolerance for mode finding.  Subsampling may also be more robust in the face of minor modes (i.e., local optima).  

One of our goals in developing Pathfinder was to find a drop-in replacement to initialize the warmup phase of MCMC sampling.  For Gibbs samplers, that would only involve finding a reasonable starting point.   In contrast, adaptive Metropolis and Hamiltonian Monte Carlo samplers require us to generate an approximate covariance matrix for either a proposal distribution or to pre-condition Hamiltonians.  Thus finding initial values is only the first phase of warmup and adaptation.  In Stan, the second phase of warmup involves exploring the posterior to estimate the posterior covariance in either diagonal or dense form.  Following \cite{bales2019selecting}, we have verified that the estimated inverse Hessian from L-BFGS is a reasonable initalization for such adaptation, and may in fact be accurate enough to bypass Stan's phase II adaptation, at least for low-dimensional models where it is easier to estimate covariance.  This would remove the remaining serial processing bottleneck for efficient parallel MCMC.  In more difficult problems, it should be possible to use the information provided by Pathfinder to perform warmup adaptation more efficiently.

With a few dozen test models of fairly low dimension and a handful of higher-dimensional examples, we have only scratched the surface of potential applications.  We believe the use of local curvature information in L-BFGS should help with fitting in both higher dimensions and in situations with poor conditioning locally due to varying curvature, compared to either optimizing the ELBO directly using stochastic gradient methods or subsampling short chains of Hamiltonian Monte Carlo.



\section*{Acknowledgements}
We thank Dan Simpson, Ben Bales, Colin Carroll, and Philip Greengard for helpful comments and Matt Hoffman for motivation, discussion, and clarification.  Andrew Gelman and Lu Zhang thank the U.S. National Science Foundation (grant 2055251), National Institutes of Health (grant 1R01AG06714901), Office of Naval Research (grant N000141912204), Institute for Education Sciences (grant R305D190048), Alfred P. Sloan Foundation (grant G201912491), and Schmidt Futures for partial support of this work. Aki Vehtari thanks the Academy of Finland Flagship programme: Finnish Center for Artificial Intelligence, FCAI, for partial support of this work.

\vskip 0.2in
\bibliographystyle{ba}  
\bibliography{references} 

\appendix

\section{Derivation of (\ref{eq: Hk_factor})}\label{appendix: Hk_factor_deriv}
From \eqref{eq: Hk_reformat}, we have
\begin{equation*}
	\begin{aligned}
		\samp{\Sigma}{l} &= 
		\textrm{diag}(\alpha^{\frac{1}{2}}) \left( \idenI + \underbrace{\textrm{diag}(\alpha^{-\frac{1}{2}}) \cdot \beta}_{{Q} \cdot \tilde{R}} \cdot \gamma \cdot \beta^{\top} \cdot \textrm{diag}(\alpha^{-\frac{1}{2}}) \right) \textrm{diag}(\alpha^{\frac{1}{2}})\\
		&= 
		\textrm{diag}(\alpha^{\frac{1}{2}}) \left( \idenI + {Q} \cdot \tilde{R} \cdot \gamma \cdot \tilde{R}^\top \cdot {Q}^\top \right) \textrm{diag}(\alpha^{\frac{1}{2}}) \\
		&= 
		\textrm{diag}(\alpha^{\frac{1}{2}}) \left( \idenI + 
		\begin{bmatrix} {Q} & {P} \end{bmatrix} \begin{bmatrix}
			\tilde{R} \cdot \gamma \cdot \tilde{R}^\top & 0\\
			0 & 0 \end{bmatrix}  \begin{bmatrix} {Q}^\top \\ {P}^\top \end{bmatrix} \right) \textrm{diag}(\alpha^{\frac{1}{2}}) \\
		&= \textrm{diag}(\alpha^{\frac{1}{2}}) \left( 
		\underbrace{\begin{bmatrix} {Q} & {P} \end{bmatrix} \begin{bmatrix}
				\idenI & 0\\
				0 & \idenI \end{bmatrix}  \begin{bmatrix} {Q}^\top \\ {P}^\top \end{bmatrix}}_{\idenI} + 
		\begin{bmatrix} {Q} & {P} \end{bmatrix} \begin{bmatrix}
			\tilde{R} \cdot \gamma \cdot \tilde{R}^\top & 0\\
			0 & 0 \end{bmatrix}  \begin{bmatrix} {Q}^\top \\ {P}^\top \end{bmatrix} \right) \textrm{diag}(\alpha^{\frac{1}{2}})\\
		&= \textrm{diag}(\alpha^{\frac{1}{2}})
		\begin{bmatrix} {Q} & {P} \end{bmatrix} \begin{bmatrix}
			\idenI + \tilde{R} \cdot \gamma \cdot \tilde{R}^\top  & 0\\
			0 & \idenI \end{bmatrix}  \begin{bmatrix} {Q}^\top \\ {P}^\top \end{bmatrix}  \textrm{diag}(\alpha^{\frac{1}{2}}) \qquad \qquad
		\text{\small [because $\idenI +\tilde{R} \cdot \gamma \cdot \tilde{R}^\top = \tilde{L} \tilde{L}^\top$]} \\
		&= \textrm{diag}(\alpha^{\frac{1}{2}}) 
		\begin{bmatrix} {Q} & {P} \end{bmatrix} \begin{bmatrix}
			\tilde{L} \tilde{L}^\top & 0\\
			0 & \idenI \end{bmatrix}  \begin{bmatrix} {Q}^\top \\ {P}^\top \end{bmatrix} \textrm{diag}(\alpha^{\frac{1}{2}})\\
		&= \textrm{diag}(\alpha^{\frac{1}{2}}) 
		\begin{bmatrix} {Q} & {P} \end{bmatrix} \begin{bmatrix}
			\tilde{L} & 0\\
			0 & \idenI \end{bmatrix}  \begin{bmatrix}
			\tilde{L}^\top & 0\\
			0 & \idenI \end{bmatrix}  \begin{bmatrix} {Q}^\top \\ {P}^\top \end{bmatrix} \textrm{diag}(\alpha^{\frac{1}{2}}) \\
		&= \underbrace{\textrm{diag}(\alpha^{\frac{1}{2}}) 
			\begin{bmatrix} {Q}\tilde{L} & {P} \end{bmatrix}}_{T}
		\begin{bmatrix} \tilde{L}^\top{Q}^\top \\ {P}^\top \end{bmatrix} \textrm{diag}(\alpha^{\frac{1}{2}}).
	\end{aligned}\;
\end{equation*}

\section{1-Wasserstein distance}\label{appendix: 1_wasserstein}
The 1-Wasserstein distance between two probability measures is based on the amount of effort it would take to rearrange one probability measure to look like the other, where effort is measured according to the distance mass is moved \citep{Craig_2016, Villani_2009, McCann_1995}.  If each distribution is viewed as a unit volume of earth piled according to the density, then the 1-Wasserstein distance is the minimum cost of turning one pile into the other, with ``cost'' defined as the amount of earth that needs to be moved times the mean distance it has to be moved. Unlike KL divergence, Wasserstein distance is a proper distance metric in that it is symmetric, all distances are non-negative, it obeys the triangle inequality because mass can be moved in two steps, and distance between a distribution and itself is zero, because no probability mass needs to be adjusted.  

\begin{figure}[t!]
	\centering{
		\subfloat[1-W distance = 0.034]{\includegraphics[width=0.33\textwidth, keepaspectratio]{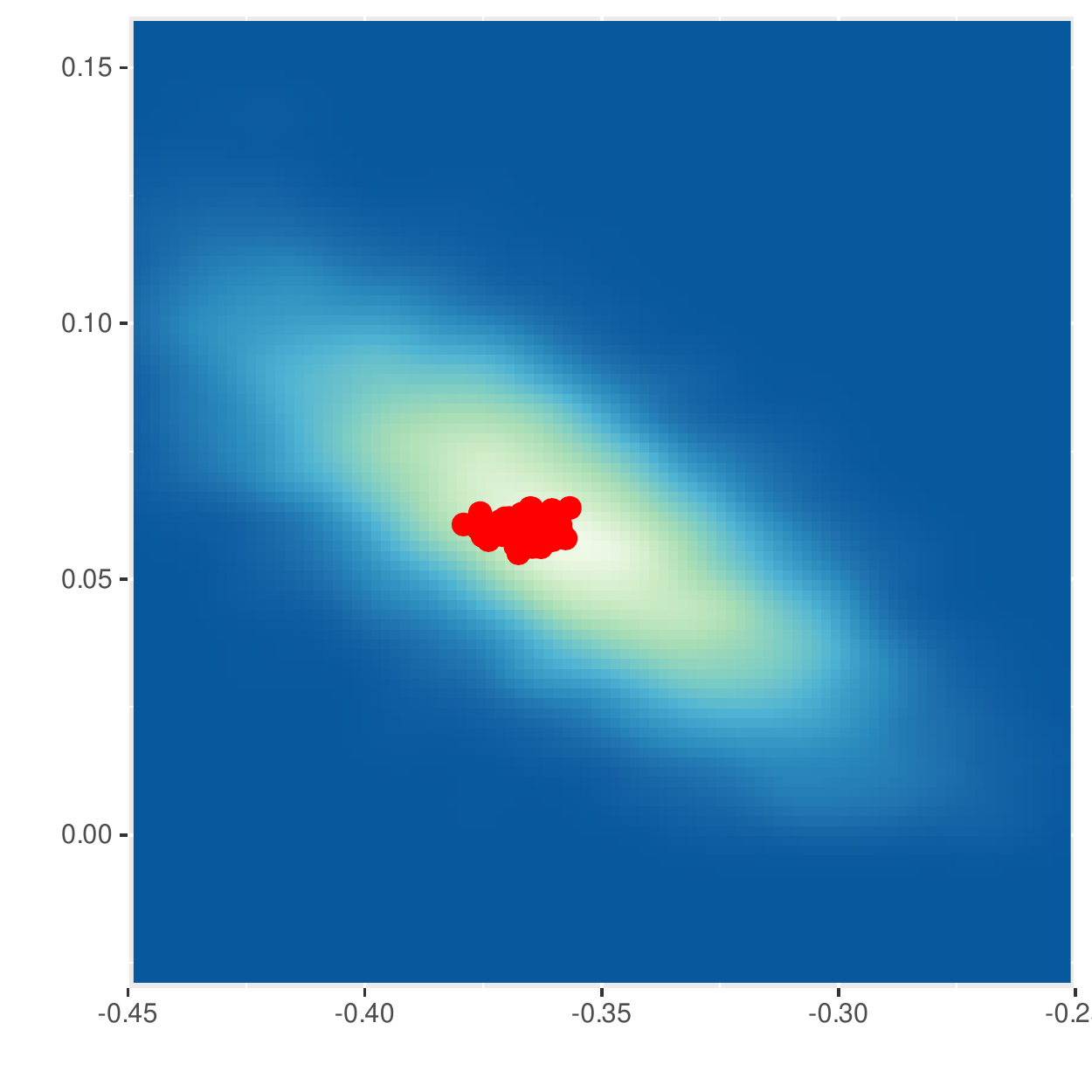}}
		\hfill
		\subfloat[1-W distance = 0.017]{\includegraphics[width=0.33\textwidth, keepaspectratio]{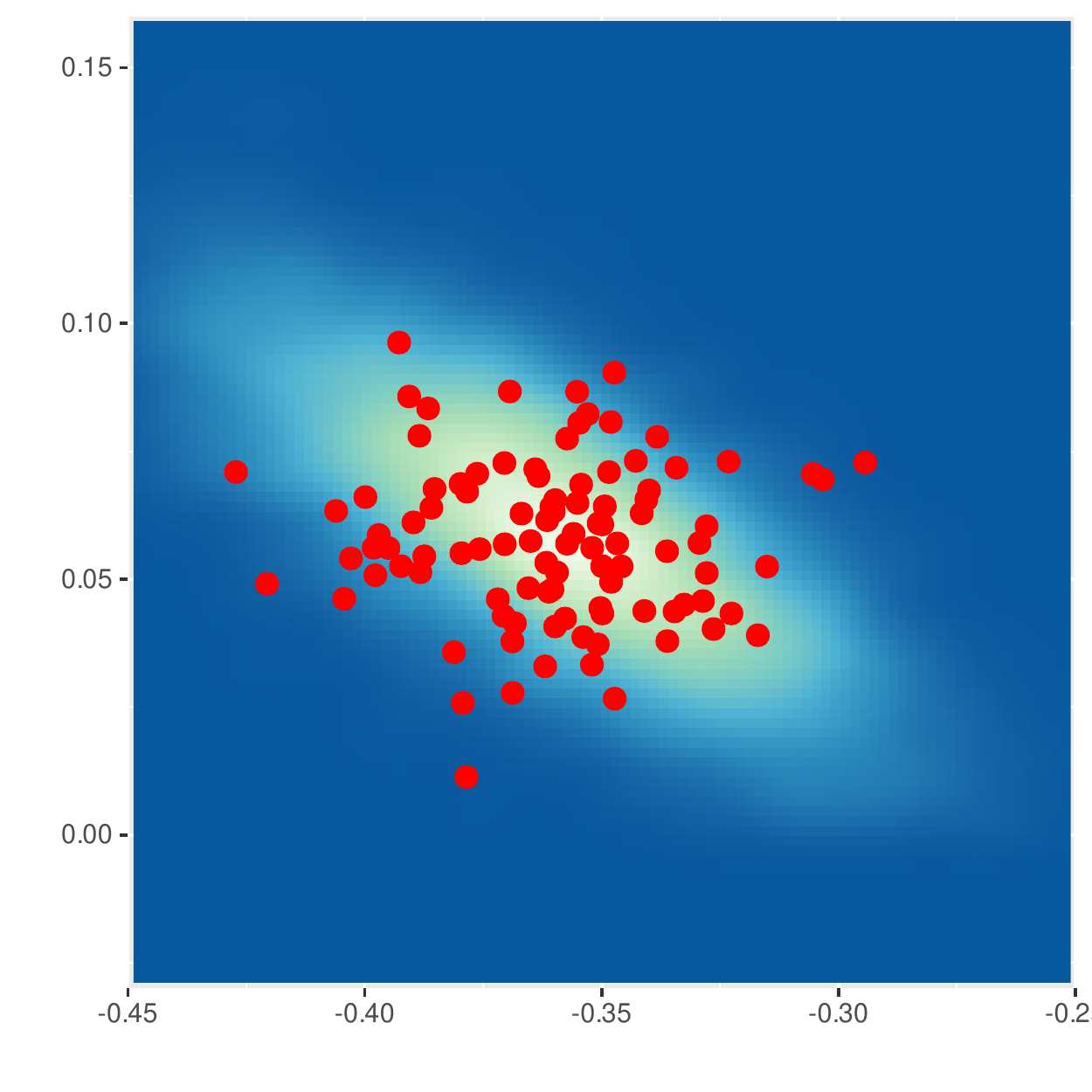}}
		\hfill
		\subfloat[1-W distance = 0.009]{\includegraphics[width=0.33\textwidth, keepaspectratio]{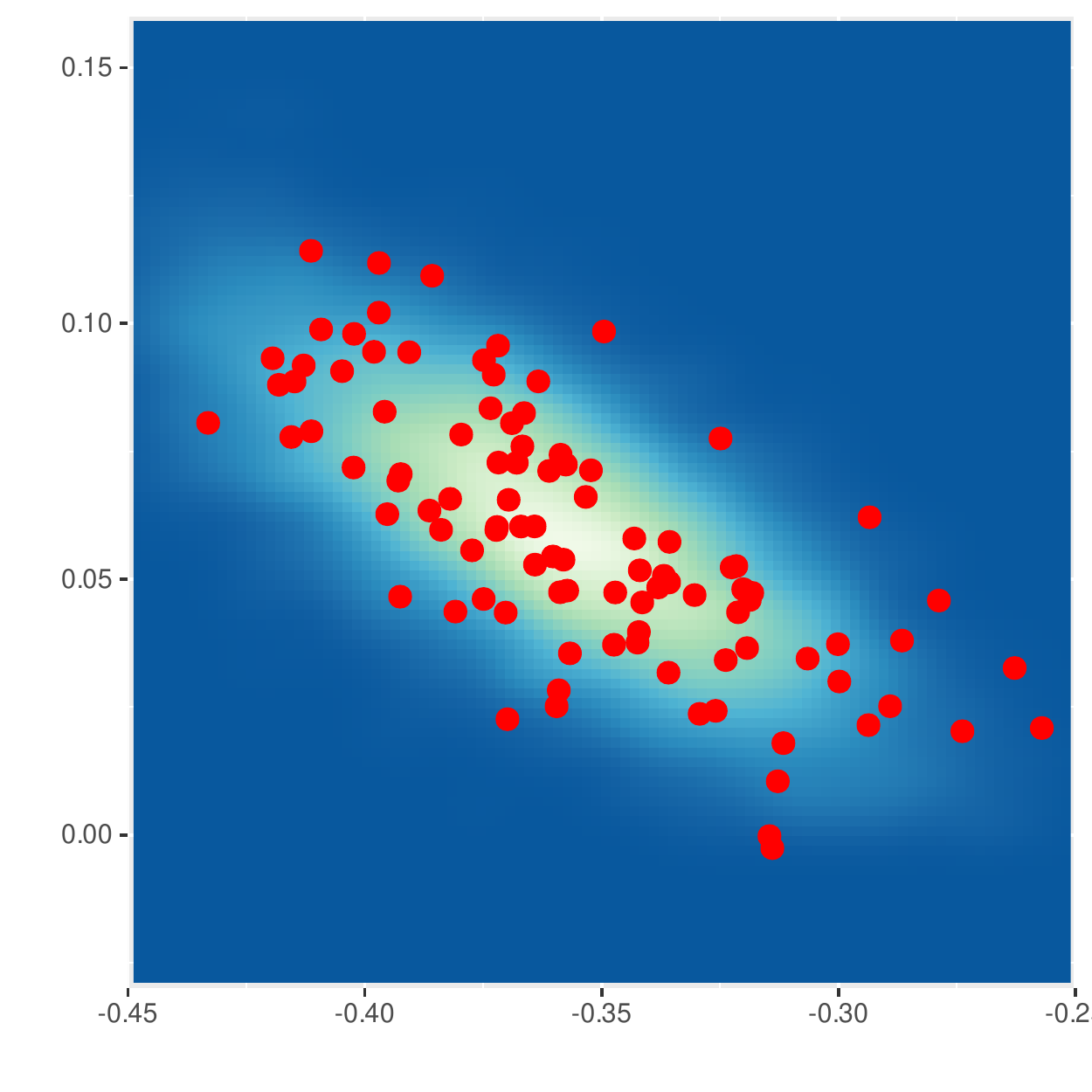}}
	}
	\caption{\it Scatterplot of the 100 approximate draws for the \texttt{dogs-dogs\_log} example from \texttt{posteriordb}.  Higher target densities are shaded lighter.  The subtitles provide 1-Wasserstein (1-W) distances to the target density.}\label{fig: 1-ark_1}
\end{figure}

Both multi-path Pathfinder and Stan's current no-U-turn sampler generate nonparametric approximations. In simulation studies in Section~\ref{sec: simulation}, we use the empirical distribution of 100 approximate draws to measure the Wasserstein distance for approximations generated by multi-path Pathfinder and Stan's no-U-turn sampler. We use the empirical distribution based on the 10,000 reference draws from \texttt{posteriordb} to represent the target posterior distribution. In order to compare multi-path Pathfinder and Stan's no-U-turn sampler to single-path Pathfinder and ADVIs, a discrete form of Wasserstein distance is applied to 100 samples drawn from the approximate distributions produced by single-path Pathfinder and ADVIs. In summary, we estimate distances between reference draws from the target posterior and the approximate draws from Pathfinder (single and multi-path), ADVI (with diagonal and dense covariance), and Stan's current no-U-turn sampler. We provide an example in Figure~\ref{fig: 1-ark_1} to demonstrate the 1-Wasserstein distances between reference samples and approximate draws with different patterns.\footnote{We use the function \texttt{wasserstein()} from the R package \texttt{transport} \citep{transport_pkg} to calculate the 1-Wasserstein distance between two sets of draws.}  

\section{Importance sampling}\label{appendix: importance_sampling}
Importance sampling is a method for adjusting expectation estimates based on an approximate proposal distribution so that they more closely resemble a target distribution. Rather than weighting the draws equally, importance sampling reweights them according to the ratio of the target density to the proposal density.  
If the proposal distribution is $q(\theta)$ and the target $p(\theta)$, the importance weight of a draw $\theta \sim q(\theta)$ is
$w(\theta) = p(\theta) /q(\theta)$.
The importance sampling estimate of an expectation with respect to importance weighted expectation estimate over a target density $p$ given a sequence of proposal draws $\samp{\theta}{1},\ldots, \samp{\theta}{M} \sim q(\theta)$ is
\[
\mathbb{E}_{p(\theta)}[f(\theta)]
\approx \frac{1}{M} \sum_{m = 1}^M w(\samp{\theta}{m}) \cdot f(\samp{\theta}{m}).
\]

\section{ELBO comparison for simulation studies in Section~\ref{subsec: sim_pf_ADVI} }\label{appendix: ELBO_compar}

In the body of the paper, we used Wasserstein-1 distance to evaluate the quality of approximate draws from a target distribution.  In
Figure~\ref{fig: pfvsADVI_compar}, we show the corresponding values of the ELBO derived by single-path Pathfinder and ADVI in both its diagonal (``mean-field'') and dense (``full rank'') versions for the collection of \texttt{posteriordb} models we considered in Section~\ref{subsec: sim_pf_ADVI}.
In all cases, we use 100 approximate draws to compute a Monte Carlo estimate of ELBO by \eqref{eq: ELBO_est}.   The ELBO comparison is consistent with the 1-Wasserstein distance comparison in Section~\ref{subsec: sim_pf_ADVI}.
Pathfinder results in better (higher) ELBO values than ADVI for most of the examples from \texttt{posteriordb}. Meanwhile, for model \verb|mcycle_gp-accel_gp|, \verb|eight_schools-eight_schools_noncentered| and \verb|eight_schools-eight_schools_centered|, the advantages of ADVI over Pathfinder are more prominent in ELBO comparison than in Wasserstein distance comparison. Based on the case studies in Section~\ref{subsec: case_study_pf}, the ELBO more heavily penalizes underdispersed approximations than Wasserstein distance when the approximate samples are close to the high probability mass region. 

\begin{figure}[t!]
	\centering{
		\includegraphics[width=.9\textwidth, keepaspectratio]{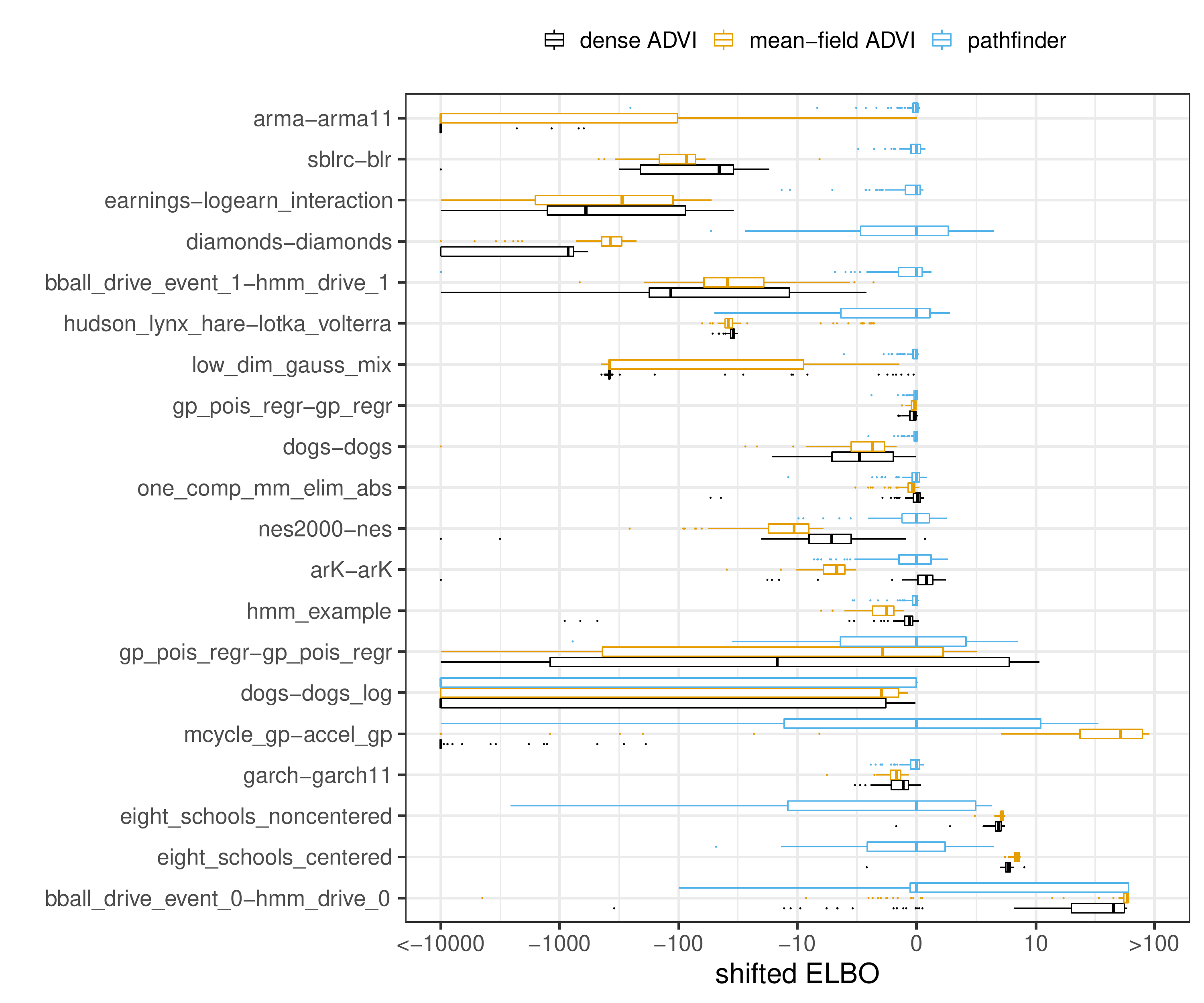}}
	\caption{\it Box plots of estimated ELBO values (higher is better) for single-path Pathfinder and ADVI for the 20 examples in \texttt{posteriordb}. Each box plot displays estimated ELBOs of 100 independent runs of (single-path) Pathfinder, mean-field ADVI, and dense ADVI. We estimate the ELBO with 100 approximate draws from each run.  All values are shifted by the median of the estimated ELBO values for single-path Pathfinder.}\label{fig: pfvsADVI_ELBO_compar}
\end{figure}

\section{Laplace approximation for simulation studies in Section~\ref{subsec: sim_pf_ADVI} }\label{appendix: Laplace_compar}

\begin{figure}[t!]
	\centering{
		\includegraphics[width=.9\textwidth, keepaspectratio]{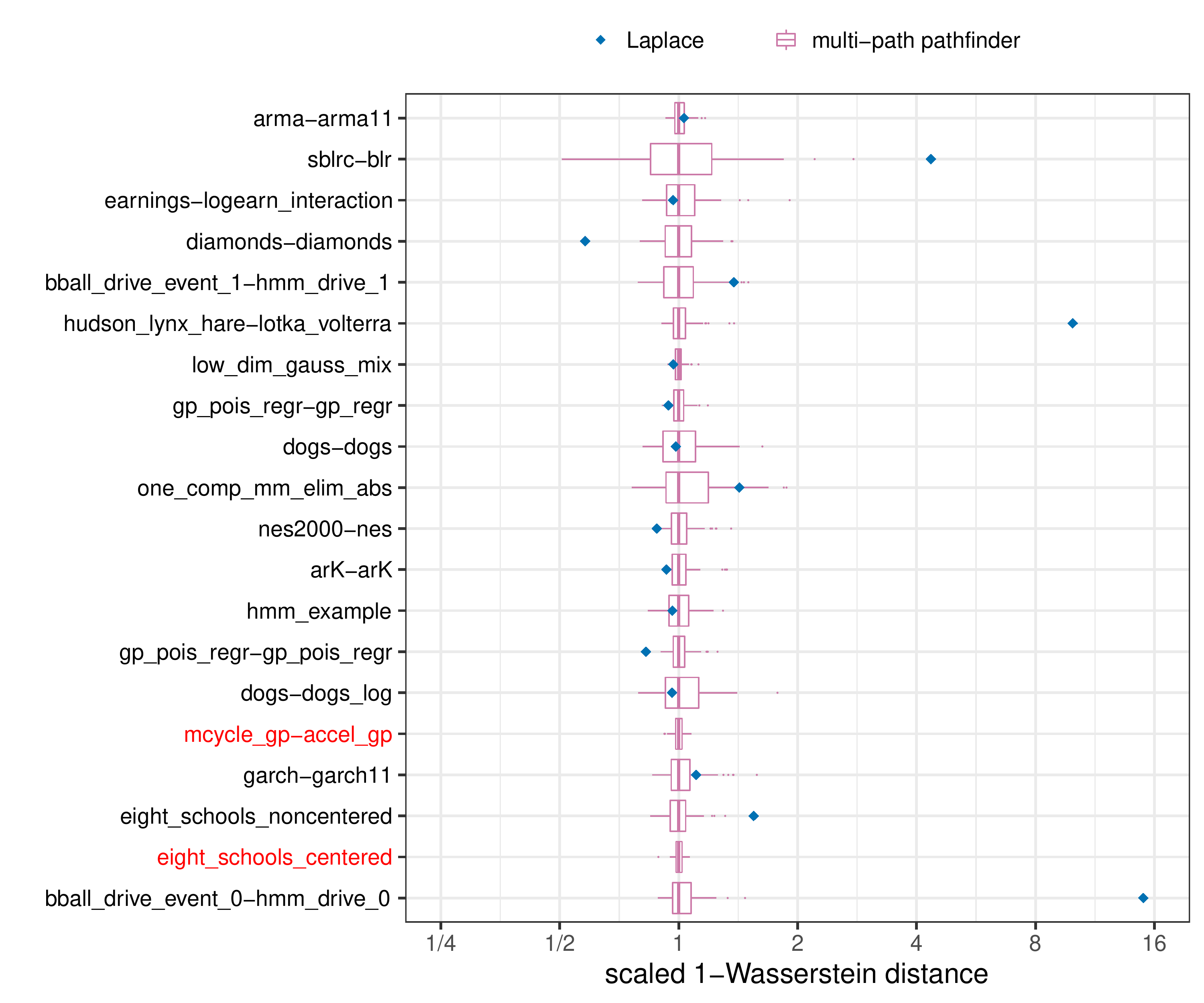}}
	\caption{\it Box plots of 1-Wasserstein distances between the reference posterior samples and approximate draws from multi-path Pathfinder and Laplace approximation for the 20 models in \texttt{posteriordb}. Each box plot displays 1-Wasserstein distances of 100 independent runs of (multi-path) Pathfinder and Laplace approximation. Distances for each model are scaled by the median of the 1-Wasserstein distances for multi-path Pathfinder. The Laplace approximation is not available for examples \texttt{eight\_schools-eight\_schools\_centered} and \texttt{mcycle\_gp-accel\_gp}, since the Hessian at their modes are singular.}\label{fig: pfvsLaplace_compar}
\end{figure}

In this section, we present an experiment to compare Pathfinder with standard Laplace approximation. The comparison is based on the same examples in the simulation study in Section~\ref{subsec: sim_pf_ADVI}. We define the standard Laplace approximation as the Laplace approximation at the mode found by optimization. To generate standard Laplace approximations, we first use L-BFGS to generate an optimization trajectory until it converges. Then we compute the Hessian of the negative log density at the end of the optimization path by function \texttt{jacobian()} from the R package \texttt{numDeriv}. The L-BFGS for finding mode and the L-BFGS in Pathfinder share the same tuning parameters except the maximum iteration. The Laplace is the Normal approximation with mean at the mode and covariance matrix equal to the inverse of the computed Hessian. We generate $100$ approximate draws from the Laplace approximation and compute the 1-Wasserstein distance between the empirical distribution of the approximate draws and the target distribution. In Figure~\ref{fig: pfvsLaplace_compar} we compare multi-path Pathfinder with Laplace approximation based on 1-Wasserstein distances. The distances are scaled by the median of the 1-Wasserstein distance for multi-path Pathfinder for each example.

Normal approximation at the mode (aka Laplace) and Pathfinder are both based on optimization. If the mode exists, Pathfinder's path will also end at the mode. In case of Laplace, we trust that the mode is a good center of the normal approximation. In case of Pathfinder we use ELBO estimates to check if some other place along the path would be better. If the ELBO could be estimated exactly, in case of low-dimensional normal posterior, Pathfinder's approximation would match the Laplace approximation. As ELBO is estimated using a small number of Monte Carlo draws, there is additional variability and Pathfinder's approximation can be little off compared to the Laplace. In Figure~\ref{fig: pfvsLaplace_compar} for nearly normal posteriors like example \verb|nes2000-nes|, due to the error of the ELBO estimates, Pathfinder might pick approximations slightly inferior to the Laplace approximation. In Section~\ref{sec: discussion} we discuss ways to improve the accuracy of ELBO estimation near the optimal ELBO. In the experiments we intentionally used low $J$ for the low-rank part of Pathfinder's approximation. Thus for close to normal posteriors with posterior dependencies, Laplace with dense covariance matrix beats the low-rank plus diagonal approximation even if that approximation would also be at the mode. The effect of this is clearly seen in case of \verb|diamonds-diamonds| posterior that has close to normal posterior with strong dependencies in 25 dimensions. If such strong dependencies would be expected, Pathfinder's accuracy can be trivially improved by increasing $J$ which naturally then increases also memory and computation costs. As shown in Figure~\ref{fig: sens_test_J}, the Wasserstein distances for single-path Pathfinder can be reduced by half with a larger $J$. On the other hand, the implementation of standard Laplace quickly become infeasible as $N$ grows. Standard Laplace not only requires $\mathcal{O}(N^3)$ flops and $\mathcal{O}(N^2)$ memory for obtaining the Cholesky decomposition of the inverse Hessian, the cost of autodiff for Hessian computation is also expensive, especially for high-dimensional problems. If the mode doesn't exist or the second derivatives don't exist at the mode (e.g., for \verb|eight_schools_centered| and \verb|mcycle_gp-accel_go| ), then the Laplace approximation fails but Pathfinder is still likely to be able to provide an approximation. Even if the mode and second derivatives at the mode exist, Pathfinder is safer for posteriors that are far from normal or multimodal. Taking \verb|sblrc-blr| for example, the posterior is skewed due to a small sample size, and we observe that Pathfinder works better than Laplace. We are thus trading some accuracy (due to the stochastic ELBO estimates) in case of close to normal posteriors to faster computation (low rank part) and better accuracy in case of no-normal distributions (use of ELBO).

\section{Pseudocode for L-BFGS and ELBO}\label{appendix: pseudocode}

\begin{algorithm}
	\caption{L-BFGS. 
	}\label{l-bfgs-algorithm.fig}
	\hspace*{\algorithmicindent} \textbf{Input:} \\
	\hspace*{\algorithmicindent} \hspace*{\algorithmicindent} $\log p$: differentiable log density function of dimension $N$\\
	\hspace*{\algorithmicindent} \hspace*{\algorithmicindent} $\rmsup{\theta}{init}$: initial value in support
	with $\log p(\rmsup{\theta}{init})$ finite
	\hfill $(\null \sim \textrm{uniform}(-2,2)^N)$ \\
	\hspace*{\algorithmicindent} \hspace*{\algorithmicindent} $J$: size of the history used to approximate the inverse Hessian \hfill  $(6)$\\
	\hspace*{\algorithmicindent} \hspace*{\algorithmicindent} $\rmsup{\tau}{rel}$: relative tolerance of log density change for convergence
	\hfill $(10^{-13})$\\
	\hspace*{\algorithmicindent} \hspace*{\algorithmicindent} $L$: maximum number of iterations \hfill $(1000)$\\
	\hspace*{\algorithmicindent} \hspace*{\algorithmicindent} $c$: pair of bounds for Wolfe condition on line search, with $0 < c_1 \ll c_2 < 1$
	\hfill $(10^{-4}, 0.9)$\\
	\hspace*{\algorithmicindent} \hspace*{\algorithmicindent} $\epsilon$: positivity threshold for updating covariance estimate
	\hfill $(2.2 \cdot 10^{-16})$\\
	\hspace*{\algorithmicindent} \textbf{Output:}\\
	\hspace*{\algorithmicindent} \hspace*{\algorithmicindent} $\samp{\theta}{1}, \ldots, \samp{\theta}{L'}$: optimization path with $L' \leq L$ and
	$\log p(\samp{\theta}{l}) < \log p(\samp{\theta}{l + 1})$ \\
	\hspace*{\algorithmicindent} \hspace*{\algorithmicindent} $\nabla \log p(\samp{\theta}{1}), \ldots, \nabla \log p(\samp{\theta}{L'})$: gradient log density of points on optimization path 
	\begin{algorithmic}[1]
		\Procedure{L-BFGS}{$\log p, \rmsup{\theta}{init}, J, \rmsup{\tau}{rel}, L, c, \epsilon$}
		\State let $\samp{\theta}{0} = \rmsup{\theta}{init}$, $S = \begin{bmatrix} \, \end{bmatrix}$, $Z = \begin{bmatrix} \, \end{bmatrix}$, and $\alpha = 1_N$, where $1_N$ denotes the $N$-vector of 1s.
		\hfill $\bigo{N}$
		\For{$l \in 0:L - 1$}\hfill $\bigo{L  J^2  N}$
		\State generate $\beta$ and $\gamma$ by \eqref{eq: beta_gamma}
		\hfill $\bigo{J^2  N}$ 
		\State let $\delta = (\textrm{diag}(\alpha) + {\beta} \cdot \gamma \cdot {\beta}^{\top}) \cdot \nabla \log p(\samp{\theta}{l})$ be the search direction
		\hfill $\bigo{J N}$ 
		\For{$\lambda \in 1, \frac{1}{2}, \frac{1}{4}, \ldots$} 
		\State let $\samp{\theta}{l + 1} = \samp{\theta}{l} + \lambda \cdot \delta$ 
		\hfill $\bigo{N}$ 
		\State break if the Wolfe conditions are satisfied,
		\hfill $\bigo{N}$ 
		\begin{eqnarray*}
			\log p(\samp{\theta}{l + 1})
			& \!\!\geq \!\! &
			\log p(\samp{\theta}{l}) + c_1 \cdot \nabla \log p(\samp{\theta}{l})^{\top} \cdot (\lambda \cdot \delta)
			\qquad \qquad \qquad \qquad \mbox{ }
			\\
			\nabla \log p(\samp{\theta}{l + 1})^{\top} \cdot \delta
			& \leq &
			c_2 \cdot \nabla \log p(\samp{\theta}{l})^{\top} \cdot \delta
			\qquad \qquad \qquad \qquad \mbox{}
		\end{eqnarray*}
		\EndFor
		\State return if 
		$\displaystyle 
		\frac{\log p(\samp{\theta}{l + 1}) - \log p(\samp{\theta}{l})}
		{\left| \, \log p(\samp{\theta}{l}) \, \right|}
		< \rmsup{\tau}{rel}$   
		\State let $S_{l + 1} = \samp{\theta}{l+1} - \samp{\theta}{l}$ and $Z_{l+1} = \nabla \log p(\samp{\theta}{l}) - \nabla \log p(\samp{\theta}{l+1})$,
		\If {$S_{l+1}^\top Z_{l+1} > \epsilon \cdot \|Z_{l+1}\|^2\;$ (Condition 3.9 of \citet{byrd1995limited})}
		\State if more than $J - 1$ updates  are stored, delete the first columns of $S$ and $Z$
		\State let $S = \begin{bmatrix} S & S_{l + 1} \end{bmatrix}$
		and $Z = \begin{bmatrix} Z & Z_{l + 1} \end{bmatrix}$
		\State let $\alpha
		=\frac{\displaystyle \|Z_{l+1}\|^2}
		{\displaystyle S_{l+1}^\top \cdot Z_{l+1}} \cdot 1_N$ 
		\hfill $\bigo{N}$
		\EndIf
		\EndFor 
		\EndProcedure
	\end{algorithmic}
\end{algorithm}

\begin{algorithm}
	\caption{ELBO estimation}\label{ELBO-est.fig}
	\hspace*{\algorithmicindent} \textbf{Input:} \\
	\hspace*{\algorithmicindent} \hspace*{\algorithmicindent}
	$\log p(\samp{\phi}{1}), \ldots, \log p(\samp{\phi}{K})$: target log densities \\
	\hspace*{\algorithmicindent} \hspace*{\algorithmicindent} $\log q(\samp{\phi}{1}), \ldots, \log q(\samp{\phi}{K})$: log densities of approximation distribution \\
	\textbf{Output:} \\
	\hspace*{\algorithmicindent} \hspace*{\algorithmicindent}
	$\widehat{\footnotesize \textrm{ELBO}}$: Monte Carlo estimate of the evidence lower bound
	\begin{algorithmic}[1]
		\Procedure{ELBO}{$\log p(\samp{\phi}{1:K}), \log q(\samp{\phi}{1:K})$}
		\State let
		$\widehat{\textrm{\footnotesize ELBO}} = \frac{1}{K}\sum_{k = 1}^K \log p (\samp{\phi}{k}) - \log q(\samp{\phi}{k})$
		\hfill $\bigo{K}$
		\EndProcedure
	\end{algorithmic}
\end{algorithm}

\label{lastpage}

\end{document}